\documentclass{article}

 \usepackage[final]{neurips_2022}

\usepackage[utf8]{inputenc} 
\usepackage[T1]{fontenc}    
\usepackage{hyperref}       
\usepackage{url}            
\usepackage{booktabs}       
\usepackage{amsfonts}       
\usepackage{nicefrac}       
\usepackage{microtype}      
\usepackage{xcolor}         

\usepackage{xcolor}
\usepackage{siunitx}
\usepackage{tabularx}
\usepackage{graphicx}
\usepackage[nolist]{acronym}
\usepackage{hyperref}
\usepackage{appendix}
\usepackage[position=b]{subcaption}
\usepackage{grffile}
\usepackage{authblk}
\usepackage{floatrow}
\newfloatcommand{capbtabbox}{table}[][\FBwidth]
\usepackage{url}
\usepackage{todonotes}
\setuptodonotes{inline}
\usepackage[inline]{enumitem}
\usepackage{amsmath}
\usepackage[capitalise,noabbrev]{cleveref}

\title{Towards Automated Design of Bayesian Optimization via Exploratory Landscape Analysis}

\author[1]{Carolin Benjamins}
\author[2]{Anja Jankovic}
\author[2,3]{Elena Raponi}
\author[2]{Koen van der Blom}
\author[1]{Marius Lindauer}
\author[2]{Carola Doerr}
\affil[1]{Institute of AI, Leibniz University Hannover, Germany}
\affil[2]{Sorbonne Universit\'e, CNRS, LIP6, Paris, France}
\affil[3]{TUM School of Engineering and Design, TU München, Germany}

\begin{document}

\begin{acronym}
\acro{DoE}{\emph{design of experiment}}
\acro{BO}{Bayesian optimization}
\acro{AF}{\emph{acquisition function}}
\acro{EI}{Expected Improvement}
\acro{PI}{Probability of Improvement}
\acro{UCB}{Upper Confidence Bound}
\acro{ELA}{exploratory landscape analysis}
\acro{GP}{Gaussian Process}
\acro{TTEI}{Top-Two Expected Improvement}
\acro{TS}{Thompson Sampling}
\acro{DAC}{Dynamic Algorithm Configuration}
\acro{AC}{Algorithm Configuration}
\acro{CMA-ES}{CMA-ES}
\acro{AS}{algorithm selection}
\acro{PIAS}{per-instance algorithm selection}
\acro{PIAC}{per-instance algorithm configuration}
\acro{AFS}{Acquisition Function Selector}
\acro{VBS}{\emph{virtual best solver}}
\acro{RF}{random forest}
\end{acronym}

\maketitle

\begin{abstract}
Bayesian optimization (BO) algorithms form a class of surrogate-based heuristics, aimed at efficiently computing high-quality solutions for numerical black-box optimization problems. The BO pipeline is highly modular, with different design choices for the initial sampling strategy, the surrogate model, the acquisition function (AF), the solver used to optimize the AF, etc. We demonstrate in this work that a dynamic selection of the AF can benefit the BO design. More precisely, we show that already a na\"ive random forest regression model, built on top of exploratory landscape analysis features that are computed from the initial design points, suffices to recommend AFs that outperform any static choice, when considering performance over the classic BBOB benchmark suite for derivative-free numerical optimization methods on the COCO platform. Our work hence paves a way towards AutoML-assisted, on-the-fly BO designs that adjust their behavior on a run-by-run basis. 
\end{abstract}

\section{Introduction}

In optimization we often encounter black-box problems 
having no explicit formulation of the underlying function or its derivatives. To identify high-quality solutions for such functions, black-box algorithms rely on an iterative process of querying the quality of the solution candidates it suggests, and adjusting their strategy to obtain the next set of candidate solutions. 
This process is particularly challenging when the maximal number of evaluations that can be made before a final recommendation is expected is very small compared to the size of the design space.\\
A standard approach for such settings is \ac{BO}~\citep{mockus_bayesian_2012}, a family of surrogate-based optimization algorithms that obtains its candidate solutions via the following process:
a first set of candidates, known as the \emph{initial design} or \ac{DoE}, is obtained from a classic sampling strategy, e.g., (quasi-)random distributions or space filling designs.
After evaluation, these points are used to build a surrogate model, an approximation of the unknown objective function capturing the uncertainty.
An \ac{AF} (also: \emph{infill criterion}) determines which point to sample next based on the predicted mean and variance of the surrogate model, requiring no evaluations of the true problem.
As long as the budget is not exhausted, the surrogate model is updated and the algorithm repeats the last steps.\\
All steps of the \ac{BO} pipeline are subject to different design choices, impacting its performance
~\citep{lindauer-dso19a,cowenrivers-arxiv21a,BossekDK20}.
Of particular importance for the overall \ac{BO} performance is the balance between exploration and exploitation to efficiently find the optimal solution, which is strongly influenced by the choice of the \ac{AF}.
Although many different \ac{AF}s have been introduced (e.g., \acf{PI}, \acf{EI}, \ac{UCB}~\citep{forrester_engineering_2008} and \ac{TS}~\citep{thompson-biomet33a}), there are no guidelines on how to select the most appropriate \ac{AF} given the characteristics of the problem at hand. 
Furthermore, \ac{AF} design decisions have predominantly considered a \emph{static} choice over the entire \ac{BO} procedure.
Only few works have considered \emph{dynamic} choices when selecting \ac{AF}s~\citep{hoffman-uai11,kandasamy-jmlr2020,cowenrivers-arxiv21a}, and similarly to insights obtained in other domains of optimization, have shown the potential of opting for dynamic combinations \ac{AF} rather than a static choice.\\
Orthogonal to this, performance gains may also be possible through the application of \ac{AS} techniques~\citep{rice-aic76a}.
Given sufficiently heterogeneous problem instances, and \emph{based exclusively on their characteristics}, choosing different \ac{AF}s or \ac{AF} schedules might result in better performance than applying the same \ac{AF} (schedule) to all problems, which is known as \ac{PIAS}~\citep{kerschke_automated_2019}.
More importantly, we observe that the \emph{properties of the initial design itself} further influence the choice of the \ac{AF}, so we may want to select different \ac{AF}s per run, even for the same problem instance.
This is in line with \emph{per-run algorithm selection}, as suggested in~\citep{DBLP:conf/cec/JankovicVKNED22, DBLP:conf/ppsn/KostovskaJVNWED22} for a portfolio of iterative optimization heuristics.

\begin{figure}[t]
    \centering
    \includegraphics[width=0.8\textwidth]{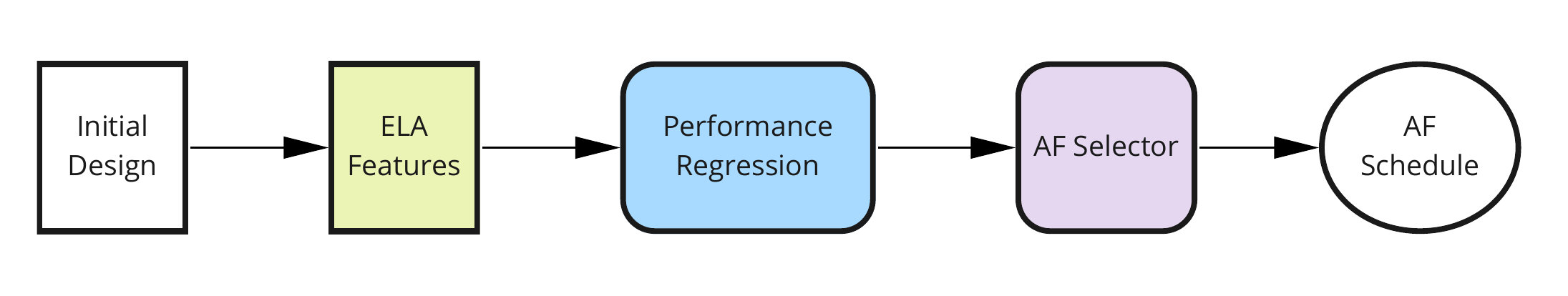}
    \caption{\textbf{How to select acquisition function (AF) schedules } Based on the evaluated initial design of a \ac{BO} run, we compute the ELA features. With those we query a trained RF and observe the predicted performance of each AF schedule. The \acl{AFS} returns the AF schedule with the best predicted performance to use for surrogate-based evaluations.}
    \label{fig:method}
\end{figure}

Motivated by recent insights from the \ac{AS} domain, we aim to understand the interplay between the problem (characterized through the initial design), the choice of the \ac{AF} (schedule), and the \ac{BO} performance.
To this end, we benchmark our approach on a test suite of well-established black-box numerical optimization problems, the BBOB functions of the COCO framework~\citep{hansen-joms21}. 
We build and train a model, called the \ac{AFS}~(\cref{fig:method}), which automatically selects the best suited \ac{AF} (schedule) for a certain problem.
To represent problems, we employ \emph{\ac{ELA}}~\citep{mersmann-gecco11}.
\ac{ELA} is a framework for the extraction and computation of low-level numerical landscape features of black-box problems (related to e.g., multimodality, separability, conditioning, presence of plateaus, etc.) from sampled points and their evaluations.
Here, we use the evaluated \emph{initial design} for computing the \ac{ELA} features. 
Since \ac{BO} anyway evaluates the initial design before starting the surrogate-based optimization process, the use of \ac{ELA} features comes at no evaluation cost.
Our \ac{AF} schedules are composed of either \ac{PI} (more exploitative), \ac{EI} (more explorative) or combinations of both.
Our key insights are that
\begin{enumerate*}[label=(\roman*)]
\item already a selector built on top of a na\"ive random forest regression model outperforms any static \ac{AF} schedule, 
\item the choice of the \ac{AF} model should depend on the problem instance, but also on the characteristics (and in particular the quality) of the initial design, and 
\item the schedules that switch from EI to PI improve over static EI and PI.   
\end{enumerate*}




\section{Experiments}
\label{ssec:exp}
Building our \acf{AFS} (see~\cref{fig:method}) consists of two steps.
First, we gather performance data of our seven AF schedules (static \ac{EI}, static \ac{PI}, and five dynamic ones: random, round robin, as well as three variants of explore-exploit schedules that switch from \ac{EI} to \ac{PI} after $\SI{25}{\percent}$, $\SI{50}{\percent}$~and $\SI{75}{\percent}$ of the budget).
Second, we train our \ac{AFS} in the following way.
We build a multi-target regression model that maps the \ac{ELA} features of the initial design to the performance of each \ac{AF} schedule.
Our \ac{AFS} then selects the schedule with the best predicted performance.\\
We evaluate our schedules on the \num{24} single-objective noiseless BBOB functions of the COCO benchmark~\citep{hansen-joms21} in dimension \num{5} and \num{5} instances with \num{60} seeds).
The size of the initial design is set to $10d=50$ data points and a budget of $40d=100$ function evaluations is used for the surrogate-based optimization part. 
The initial design is the same for every seed, so every schedule sees the same initial design.
We adapt SMAC3~\citep{lindauer-jmlr22a} to enable dynamic \ac{AF} schedules and we use a \ac{GP} as surrogate model.
We extract \ac{ELA} features based on samples of the initial design and their function evaluations using the \texttt{flacco} library~\citep{flacco2019}.
Among the \num{300} features (grouped in sets) in the library, we choose \num{38} cheap-to-compute features not requiring further function evaluations,  namely those from y-Distribution, Meta-Model, Dispersion, Information Content and Nearest-Better Clustering feature sets, following practices in the literature~\citep{BelkhirDSS16, DBLP:conf/evoW/JankovicED21}.
In our setup, \ac{ELA} feature computation for an initial design of \num{50} points takes about $\SI{0.4}{\second}$, which is negligible compared to an expensive function evaluation.
In this work, we restrict our attention to continuous problems, but \ac{ELA} may also work well for discrete (and possibly mixed-integer) search spaces~\citep{pikalov-evo22}.
We then train a standard \ac{RF}~\citep{ho1995random, scikit-learn} model ($70/30$ train/test split) to regress the performance (log regret, normalized per problem) for each schedule based on the input ELA features.
The schedule with the best predicted performance is selected by the \ac{AFS}.\\
All experiments were conducted on a Slurm CPU cluster with \num{1592} CPUs available across nodes.
For visualization, we consider the log-regret of the incumbent (best evaluated  search point) and only visualize the evaluations proposed by \ac{BO}, omitting the initial design.
In the violin plots, the log-regret of the final incumbents of all \num{60} seeds are normalized to $[0,1]$ per BBOB function.
In the convergence plots, the log-regret of the incumbents are normalized across runs to $[0,1]$ per BBOB function, and the means with the \SI{95}{\percent} confidence intervals are shown.
Ranking is computed per run, i.e.~per seed, BBOB function and instance.
The \ac{AFS} is assigned the same rank as the schedule it chose.
We only plot performance on the test data.
The plots for each function can be found in~\cref{sec:all_bbob_plots}.
You can find the repository here: \url{https://github.com/automl/BO-AFS}.

\begin{figure}[t]
    \centering
    \begin{subfigure}[b]{0.25\textwidth}
        \centering
        \includegraphics[width=\textwidth]{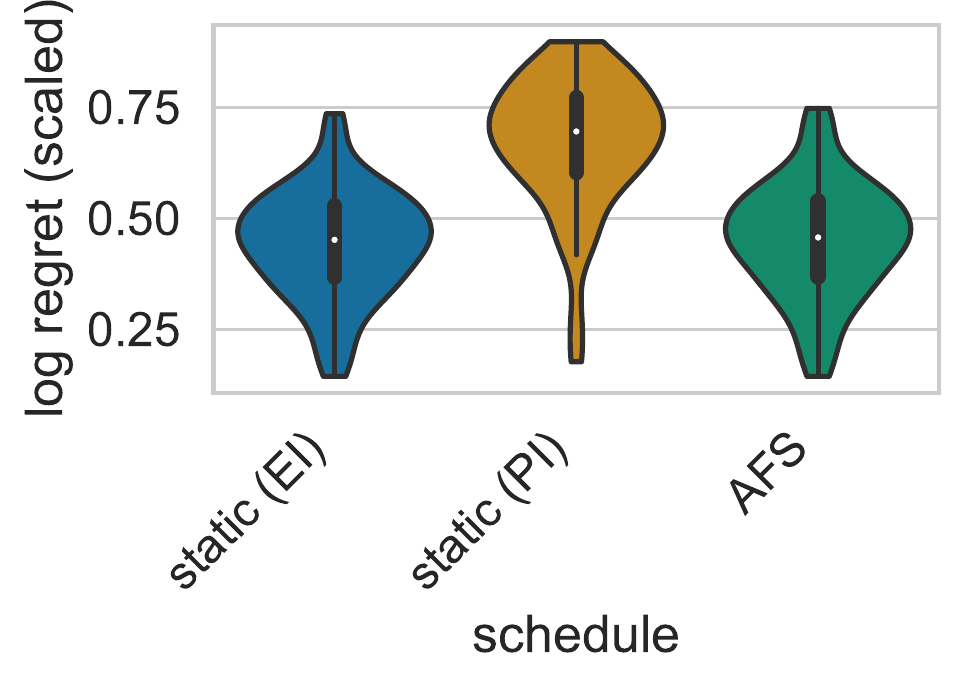}
        \caption{BBOB Function 7}
        \label{subfig:eipiafs_1}
    \end{subfigure}
    \hfill
    \begin{subfigure}[b]{0.25\textwidth}
        \centering
        \includegraphics[width=\textwidth]{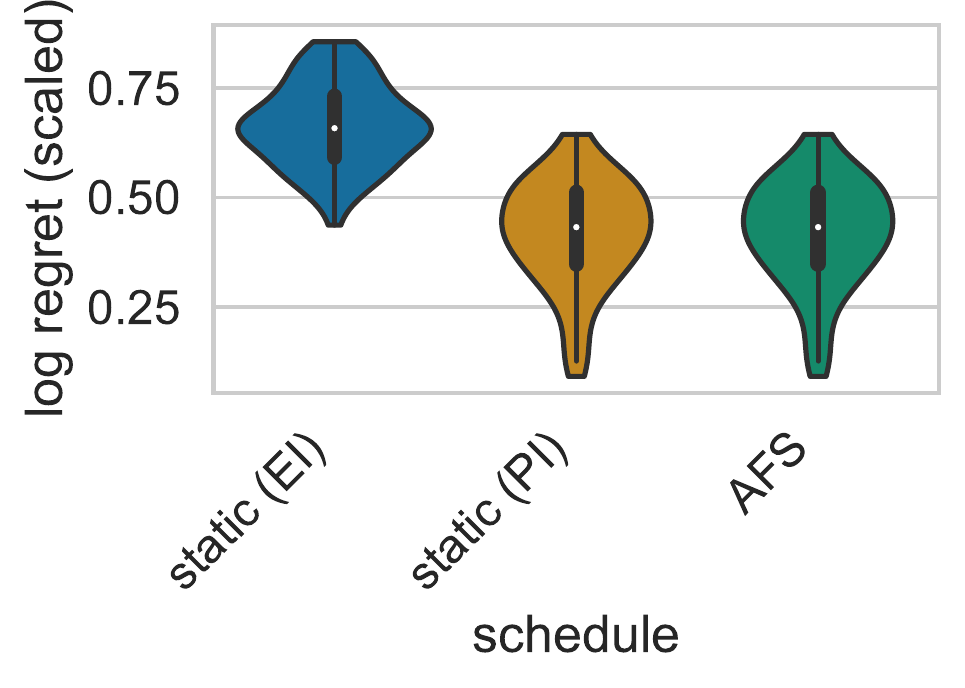}
        \caption{BBOB Function 22}
        \label{subfig:eipiafs_2}
    \end{subfigure}
    \hfill
    \begin{subfigure}[b]{0.45\textwidth}
        \centering
        \includegraphics[width=\textwidth]{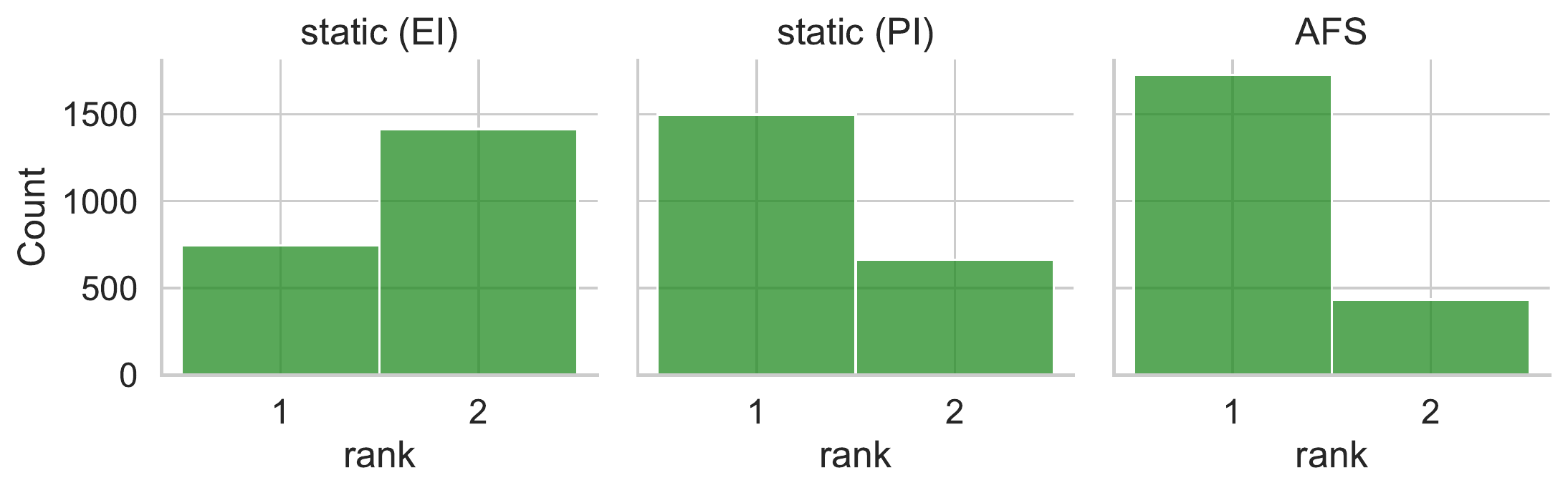}
        \caption{Ranks aggregated over all BBOB functions.}
        \label{subfig:eipiafs_rank}
    \end{subfigure}
    \caption{\ac{AFS} selecting between static PI and EI adopts the shape of the better performing schedule and ranks first.}
    \label{fig:eipiafs}
\end{figure}

\paragraph{Results}
Before we learn to select from our portfolio containing static and dynamic schedules, we first consider selecting either static \ac{PI} or static \ac{EI}.
We can observe that \ac{AFS} adopts the shape of the better performing acquisition function (\cref{subfig:eipiafs_1,subfig:eipiafs_2}) and aggregated over all functions ranks first (\cref{subfig:eipiafs_rank}), demonstrating the potential to select between \ac{AF}s based on the initial design.
\begin{figure}[ht]
\begin{floatrow}
\ffigbox{%
  \includegraphics[width=0.48\textwidth]{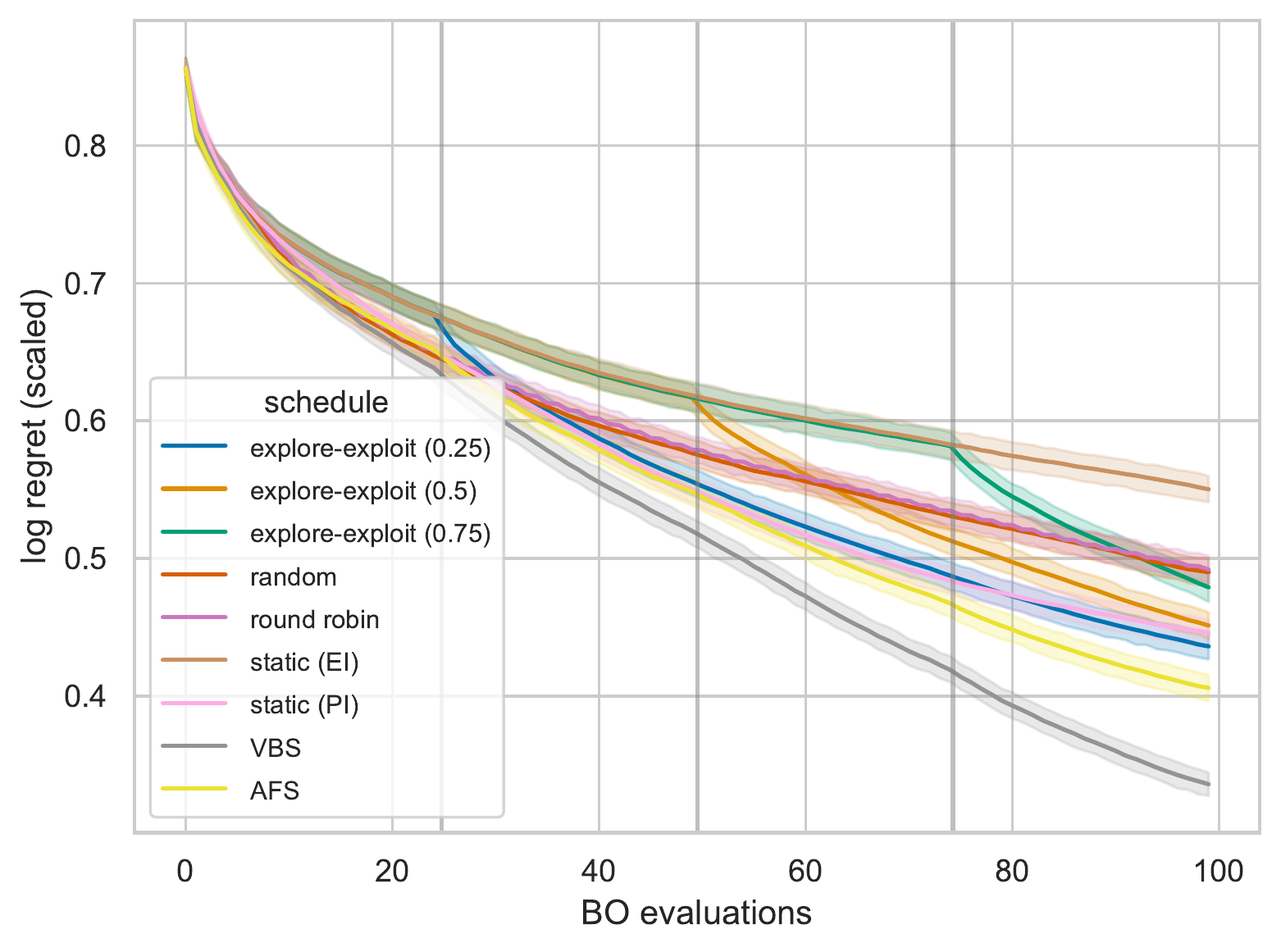}
}{%
  \caption{Switching \ac{AF}s and especially selecting \ac{AF} schedules (\ac{AFS}) is beneficial. Scaled log-regret averaged over \num{5} instances of all \num{24} BBOB functions for \num{5} dimensions with the \SI{95}{\percent} confidence interval for the schedules from~\cref{tab:schedules}.}
    \label{fig:convergence_all_ci}
} \quad
\ffigbox{%
 \includegraphics[width=.48\textwidth]{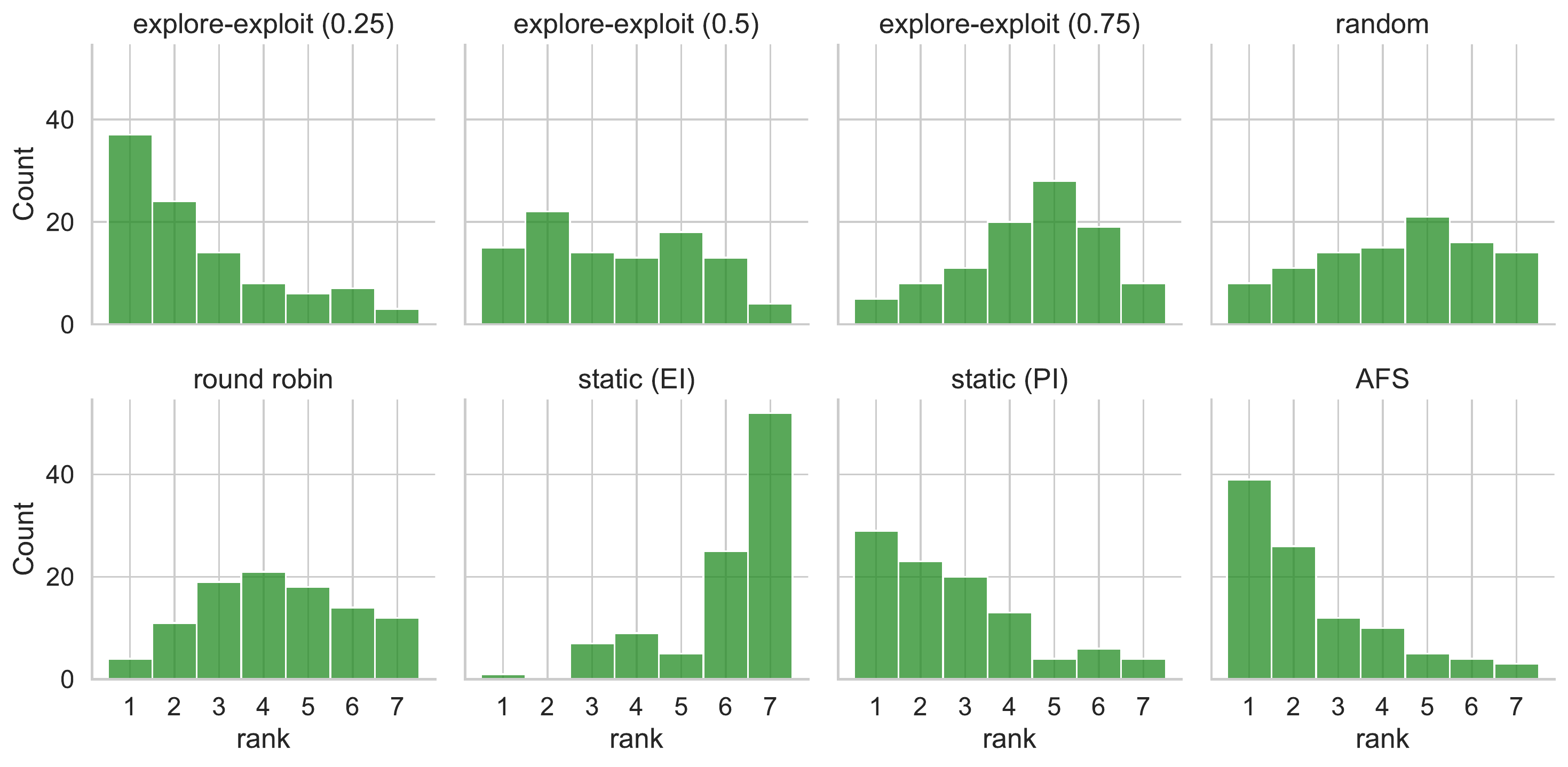}
}{
\caption{Ranks of schedules, aggregated over all BBOB functions, instances and seeds. \ac{AFS} has the most favorable rank distribution.}
    \label{fig:rank}
}

\end{floatrow}
\end{figure}
We then extend the \ac{AFS} to select between all seven proposed schedules.
We introduce the \ac{VBS}, which is our oracle, always selecting the best schedule per run.
The general optimal choice of static or dynamic \ac{AF} schedules depends on the problem at hand (see figures in~\cref{sec:all_bbob_plots}).
As a rule of thumb, switching from \ac{EI} to \ac{PI} after a certain percentage of the budget of total number of function evaluations is beneficial and can boost performance, see~\cref{fig:convergence_all_ci}.
However, the optimal switching point is highly problem-dependent and likely run-dependent.
On average, switching after $\SI{25}{\percent}$ seems to be a viable general rule if the problem's properties are unknown.
Moreover, due to the initial design, we gain information about the problem landscape based on \ac{ELA} features for that specific run.
Already with a simple, non-tuned \ac{RF}, we are able to select well-performing schedules with the \ac{AFS}.
In this case, the \ac{AFS} (apart from the \ac{VBS}) also shows the best anytime-performance which is not necessarily guaranteed as the \ac{AFS} is trained on the final performance of the other schedules.
Again, we rank across all functions and instances and observe that the mass of ranks for \ac{AFS} is centered on the first, in contrast to all other schedules, see~\cref{fig:rank}.
In summary, our \ac{AFS} is able to propose appropriate schedules from a portfolio based on the initial design per run.

\section{Conclusion and Future Work}
Using a standard test bed for benchmarking single-objective numerical black-box optimization techniques, the BBOB suite of the COCO environment, we have shown in this work that a per-run selection of the acquisition function can benefit the performance of \acf{BO} approaches.
More precisely, we have seen that a landscape-aware selection of the \ac{AF}, choosing from a portfolio of seven different \ac{AF} schedules, outperforms any static choice.
This advantage was realized with a na\"ive random forest regression approach, and it seems likely that further improvements can be expected from a proper model selection, e.g., using \texttt{auto-sklearn}~\citep{feurer-arxiv20a}. \\
Our long-term objective is the development of modular \ac{BO} algorithms that are trained to select their modules, including the \ac{AF}, on the fly, i.e., during the optimization process.
Such \emph{dynamic algorithm configuration} (DAC)~\citep{biedenkapp-ecai20a,biedenkapp-gecco22a} approaches were recently shown capable of outperforming classic (static) \emph{hyperparameter optimization} approaches.
A key issue in the design of DAC is the identification of features that the trained model can use to guide its selection.
Our results provide strong motivation to consider exploratory landscape analysis for this purpose. Blending the \ac{ELA} features with information obtained from the surrogate model clearly is a natural next step for our work. \\
In our study, we have used a relatively large initial design size ($10d$ points or $1/3$ of the total budget).
Considering~\citep{BossekDK20}, an intermediate decision whether to continue with the initial design or to start the surrogate-based optimization should bring additional performance gains.
To realize this potential, a thorough investigation of suitable \ac{ELA} features is needed, since some features cannot be considered reliable when based on such comparatively small sample sizes~\citep{BelkhirDSS16}. \\
We see our work as a continuation of the per-run algorithm selection approach suggested in~\citep{DBLP:conf/cec/JankovicVKNED22, DBLP:conf/ppsn/KostovskaJVNWED22}.
In particular, we believe that it is not only the instance per se that should guide the selection of algorithm components and configurations, but also the particular trajectory that the algorithm follows in a particular run.
In the future, we aim to develop a deeper understanding for when to switch from initial design sampling to using explorative \ac{AF}s or exploitative ones.
While in this first proof-of-concept study we have only considered EI and PI as \ac{AF}s, in future studies we intend to include other \ac{AF}s such as \acf{UCB}~\citep{forrester_engineering_2008}, \acf{TTEI}~\citep{qin_improving_2017}, and \acf{TS}~\citep{thompson-biomet33a}.

\subsection*{Acknowledgments}
Our work has been financially supported by the the ANR T-ERC project \emph{VARIATION} (ANR-22-ERCS-0003-01), by the CNRS INS2I project \emph{RandSearch}, and by the PRIME programme of the German Academic Exchange Service (DAAD) with funds from the German Federal Ministry of Education and Research (BMBF), and by RFBR and CNRS, project number 20-51-15009.
Carolin Benjamins and Marius Lindauer acknowledge funding by the German Research Foundation (DFG) under LI 2801/4-1.







\bibliography{bib/lib,bib/references,bib/shortproc}
\bibliographystyle{unsrtnat}

\newpage
\section*{Checklist}


\begin{enumerate}

\item For all authors...
\begin{enumerate}
  \item Do the main claims made in the abstract and introduction accurately reflect the paper's contributions and scope?
    \answerYes{}
  \item Did you describe the limitations of your work?
    \answerYes{}
  \item Did you discuss any potential negative societal impacts of your work?
    \answerNo{We see no potential negative societal impact because our method is about speeding up an existing black-box optimization algorithm, therefore reducing required resources.}
  \item Have you read the ethics review guidelines and ensured that your paper conforms to them?
    \answerYes{}
\end{enumerate}

\item If you are including theoretical results...
\begin{enumerate}
  \item Did you state the full set of assumptions of all theoretical results?
    \answerNA{}
        \item Did you include complete proofs of all theoretical results?
    \answerNA{}
\end{enumerate}

\item If you ran experiments...
\begin{enumerate}
  \item Did you include the code, data, and instructions needed to reproduce the main experimental results (either in the supplemental material or as a URL)?
    \answerYes{\url{https://github.com/automl/BO-AFS}}
  \item Did you specify all the training details (e.g., data splits, hyperparameters, how they were chosen)?
    \answerYes{}
        \item Did you report error bars (e.g., with respect to the random seed after running experiments multiple times)?
    \answerYes{}
        \item Did you include the total amount of compute and the type of resources used (e.g., type of GPUs, internal cluster, or cloud provider)?
    \answerYes{}
\end{enumerate}

\item If you are using existing assets (e.g., code, data, models) or curating/releasing new assets...
\begin{enumerate}
  \item If your work uses existing assets, did you cite the creators?
    \answerYes{}
  \item Did you mention the license of the assets?
    \answerNA{}
  \item Did you include any new assets either in the supplemental material or as a URL?
    \answerNo{}
  \item Did you discuss whether and how consent was obtained from people whose data you're using/curating?
    \answerNA{We create our own data.}
  \item Did you discuss whether the data you are using/curating contains personally identifiable information or offensive content?
    \answerNA{}
\end{enumerate}

\item If you used crowdsourcing or conducted research with human subjects...
\begin{enumerate}
  \item Did you include the full text of instructions given to participants and screenshots, if applicable?
    \answerNA{}
  \item Did you describe any potential participant risks, with links to Institutional Review Board (IRB) approvals, if applicable?
    \answerNA{}
  \item Did you include the estimated hourly wage paid to participants and the total amount spent on participant compensation?
    \answerNA{}
\end{enumerate}

\end{enumerate}


\appendix
\newpage
\section{Acquisition Function Schedules}
\begin{table}[h]
    \centering
    \begin{tabular}{ll}
        \toprule
         Name & Schedule \\
         \midrule
         static (EI) & \includegraphics[width=3cm]{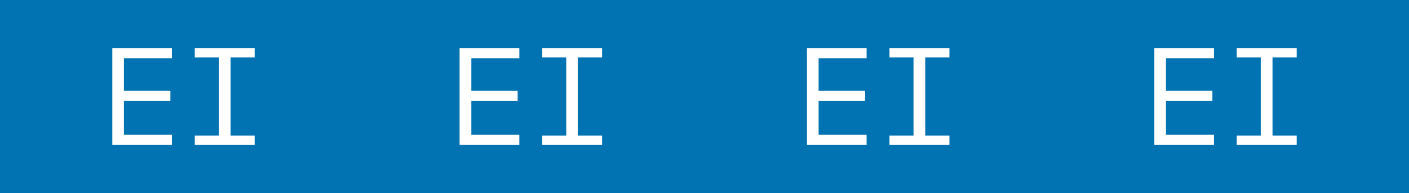}\\
         static (PI) & \includegraphics[width=3cm]{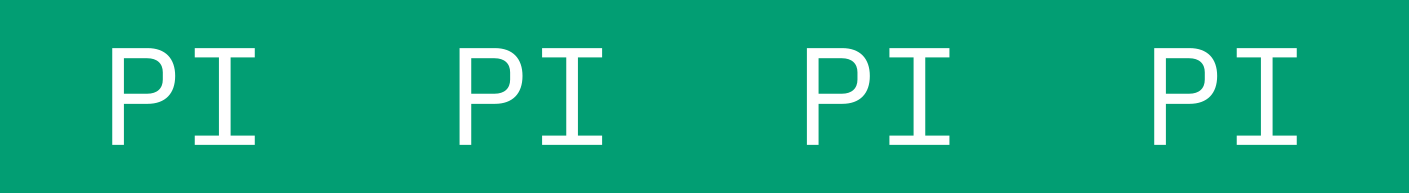}\\
         random & \includegraphics[width=3cm]{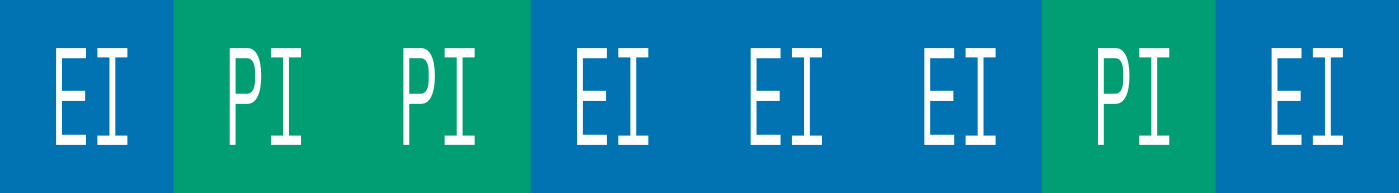}\\
         round robin & \includegraphics[width=3cm]{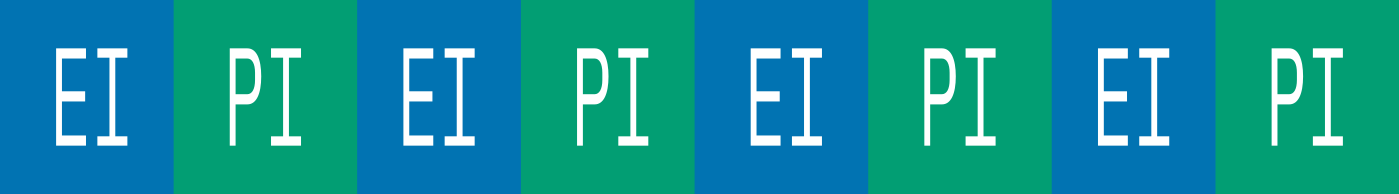}\\
         explore-exploit (\num{0.25}) & \includegraphics[width=3cm]{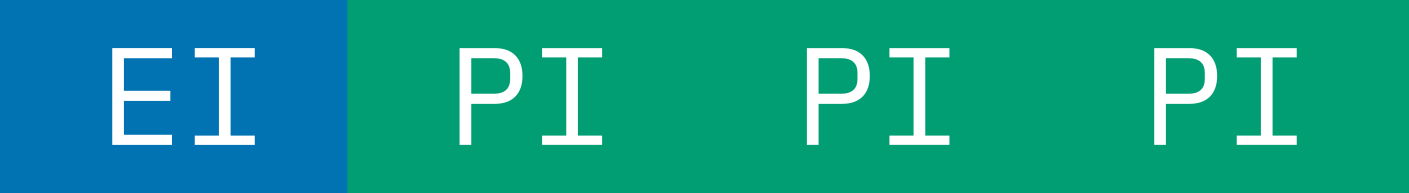}\\
         explore-exploit (\num{0.5}) & \includegraphics[width=3cm]{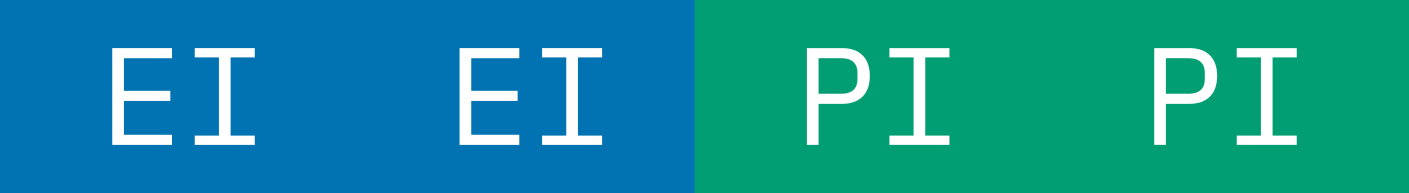}\\
         explore-exploit (\num{0.75}) & \includegraphics[width=3cm]{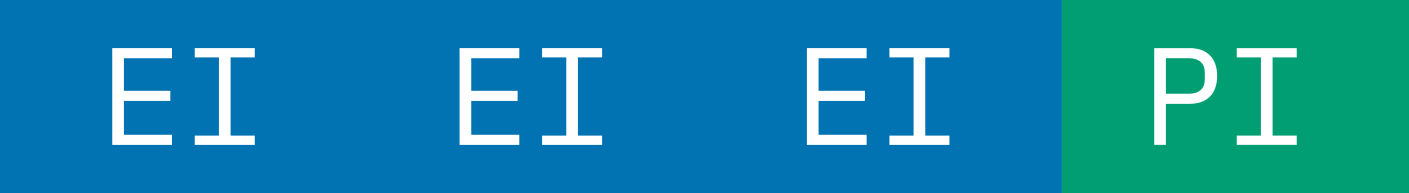}\\
         VBS \small{(Virtual Best Solver)} &
         \includegraphics[width=3cm]{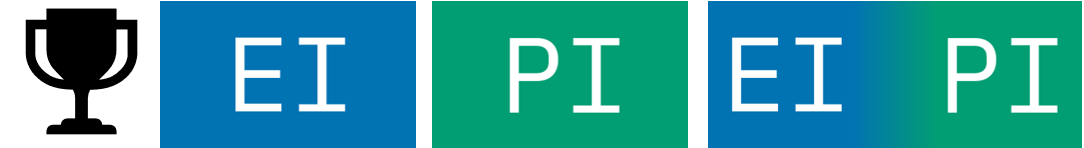}\\
         AFS \small{(\ac{AF} Selector)}&
         \includegraphics[width=3cm]{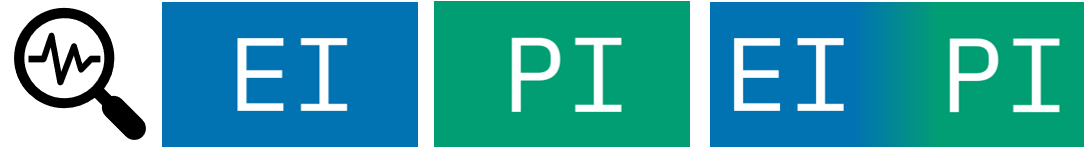}\\
         \bottomrule
    \end{tabular}
  \caption{Schedule portfolio of PI, EI and their combinations. AFS is a selector that aims to predict which of the first seven schedules is the best. VBS is a perfect selector and forms the ground truth for AFS.}%
  \label{tab:schedules}
\end{table}%

\section{Additional Results}
\label{sec:all_bbob_plots}
In this section we provide all boxplot and convergence figures for each BBOB function.
Please note that because the \ac{VBS} is selected based on final performance, it does not always show the best anytime-performance.

We first present a more in-depth discussion of the results on specific BBOB functions, focusing on the manually defined schedules.
A first observation is that switching from \ac{EI} to \ac{PI} is in general beneficial when the function landscape has an adequate global structure (F1-F19).
Here, the only exceptions are F5 and F19.
F5 is a purely linear function, where \ac{EI} performs best as it explores fast in the beginning and exploits sufficiently fast later on because of the simplicity of the landscape.
For F19, we hypothesize that only \ac{PI} is able to exploit  a lucky initial design landing close to optima.

In addition, we can see that \ac{PI} works well for uni-modal and quite smooth functions (F1-F14), while performing even better when used after the switch.
This observation is in line with \ac{PI}'s exploitative behavior.
In these cases exploiting does not miss any other optima further away in the landscape.

\begin{figure}[h]
    \centering
    \begin{subfigure}[b]{0.45\textwidth}
        \centering
        \includegraphics[width=\textwidth]{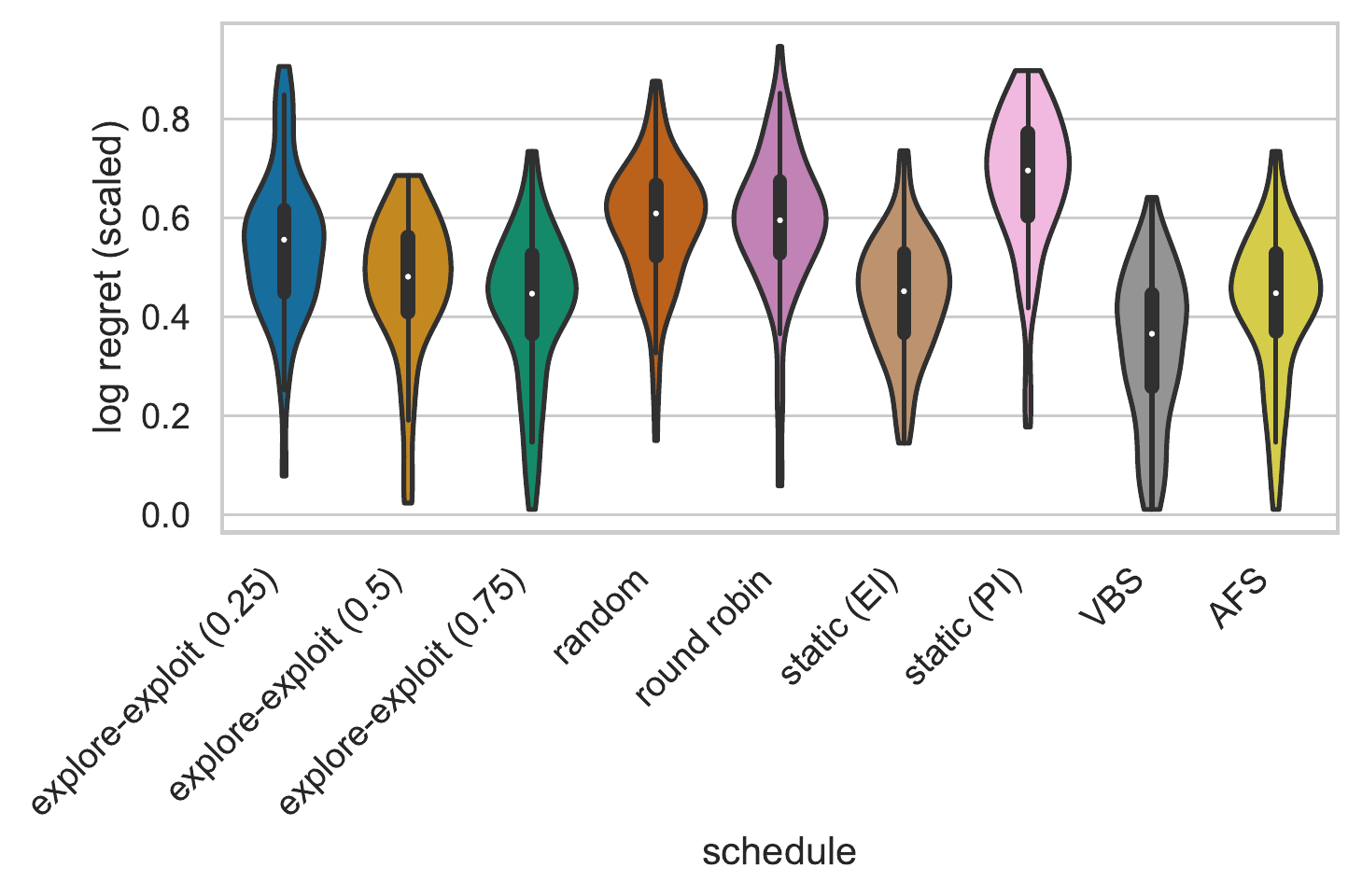}
        \caption{Final Log Regret (Scaled)}
        \label{subfig:boxplot_7}
    \end{subfigure}
    \hfill
    \begin{subfigure}[b]{0.45\textwidth}
        \centering
        \includegraphics[width=\textwidth]{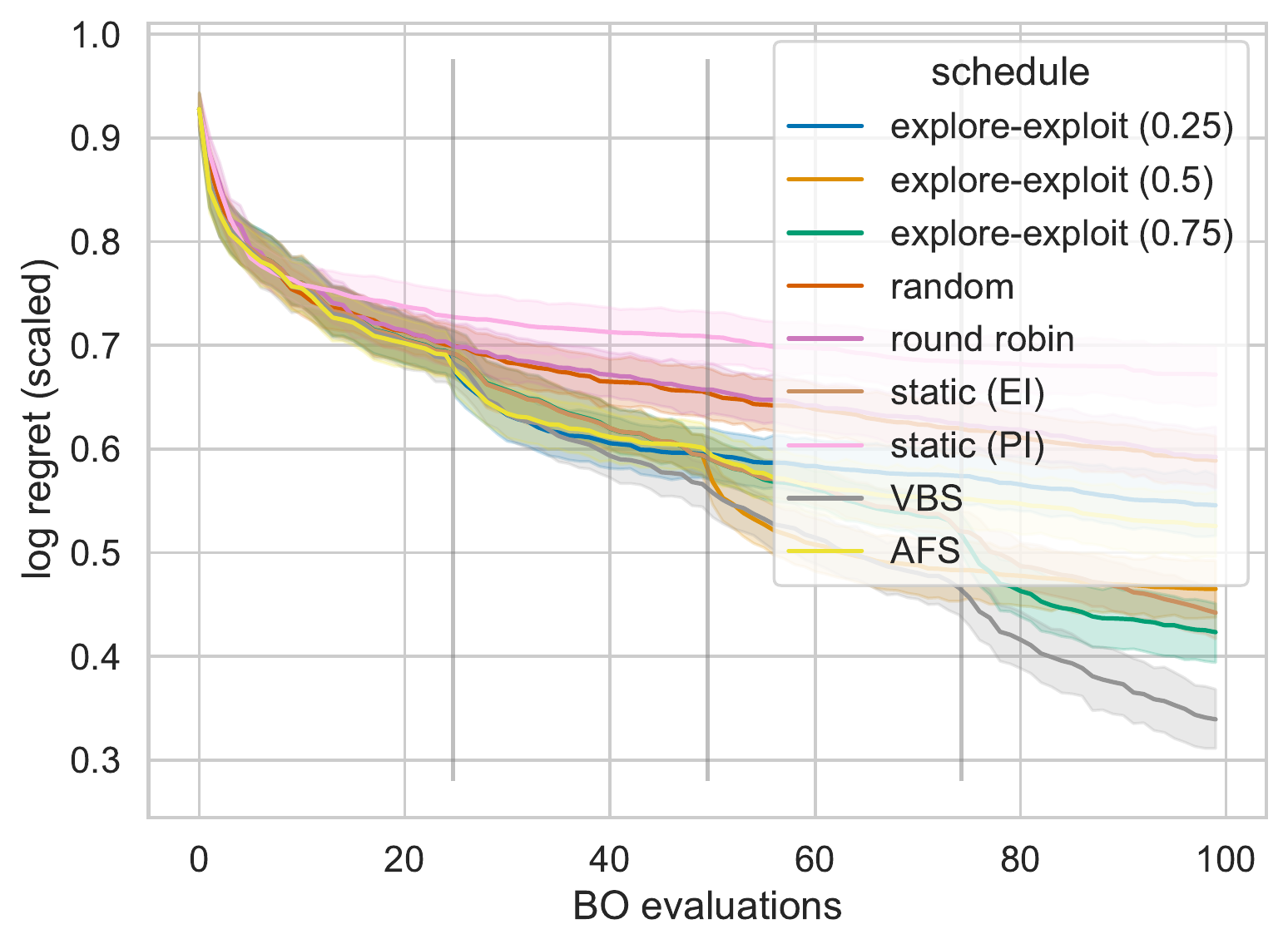}
        \caption{Log-Regret (Scaled) per Step}
        \label{subfig:convergence_7}
    \end{subfigure}\\
    \vspace*{3mm}
    \centering
    \begin{subfigure}[b]{\textwidth}
        \centering
        \includegraphics[width=\textwidth]{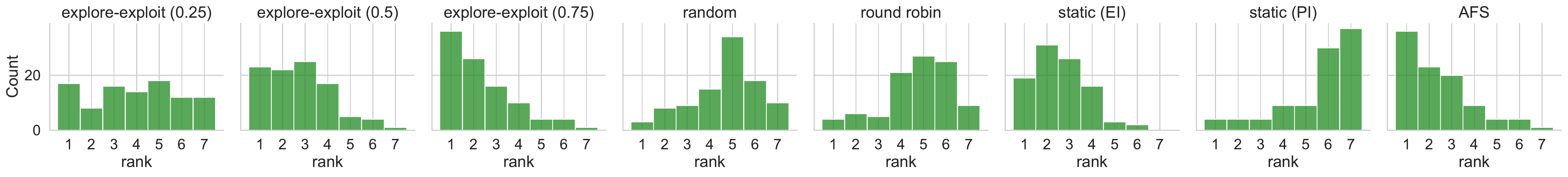}
        \caption{Rank}
        \label{subfig:rank_7}
    \end{subfigure}
    \caption{BBOB Function 7}
    \label{fig:bbob_function_7}
\end{figure}

\begin{figure}[h]
    \centering
    \begin{subfigure}[b]{0.45\textwidth}
        \centering
        \includegraphics[width=\textwidth]{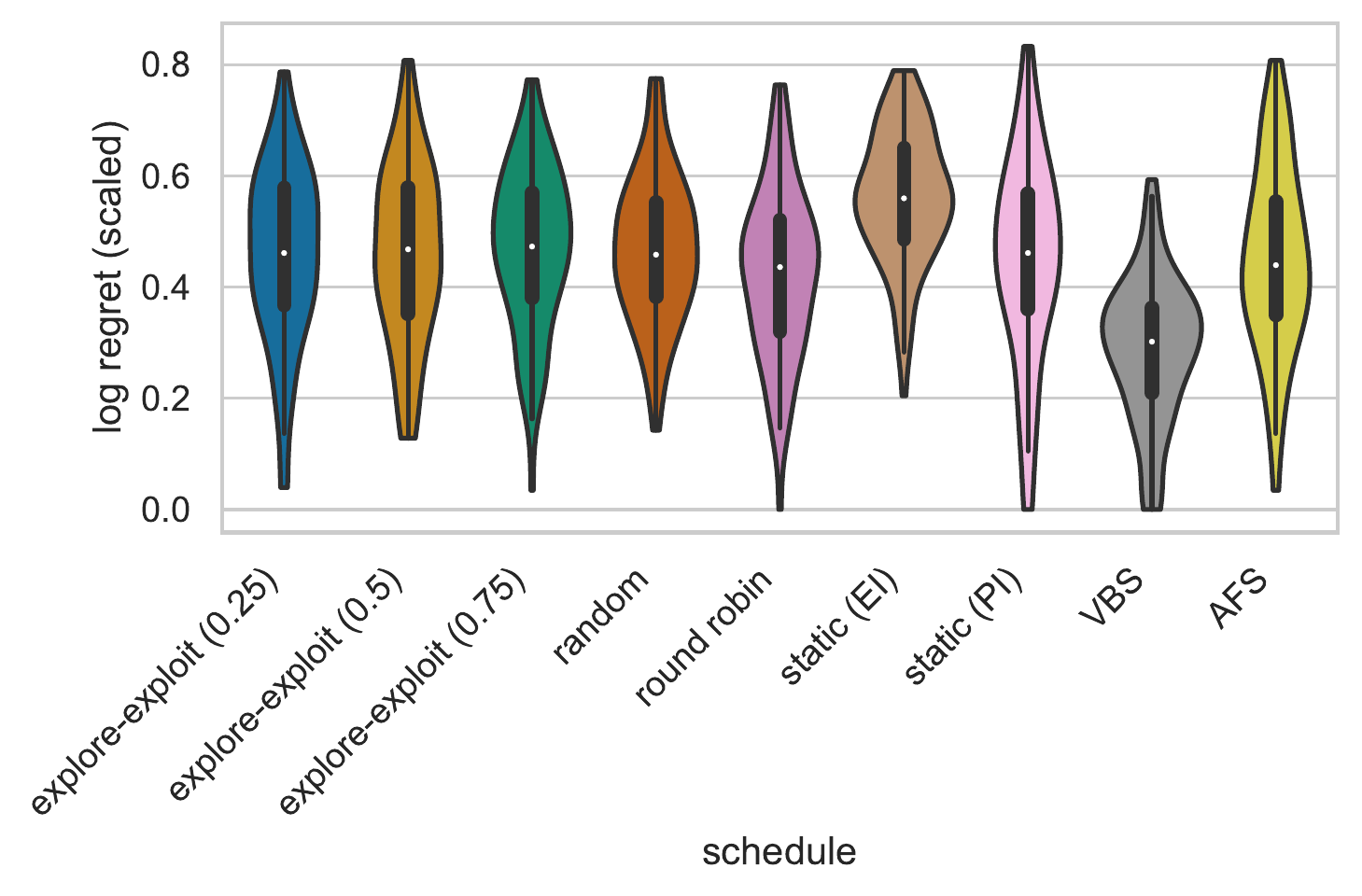}
        \caption{Final Log Regret (Scaled)}
        \label{subfig:boxplot_23}
    \end{subfigure}
    \hfill
    \begin{subfigure}[b]{0.45\textwidth}
        \centering
        \includegraphics[width=\textwidth]{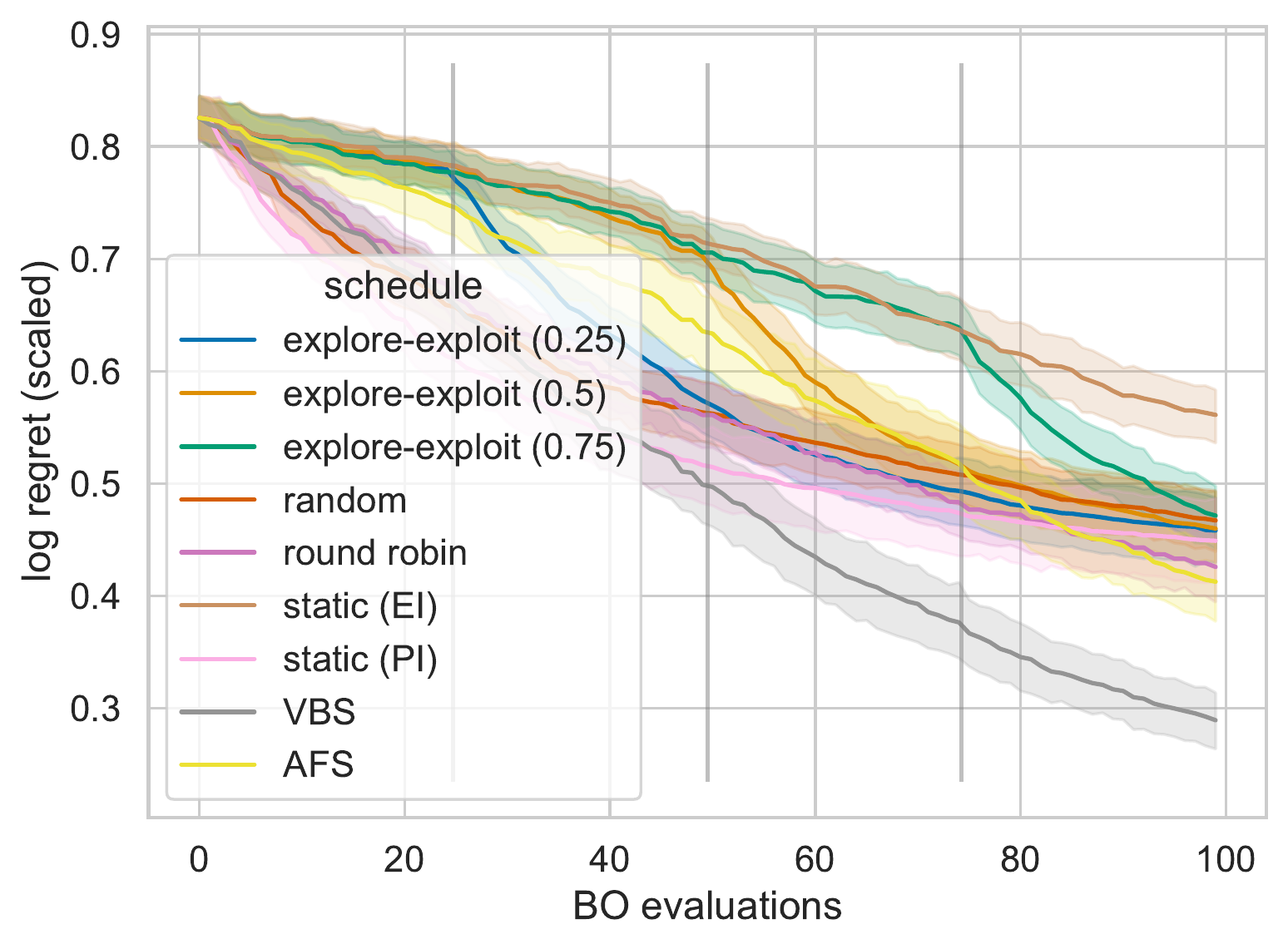}
        \caption{Log-Regret (Scaled) per Step}
        \label{subfig:convergence_23}
    \end{subfigure}\\
    \vspace*{3mm}
    \centering
    \begin{subfigure}[b]{\textwidth}
        \centering
        \includegraphics[width=\textwidth]{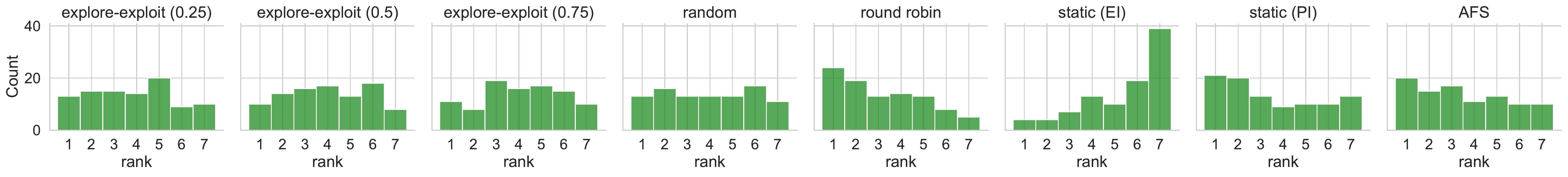}
        \caption{Rank}
        \label{subfig:rank_23}
    \end{subfigure}
    \caption{BBOB Function 23}
    \label{fig:bbob_function_23}
\end{figure}

In contrast, \ac{PI} is in general worse than \ac{EI} and also not beneficial after the switch for multi-modal functions with weak global structure (F20-F24).
Again, this is in line with our intuition because for these functions we have many important basins of attraction that have to be discovered before starting exploitation.
The only exception is F23 (\cref{fig:bbob_function_23}), a rugged function with a high number of global optima, where the probability of starting in a basin of attraction is high and thus exploitation is a viable strategy.
Also, the flatter the landscape the worse PI performs which can be also seen in F7 (slope with step, \cref{fig:bbob_function_7}).

Besides the static and the switching schedules, round robin switches from \ac{EI} to \ac{PI} and vice versa for each new function evaluation.
The round robin schedule creates a step-wise progress, but is never the best strategy.
Apparently, less frequent switching is preferable in order to take advantage of the strengths of each \ac{AF}.
On average, the random schedule performs similarly to round robin, but with a smoother progression.
Most likely, the random schedule still switches too frequently for the \ac{AF}s to work effectively.

\begin{figure}[h]
    \centering
    \begin{subfigure}[b]{0.45\textwidth}
        \centering
        \includegraphics[width=\textwidth]{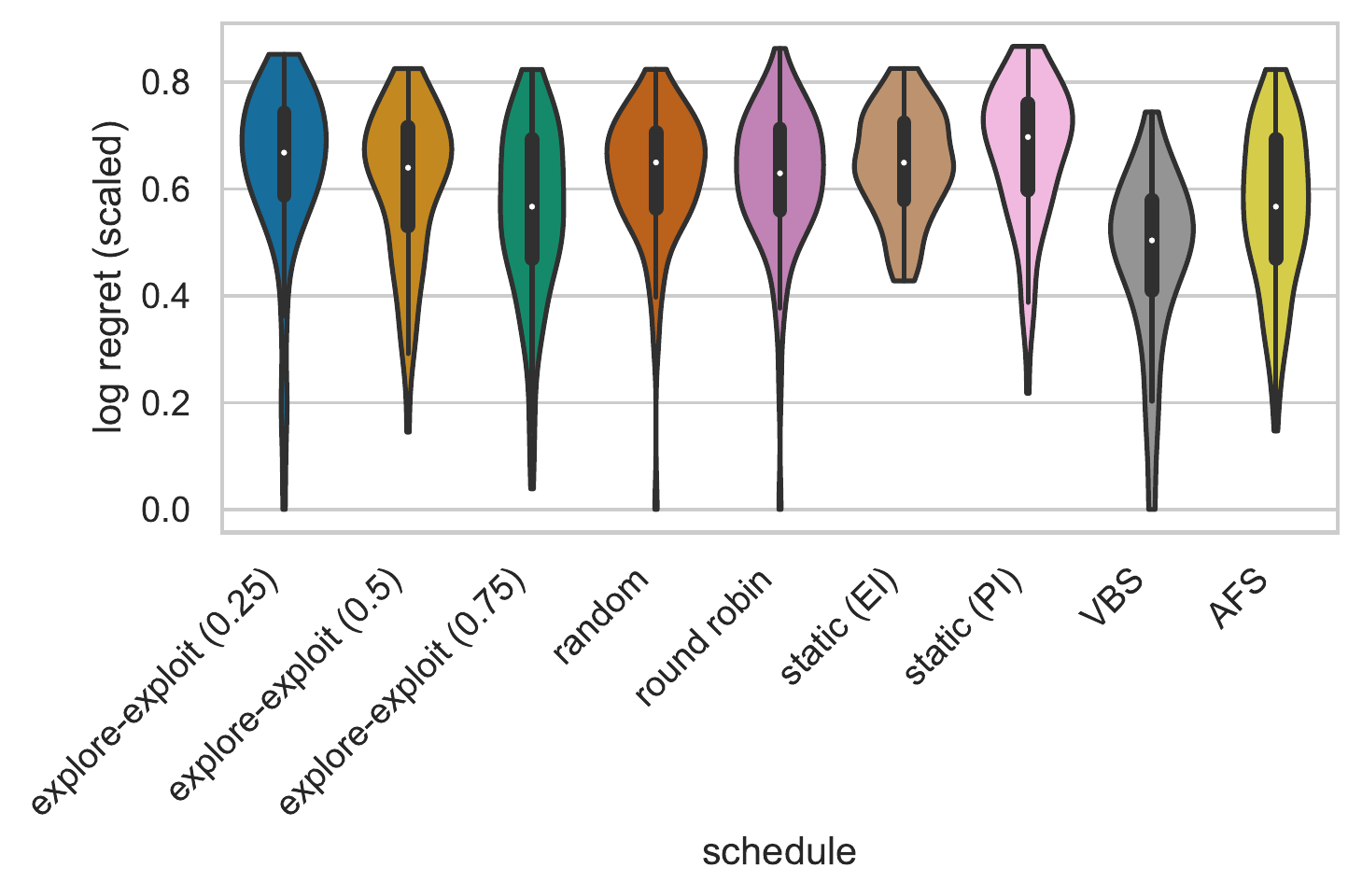}
        \caption{Final Log Regret (Scaled)}
        \label{subfig:boxplot_16}
    \end{subfigure}
    \hfill
    \begin{subfigure}[b]{0.45\textwidth}
        \centering
        \includegraphics[width=\textwidth]{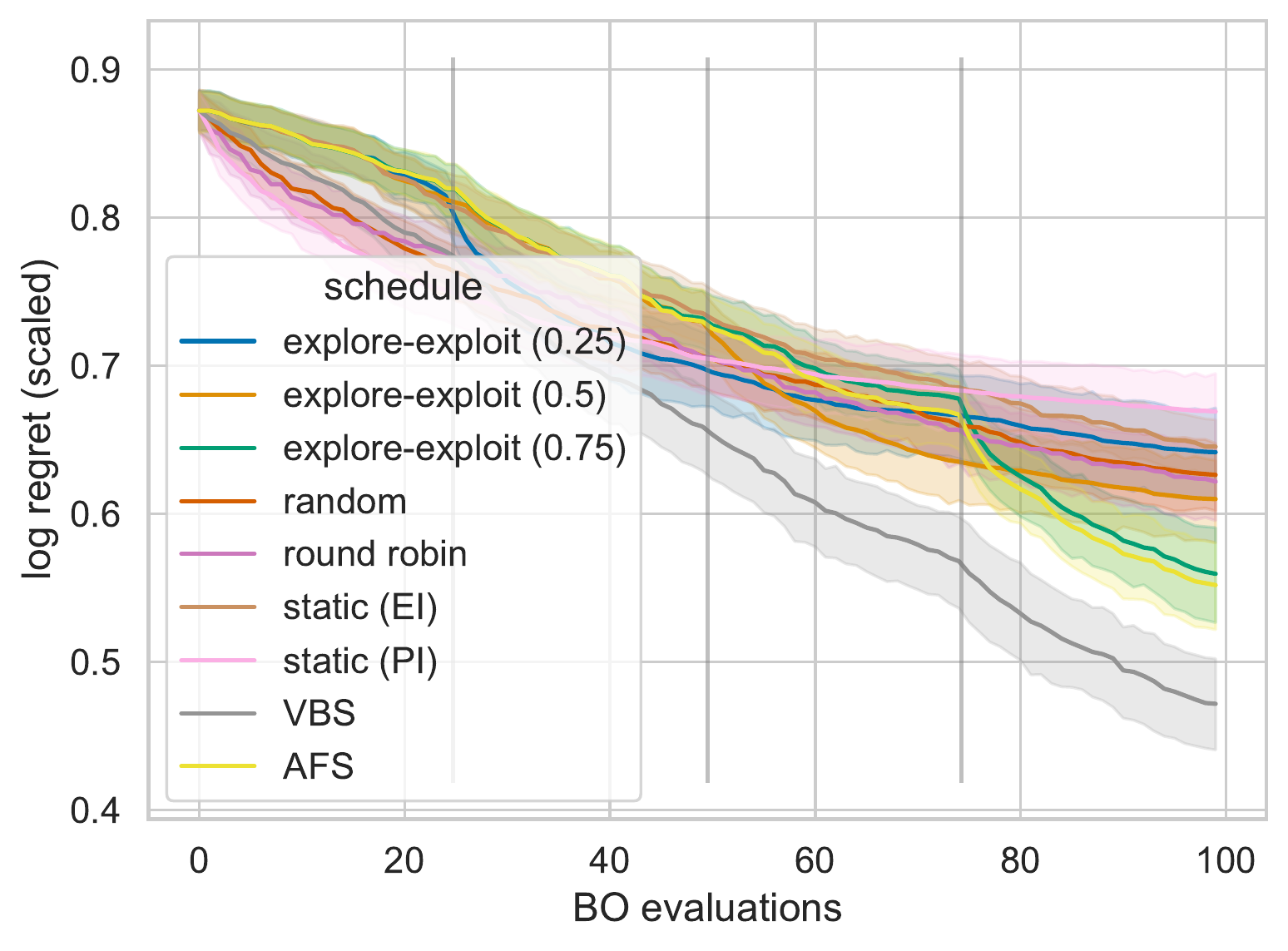}
        \caption{Log-Regret (Scaled) per Step}
        \label{subfig:convergence_16}
    \end{subfigure}\\
    \vspace*{3mm}
    \centering
    \begin{subfigure}[b]{\textwidth}
        \centering
        \includegraphics[width=\textwidth]{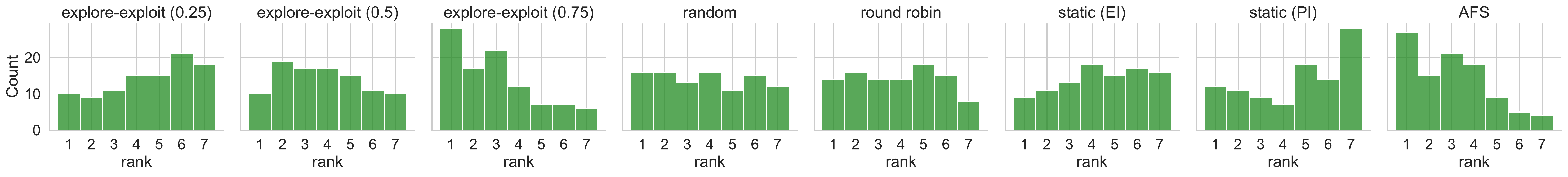}
        \caption{Rank}
        \label{subfig:rank_16}
    \end{subfigure}
    \caption{BBOB Function 16}
    \label{fig:bbob_function_16}
\end{figure}

On F16 we observe that a late switch from \ac{EI} to \ac{PI} performs best, see~\cref{subfig:convergence_16}.
We can also nicely spot the general boosts after switching after $\SI{25}{\percent}$, $\SI{50}{\percent}$ and $\SI{75}{\percent}$.
F16 has a highly rugged and moderately repetitive landscape with many local optima with evident quality difference.
Therefore, \ac{PI} can be trapped in a bad local optimum if applied too early.

\begin{figure}[h]
    \centering
    \begin{subfigure}[b]{0.45\textwidth}
        \centering
        \includegraphics[width=\textwidth]{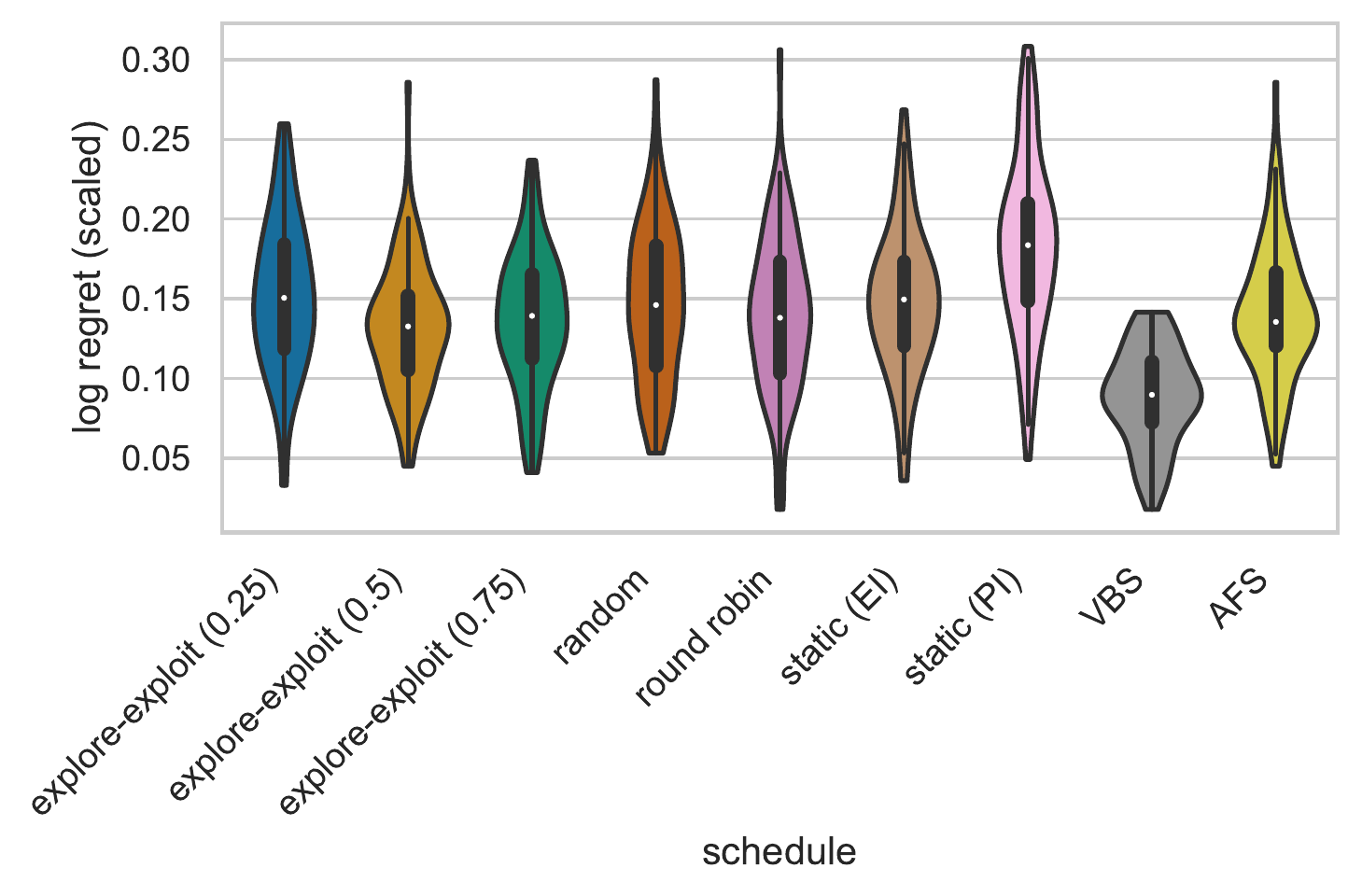}
        \caption{Final Log Regret (Scaled)}
        \label{subfig:boxplot_1}
    \end{subfigure}
    \hfill
    \begin{subfigure}[b]{0.45\textwidth}
        \centering
        \includegraphics[width=\textwidth]{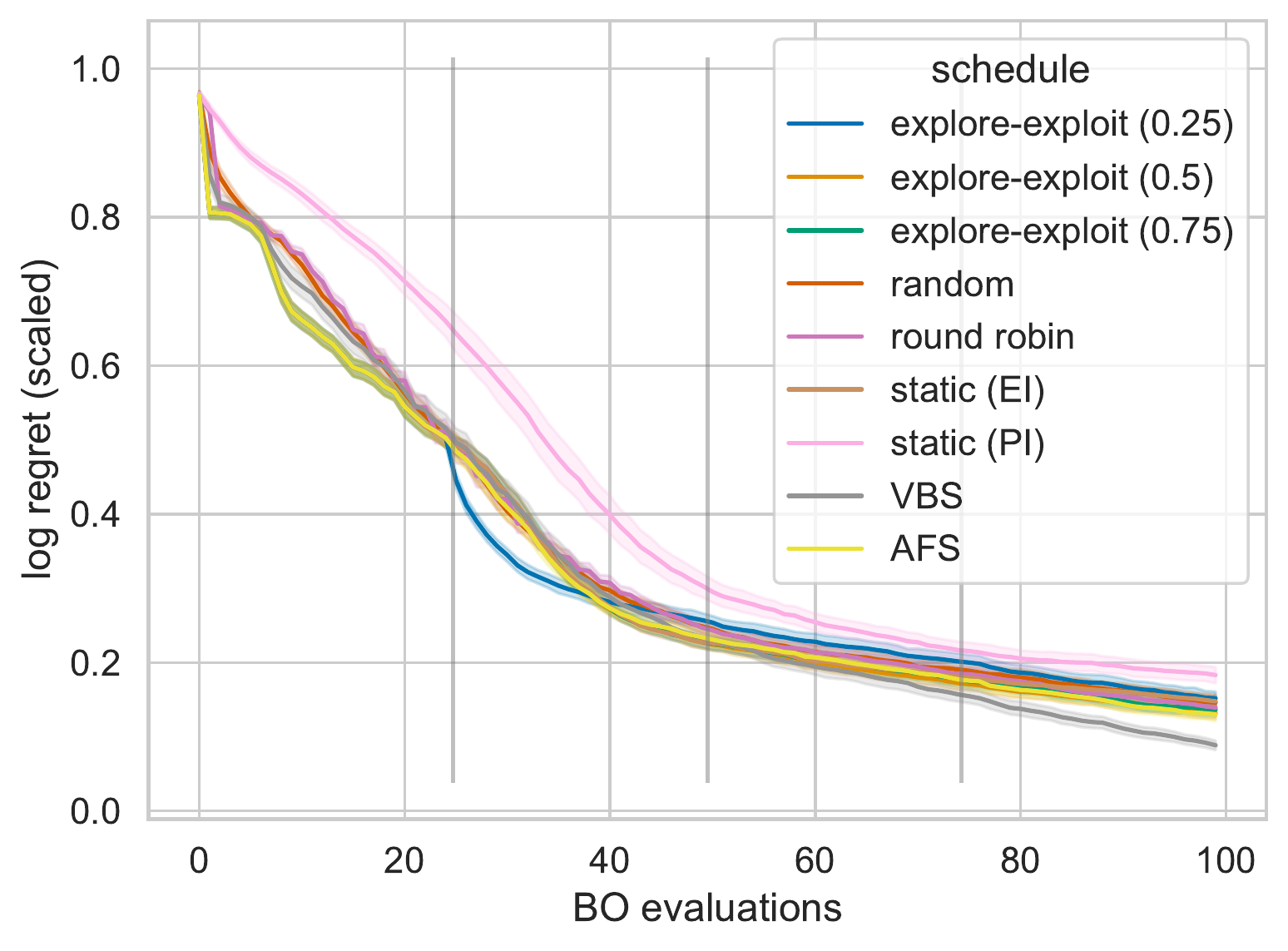}
        \caption{Log-Regret (Scaled) per Step}
        \label{subfig:convergence_1}
    \end{subfigure}\\
    \vspace*{3mm}
    \centering
    \begin{subfigure}[b]{\textwidth}
        \centering
        \includegraphics[width=\textwidth]{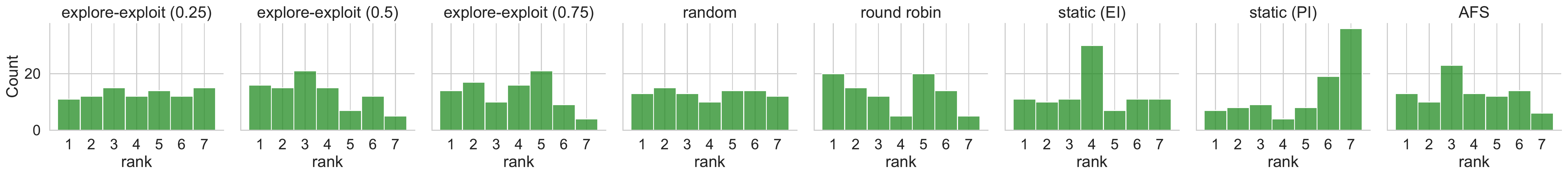}
        \caption{Rank}
        \label{subfig:rank_1}
    \end{subfigure}
    \caption{BBOB Function 1}
    \label{fig:bbob_function_1}
\end{figure}

\begin{figure}[h]
    \centering
    \begin{subfigure}[b]{0.45\textwidth}
        \centering
        \includegraphics[width=\textwidth]{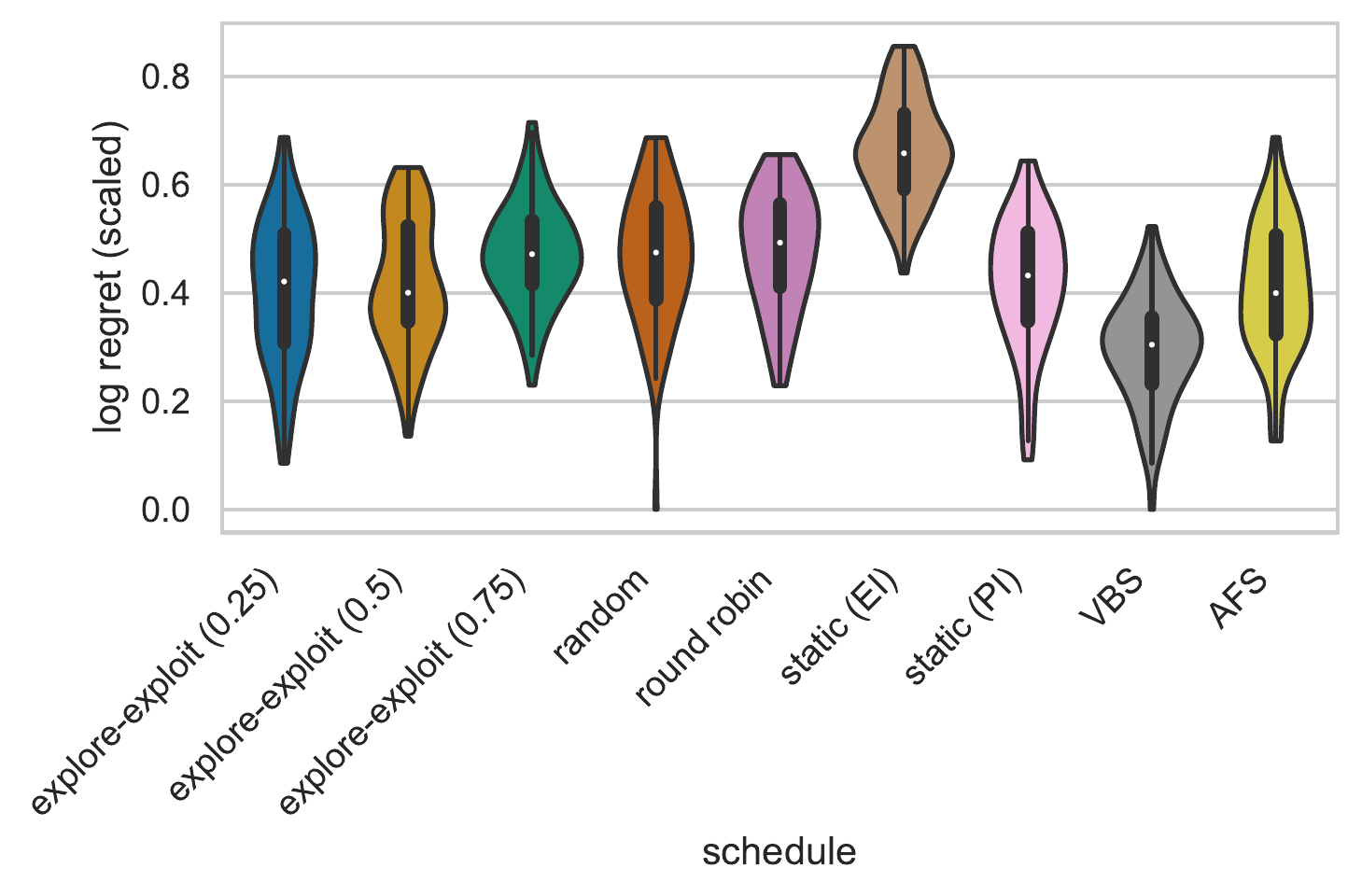}
        \caption{Final Log Regret (Scaled)}
        \label{subfig:boxplot_2}
    \end{subfigure}
    \hfill
    \begin{subfigure}[b]{0.45\textwidth}
        \centering
        \includegraphics[width=\textwidth]{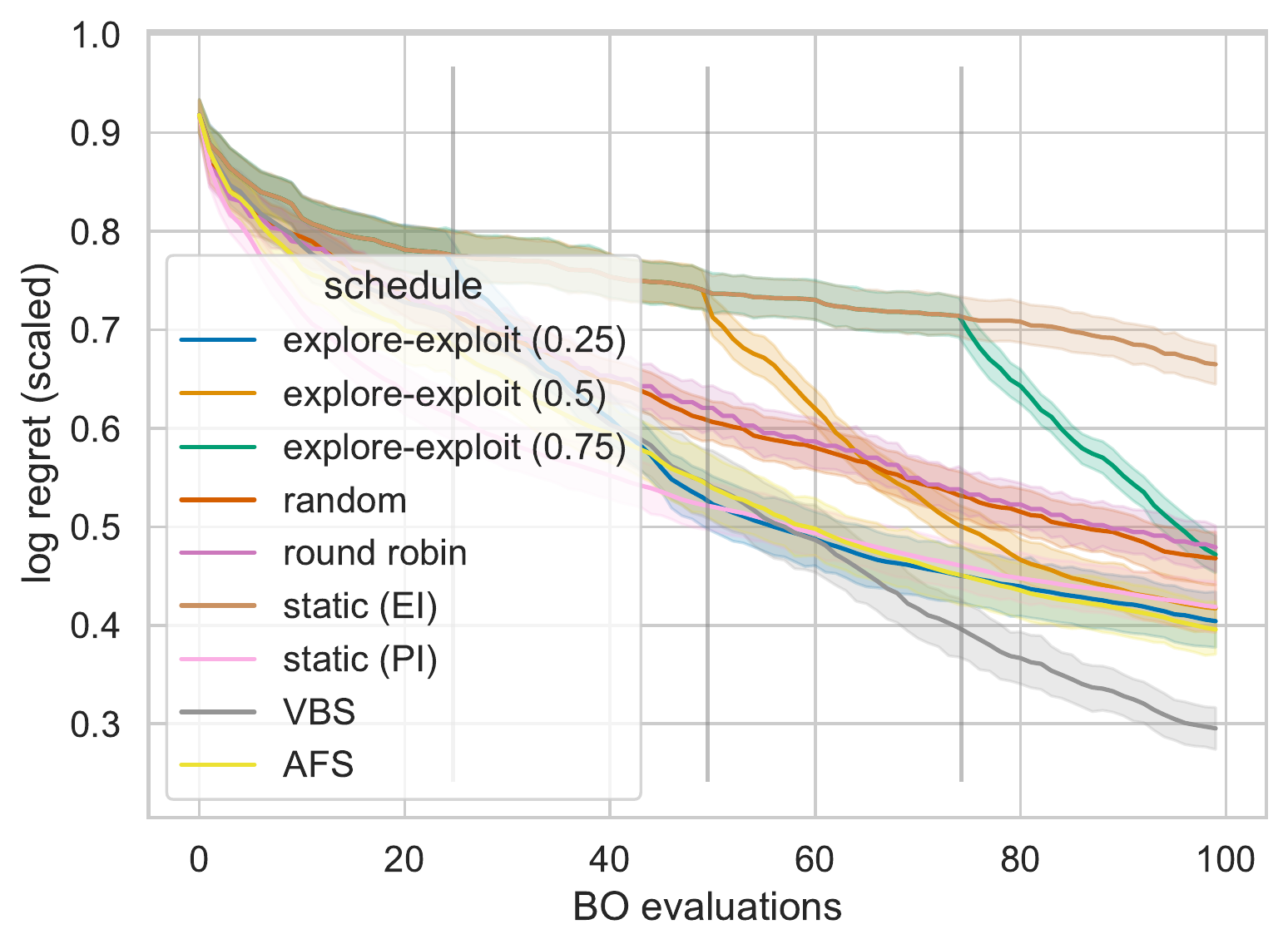}
        \caption{Log-Regret (Scaled) per Step}
        \label{subfig:convergence_2}
    \end{subfigure}\\
    \vspace*{3mm}
    \centering
    \begin{subfigure}[b]{\textwidth}
        \centering
        \includegraphics[width=\textwidth]{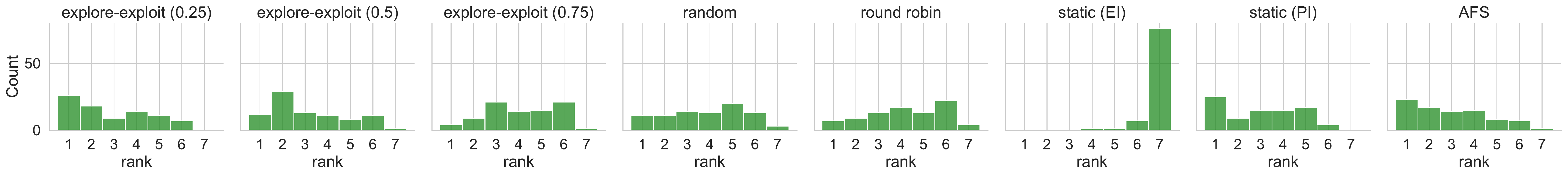}
        \caption{Rank}
        \label{subfig:rank_2}
    \end{subfigure}
    \caption{BBOB Function 2}
    \label{fig:bbob_function_2}
\end{figure}

\begin{figure}[h]
    \centering
    \begin{subfigure}[b]{0.45\textwidth}
        \centering
        \includegraphics[width=\textwidth]{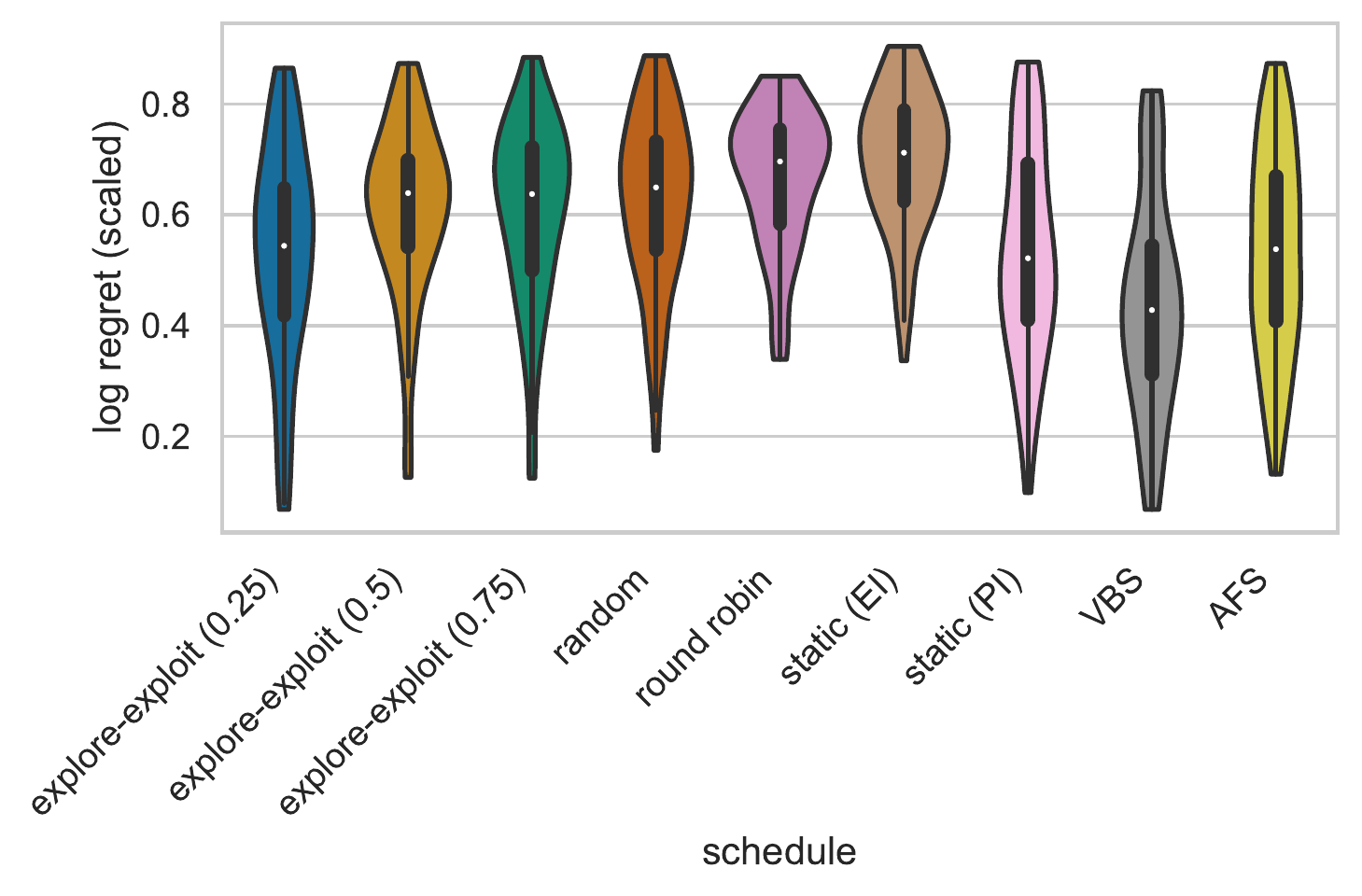}
        \caption{Final Log Regret (Scaled)}
        \label{subfig:boxplot_3}
    \end{subfigure}
    \hfill
    \begin{subfigure}[b]{0.45\textwidth}
        \centering
        \includegraphics[width=\textwidth]{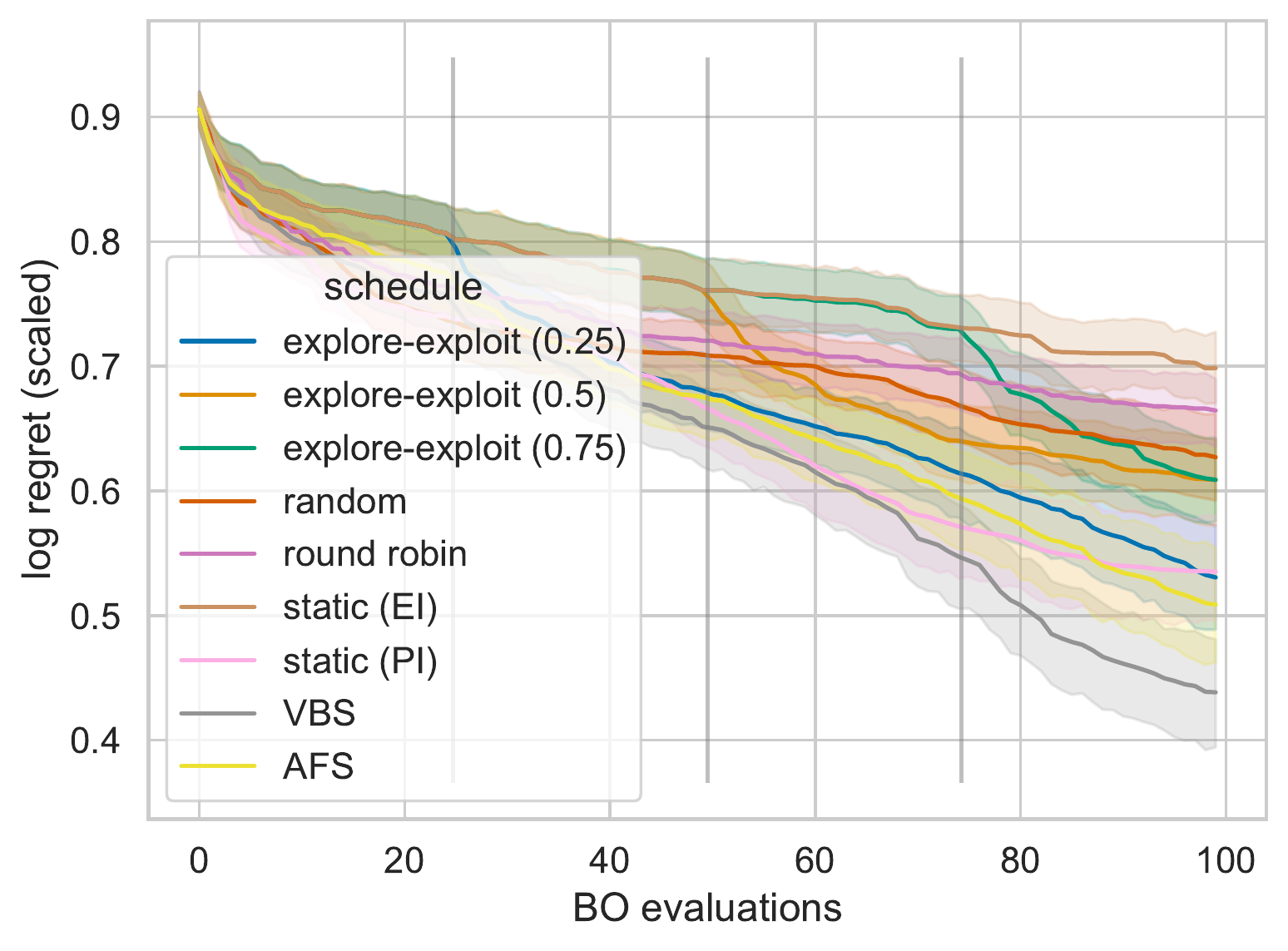}
        \caption{Log-Regret (Scaled) per Step}
        \label{subfig:convergence_3}
    \end{subfigure}\\
    \vspace*{3mm}
    \centering
    \begin{subfigure}[b]{\textwidth}
        \centering
        \includegraphics[width=\textwidth]{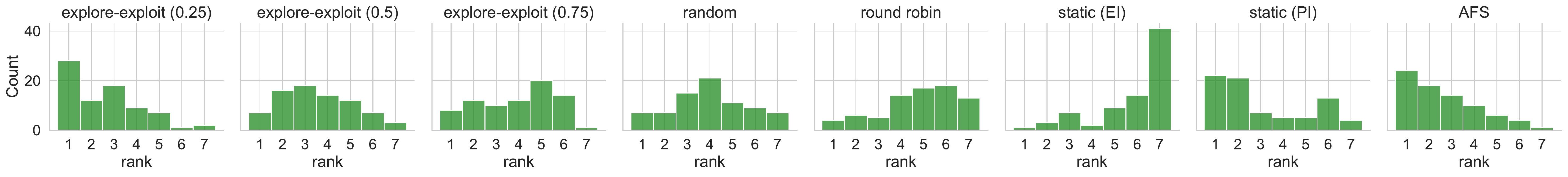}
        \caption{Rank}
        \label{subfig:rank_3}
    \end{subfigure}
    \caption{BBOB Function 3}
    \label{fig:bbob_function_3}
\end{figure}

\begin{figure}[h]
    \centering
    \begin{subfigure}[b]{0.45\textwidth}
        \centering
        \includegraphics[width=\textwidth]{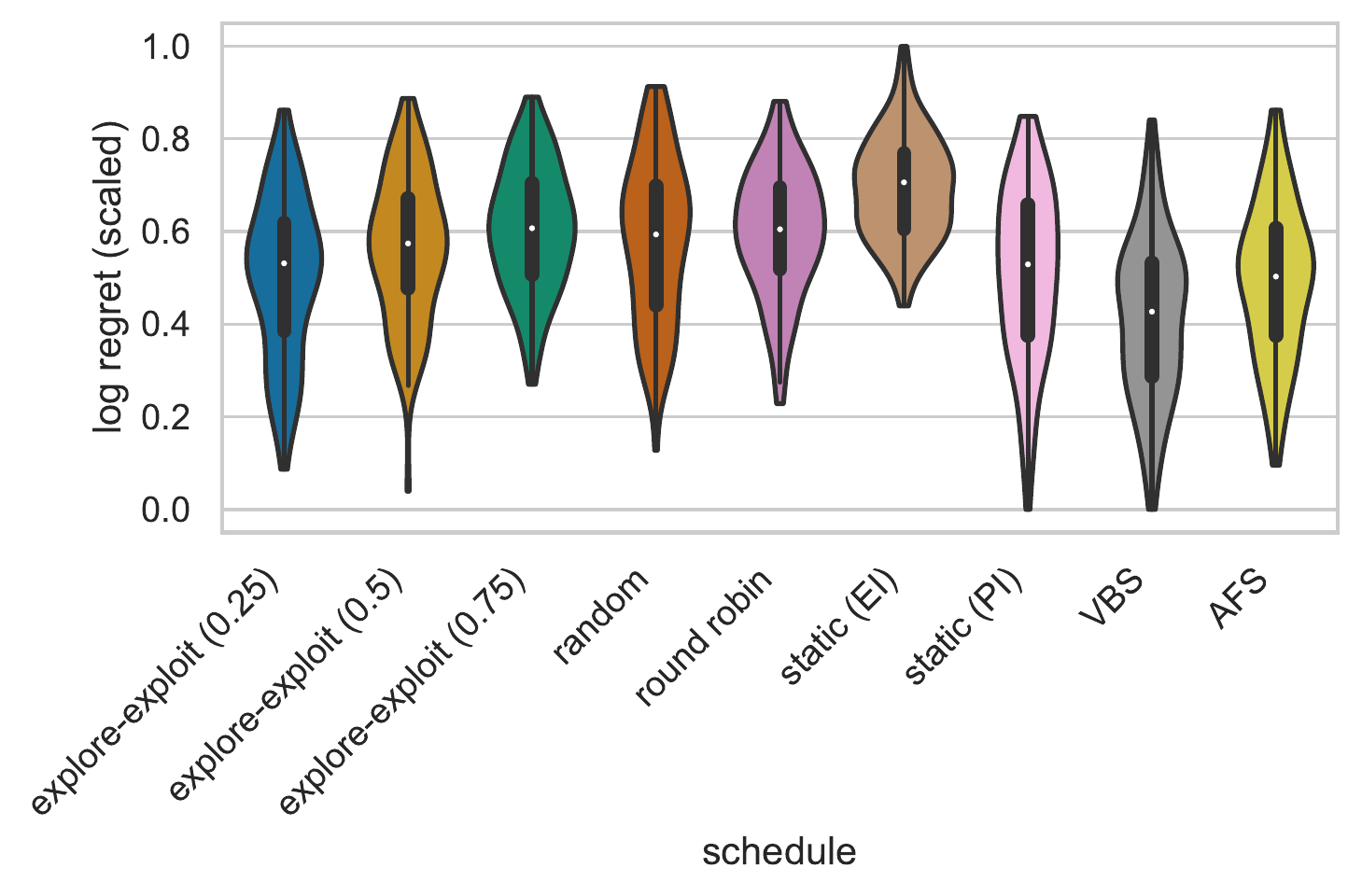}
        \caption{Final Log Regret (Scaled)}
        \label{subfig:boxplot_4}
    \end{subfigure}
    \hfill
    \begin{subfigure}[b]{0.45\textwidth}
        \centering
        \includegraphics[width=\textwidth]{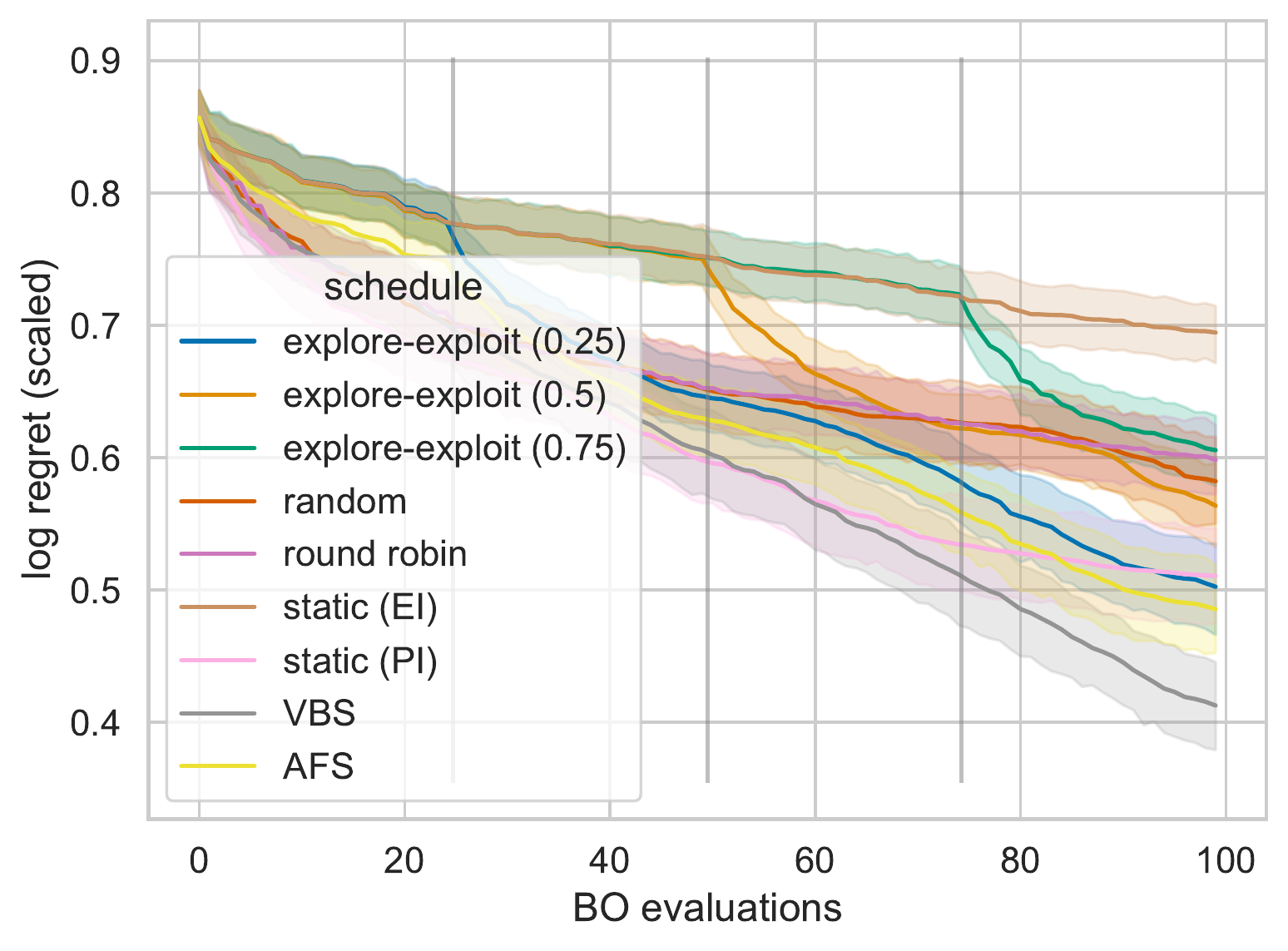}
        \caption{Log-Regret (Scaled) per Step}
        \label{subfig:convergence_4}
    \end{subfigure}\\
    \vspace*{3mm}
    \centering
    \begin{subfigure}[b]{\textwidth}
        \centering
        \includegraphics[width=\textwidth]{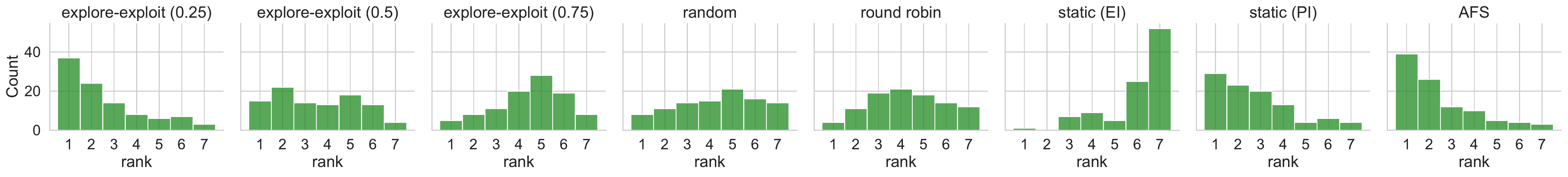}
        \caption{Rank}
        \label{subfig:rank_4}
    \end{subfigure}
    \caption{BBOB Function 4}
    \label{fig:bbob_function_4}
\end{figure}

\begin{figure}[h]
    \centering
    \begin{subfigure}[b]{0.45\textwidth}
        \centering
        \includegraphics[width=\textwidth]{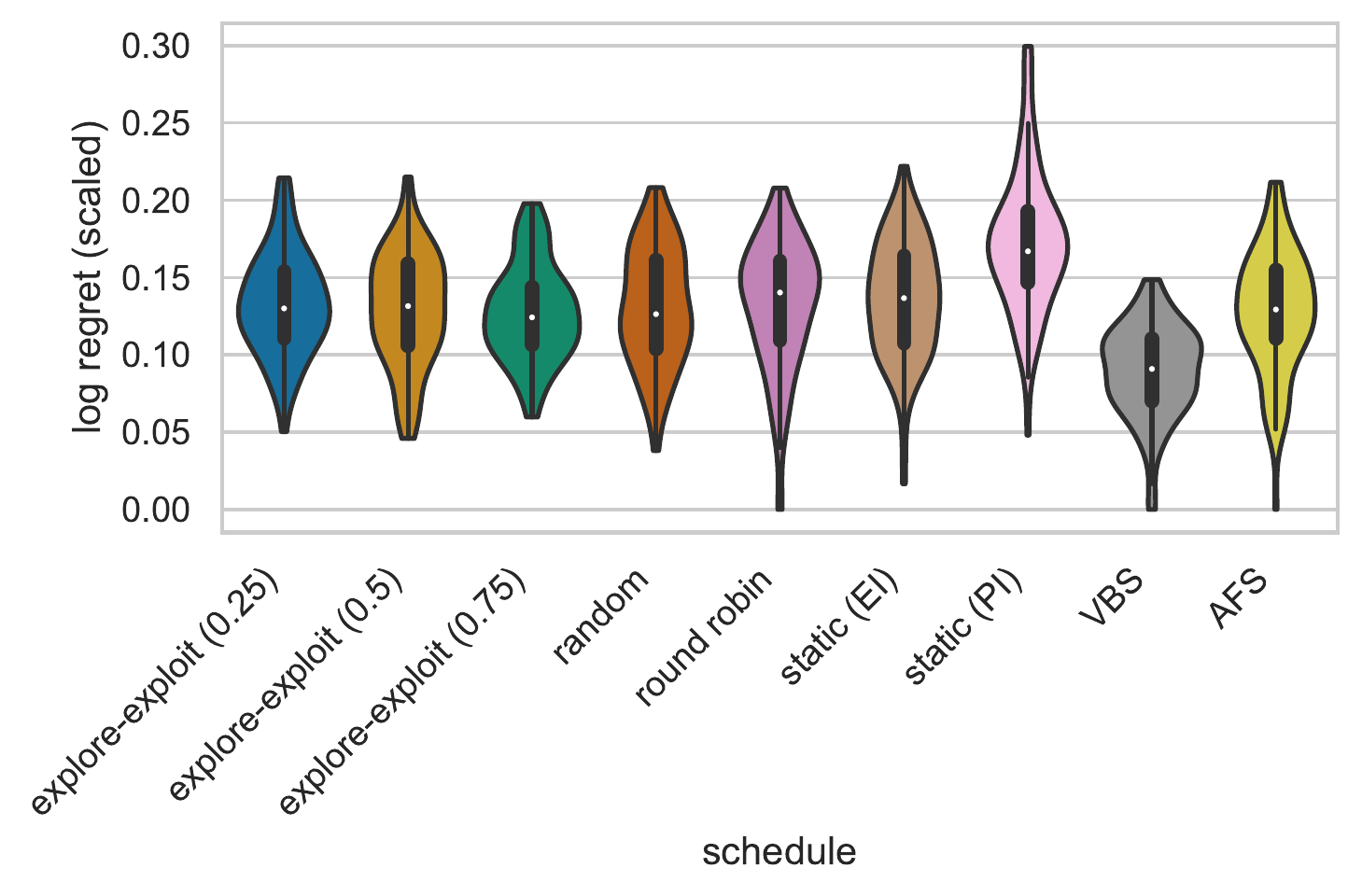}
        \caption{Final Log Regret (Scaled)}
        \label{subfig:boxplot_5}
    \end{subfigure}
    \hfill
    \begin{subfigure}[b]{0.45\textwidth}
        \centering
        \includegraphics[width=\textwidth]{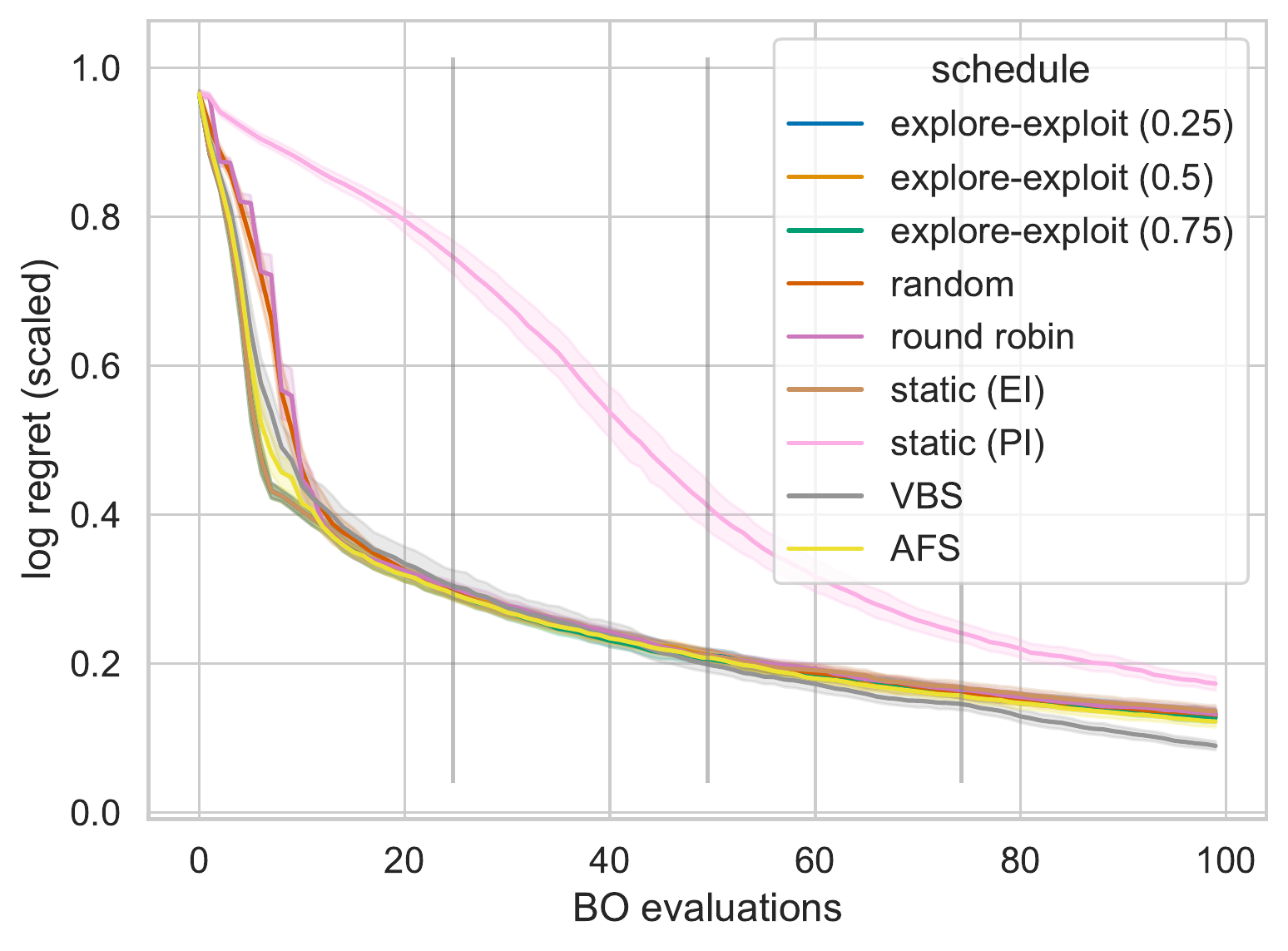}
        \caption{Log-Regret (Scaled) per Step}
        \label{subfig:convergence_5}
    \end{subfigure}\\
    \vspace*{3mm}
    \centering
    \begin{subfigure}[b]{\textwidth}
        \centering
        \includegraphics[width=\textwidth]{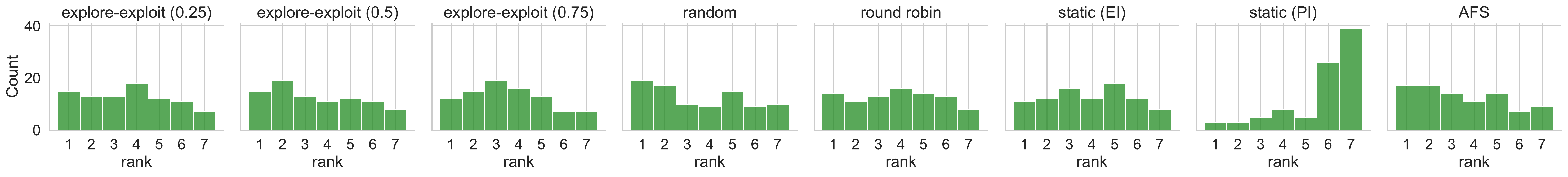}
        \caption{Rank}
        \label{subfig:rank_5}
    \end{subfigure}
    \caption{BBOB Function 5}
    \label{fig:bbob_function_5}
\end{figure}

\begin{figure}[h]
    \centering
    \begin{subfigure}[b]{0.45\textwidth}
        \centering
        \includegraphics[width=\textwidth]{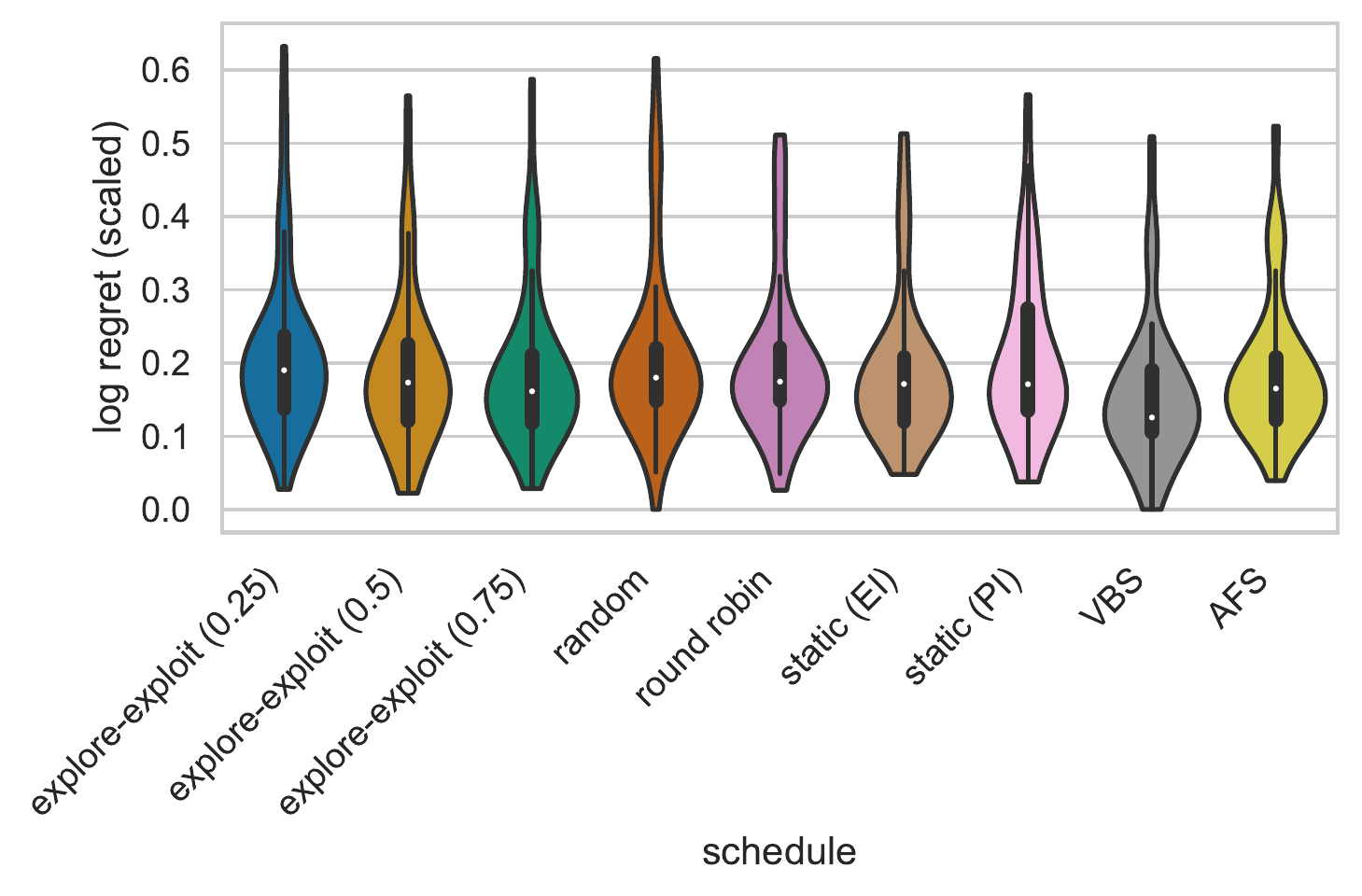}
        \caption{Final Log Regret (Scaled)}
        \label{subfig:boxplot_6}
    \end{subfigure}
    \hfill
    \begin{subfigure}[b]{0.45\textwidth}
        \centering
        \includegraphics[width=\textwidth]{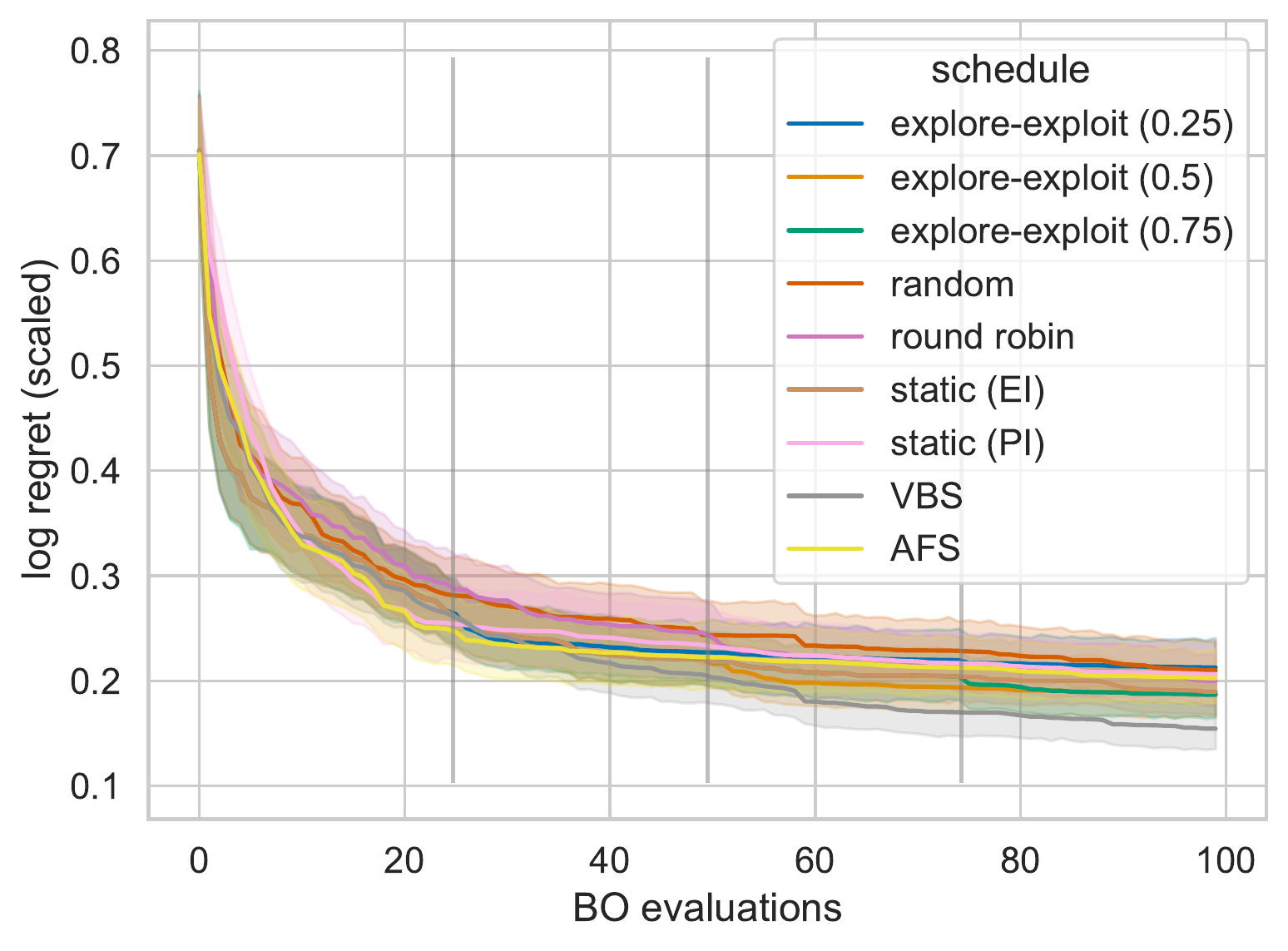}
        \caption{Log-Regret (Scaled) per Step}
        \label{subfig:convergence_6}
    \end{subfigure}\\
    \vspace*{3mm}
    \centering
    \begin{subfigure}[b]{\textwidth}
        \centering
        \includegraphics[width=\textwidth]{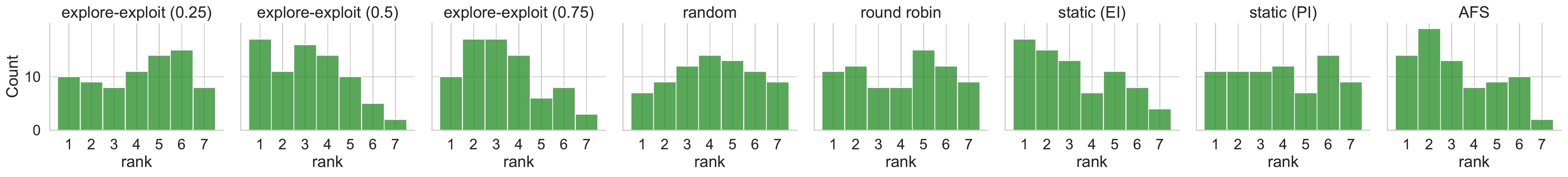}
        \caption{Rank}
        \label{subfig:rank_6}
    \end{subfigure}
    \caption{BBOB Function 6}
    \label{fig:bbob_function_6}
\end{figure}

\begin{figure}[h]
    \centering
    \begin{subfigure}[b]{0.45\textwidth}
        \centering
        \includegraphics[width=\textwidth]{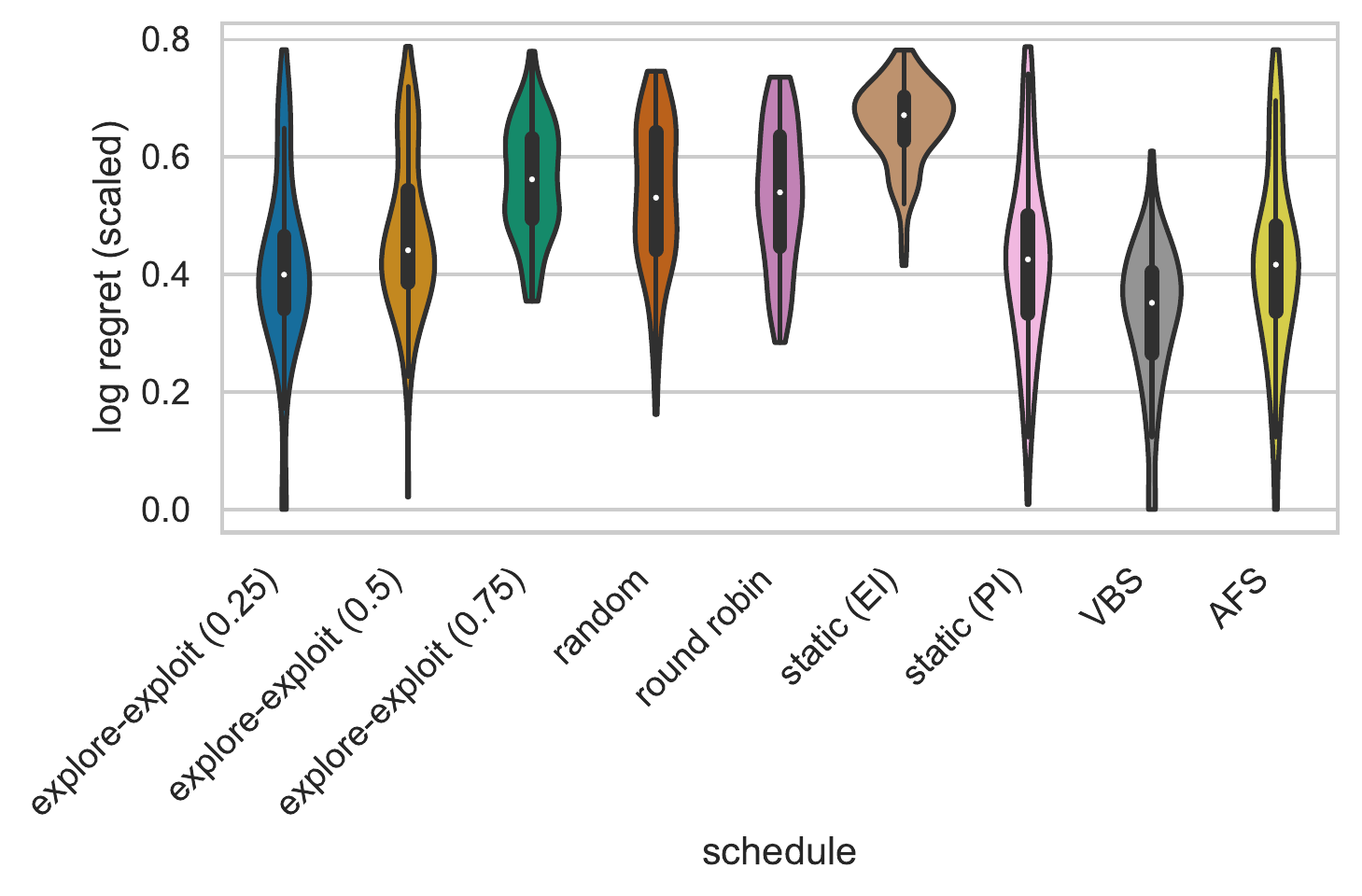}
        \caption{Final Log Regret (Scaled)}
        \label{subfig:boxplot_8}
    \end{subfigure}
    \hfill
    \begin{subfigure}[b]{0.45\textwidth}
        \centering
        \includegraphics[width=\textwidth]{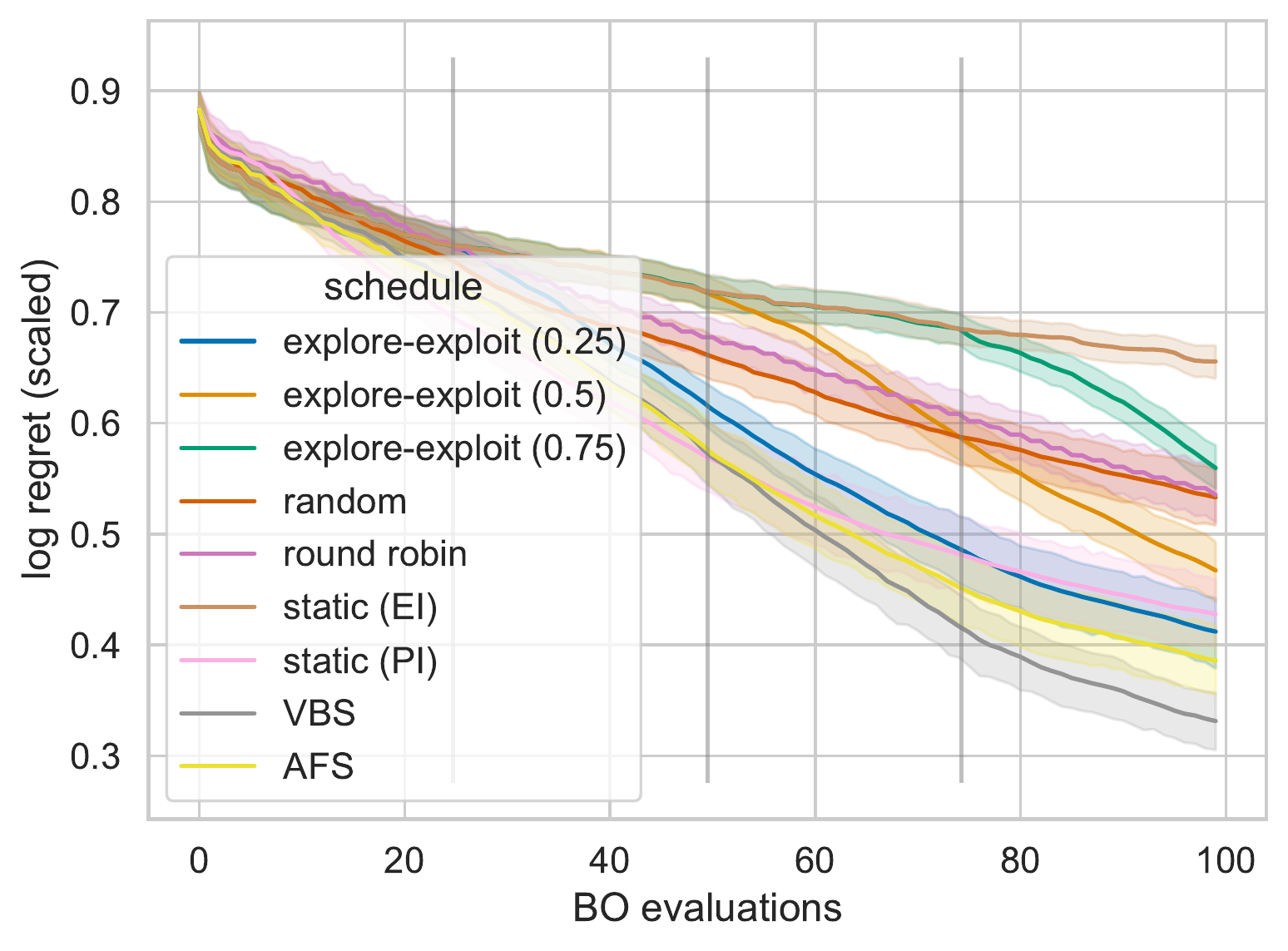}
        \caption{Log-Regret (Scaled) per Step}
        \label{subfig:convergence_8}
    \end{subfigure}\\
    \vspace*{3mm}
    \centering
    \begin{subfigure}[b]{\textwidth}
        \centering
        \includegraphics[width=\textwidth]{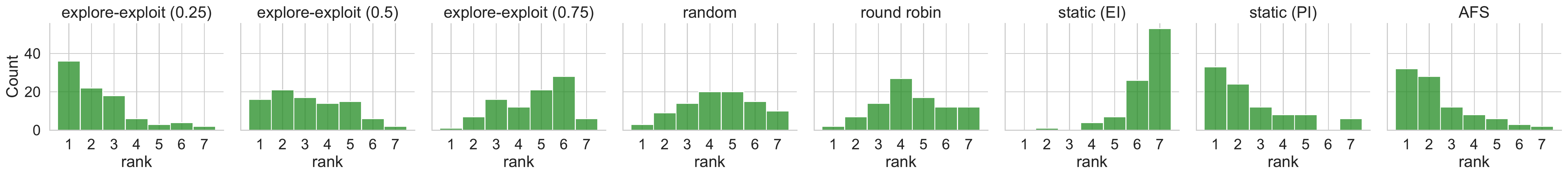}
        \caption{Rank}
        \label{subfig:rank_8}
    \end{subfigure}
    \caption{BBOB Function 8}
    \label{fig:bbob_function_8}
\end{figure}

\begin{figure}[h]
    \centering
    \begin{subfigure}[b]{0.45\textwidth}
        \centering
        \includegraphics[width=\textwidth]{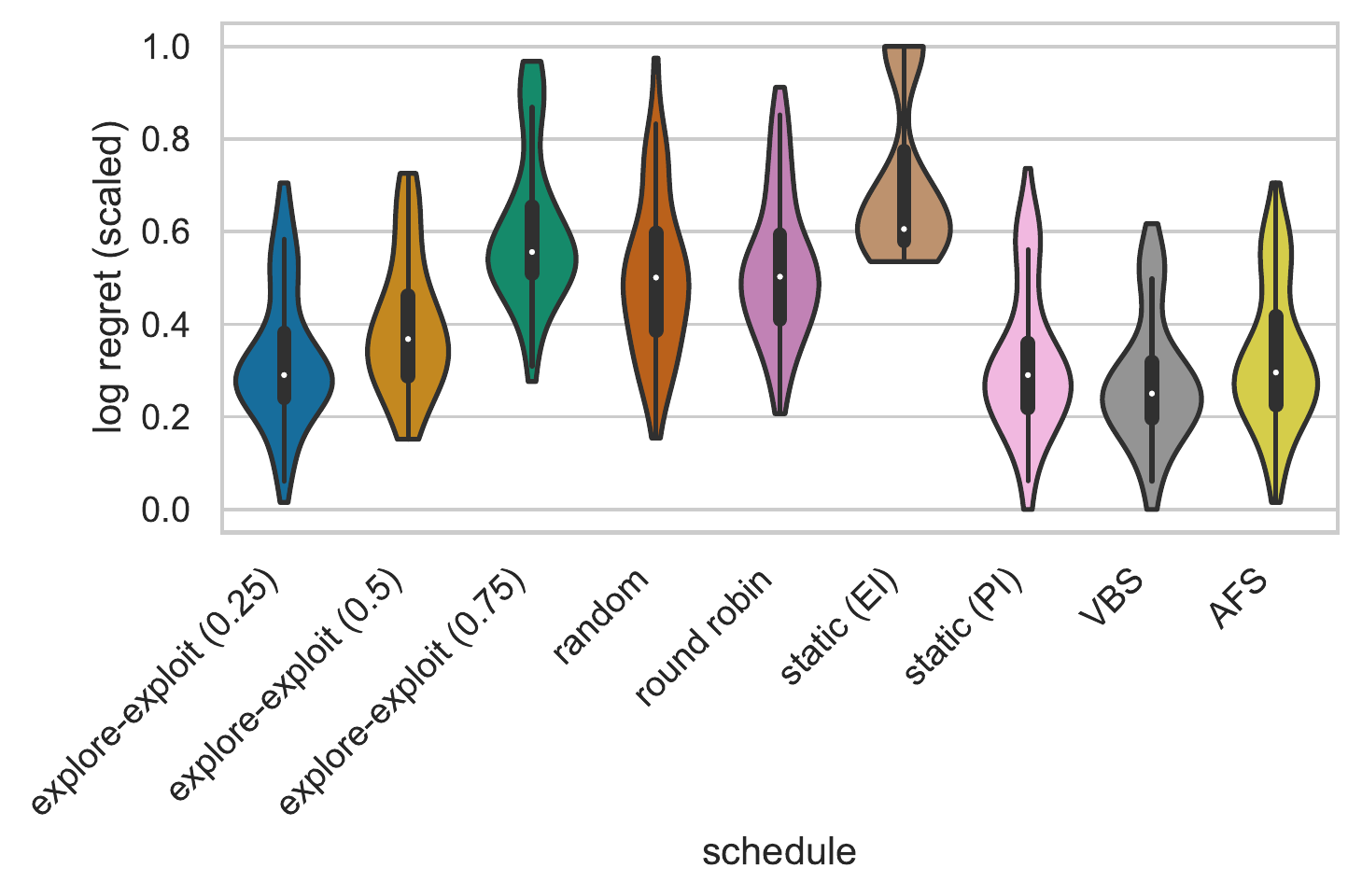}
        \caption{Final Log Regret (Scaled)}
        \label{subfig:boxplot_9}
    \end{subfigure}
    \hfill
    \begin{subfigure}[b]{0.45\textwidth}
        \centering
        \includegraphics[width=\textwidth]{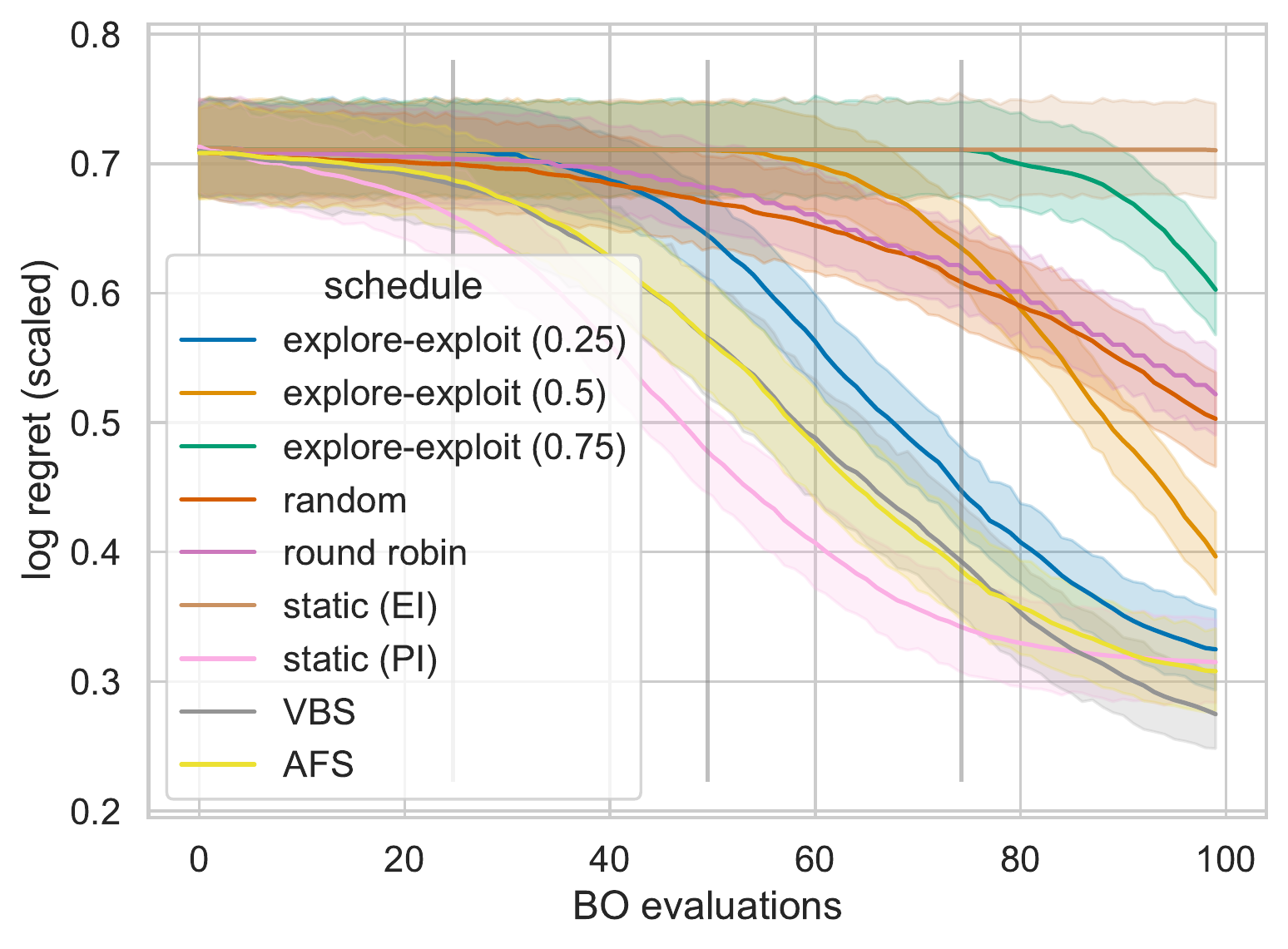}
        \caption{Log-Regret (Scaled) per Step}
        \label{subfig:convergence_9}
    \end{subfigure}\\
    \vspace*{3mm}
    \centering
    \begin{subfigure}[b]{\textwidth}
        \centering
        \includegraphics[width=\textwidth]{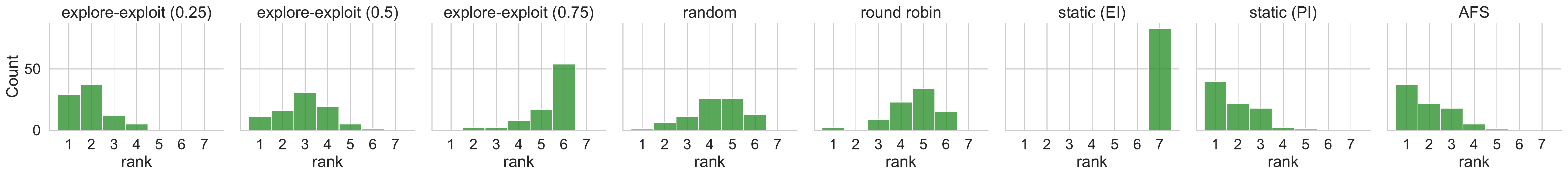}
        \caption{Rank}
        \label{subfig:rank_9}
    \end{subfigure}
    \caption{BBOB Function 9}
    \label{fig:bbob_function_9}
\end{figure}

\begin{figure}[h]
    \centering
    \begin{subfigure}[b]{0.45\textwidth}
        \centering
        \includegraphics[width=\textwidth]{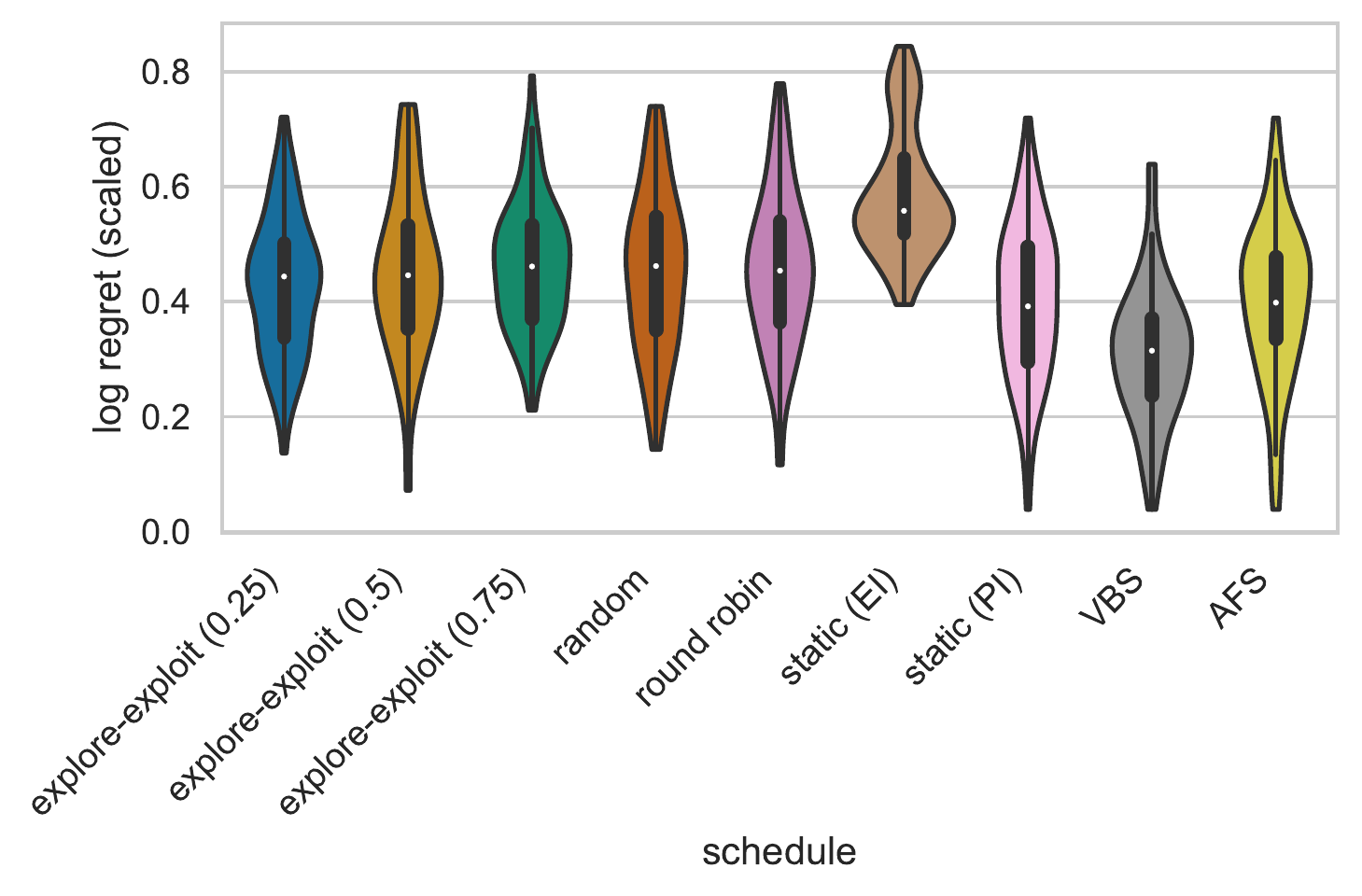}
        \caption{Final Log Regret (Scaled)}
        \label{subfig:boxplot_10}
    \end{subfigure}
    \hfill
    \begin{subfigure}[b]{0.45\textwidth}
        \centering
        \includegraphics[width=\textwidth]{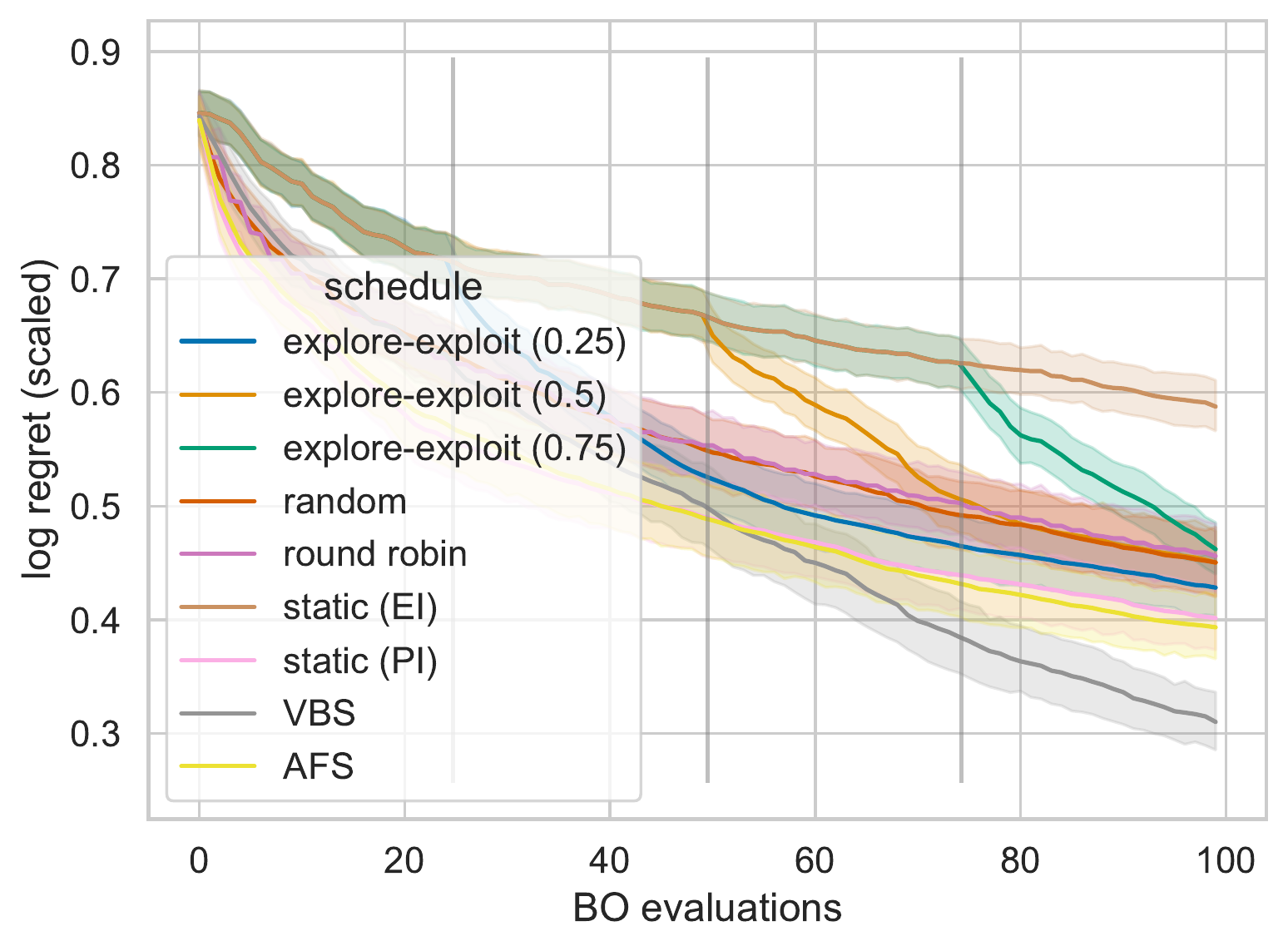}
        \caption{Log-Regret (Scaled) per Step}
        \label{subfig:convergence_10}
    \end{subfigure}\\
    \vspace*{3mm}
    \centering
    \begin{subfigure}[b]{\textwidth}
        \centering
        \includegraphics[width=\textwidth]{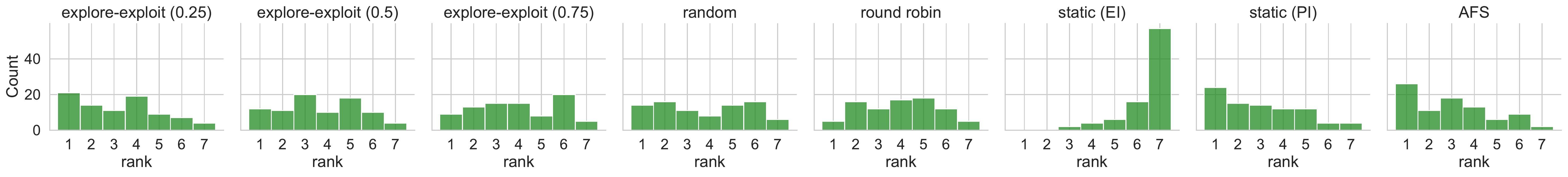}
        \caption{Rank}
        \label{subfig:rank_10}
    \end{subfigure}
    \caption{BBOB Function 10}
    \label{fig:bbob_function_10}
\end{figure}

\begin{figure}[h]
    \centering
    \begin{subfigure}[b]{0.45\textwidth}
        \centering
        \includegraphics[width=\textwidth]{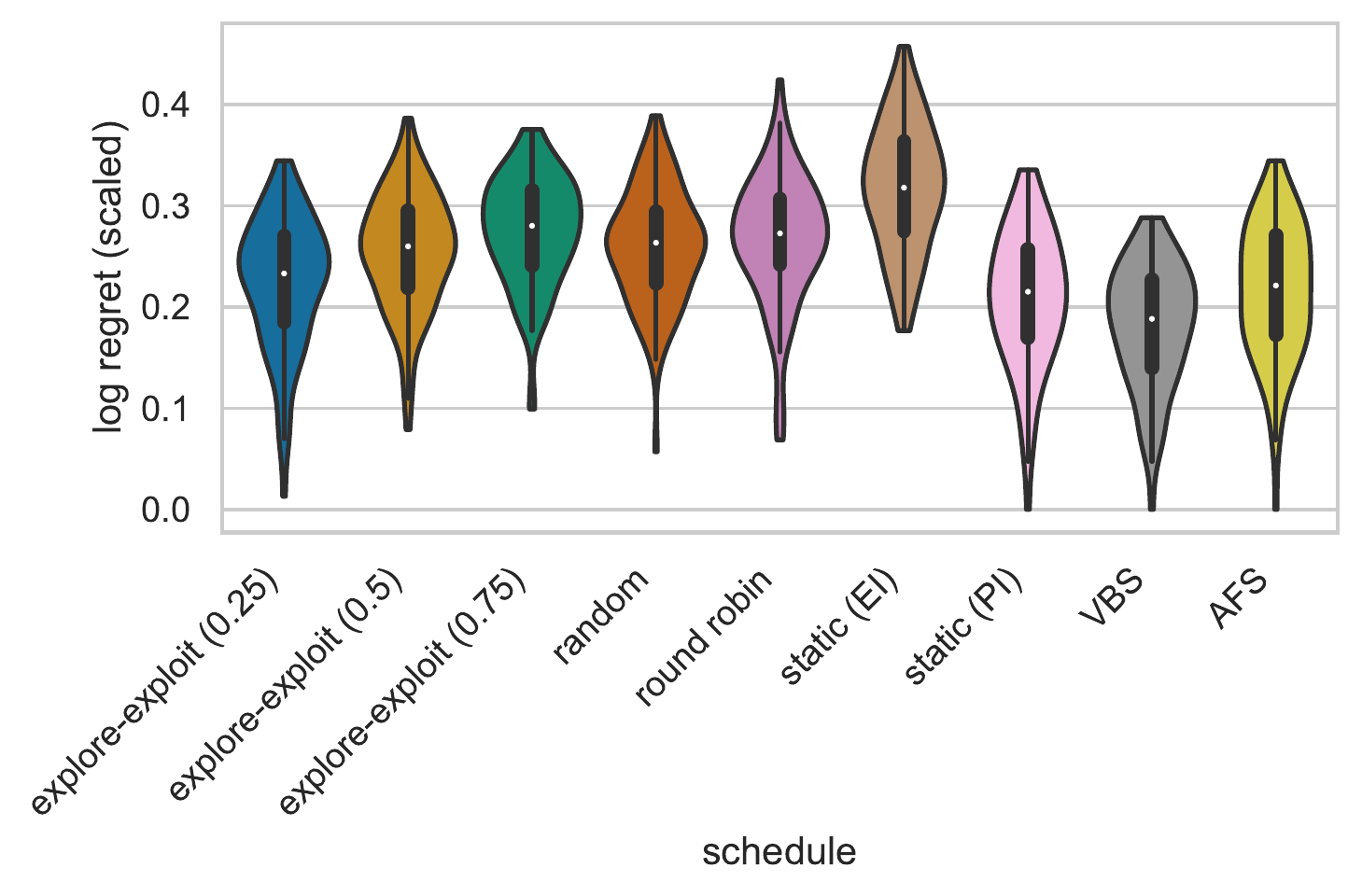}
        \caption{Final Log Regret (Scaled)}
        \label{subfig:boxplot_11}
    \end{subfigure}
    \hfill
    \begin{subfigure}[b]{0.45\textwidth}
        \centering
        \includegraphics[width=\textwidth]{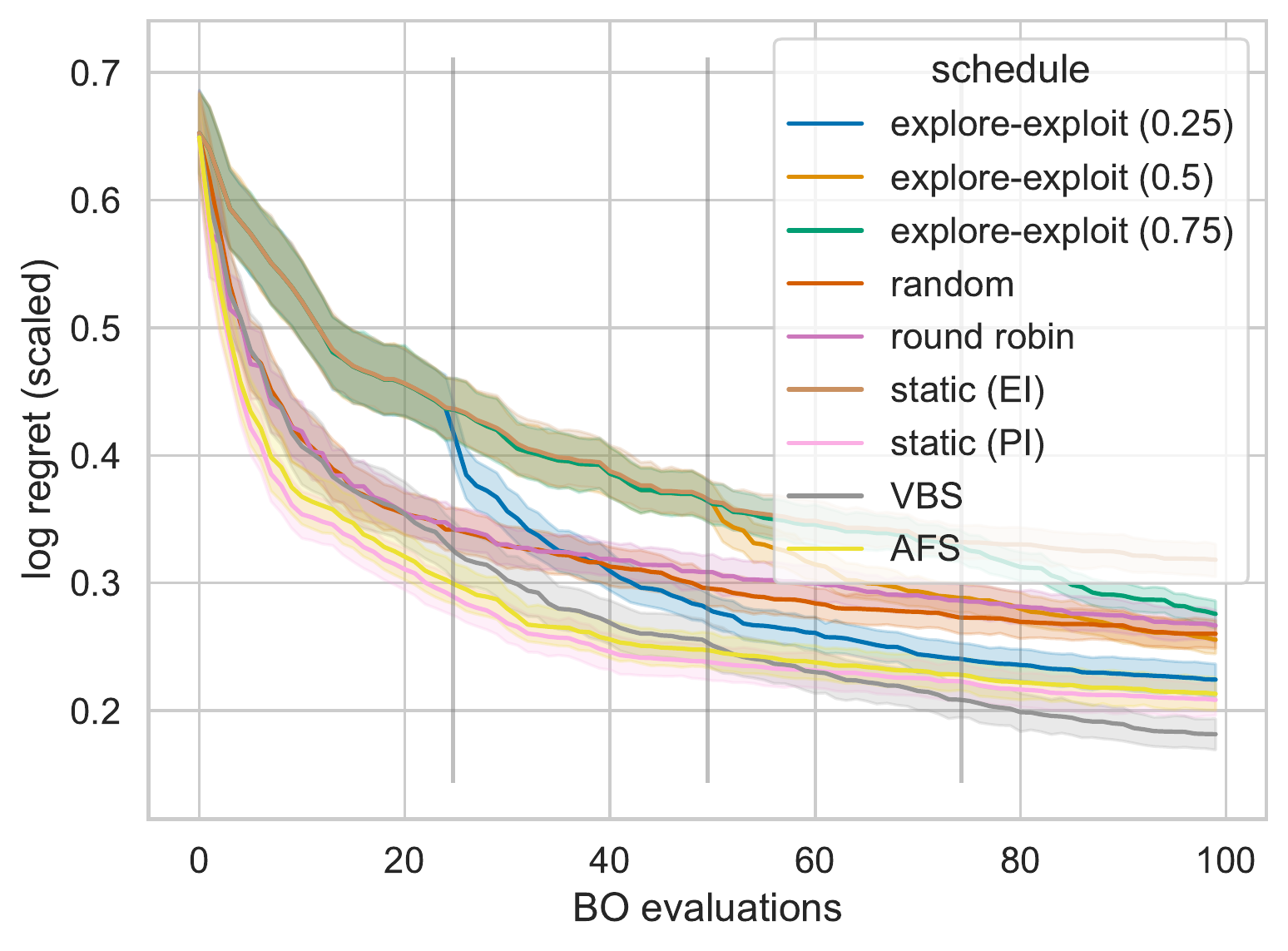}
        \caption{Log-Regret (Scaled) per Step}
        \label{subfig:convergence_11}
    \end{subfigure}\\
    \vspace*{3mm}
    \centering
    \begin{subfigure}[b]{\textwidth}
        \centering
        \includegraphics[width=\textwidth]{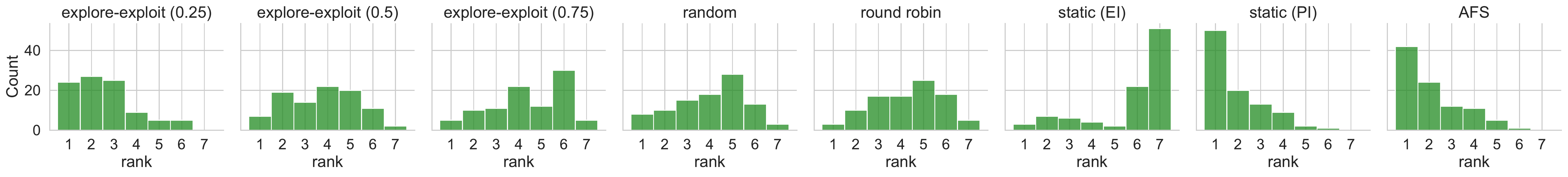}
        \caption{Rank}
        \label{subfig:rank_11}
    \end{subfigure}
    \caption{BBOB Function 11}
    \label{fig:bbob_function_11}
\end{figure}

\begin{figure}[h]
    \centering
    \begin{subfigure}[b]{0.45\textwidth}
        \centering
        \includegraphics[width=\textwidth]{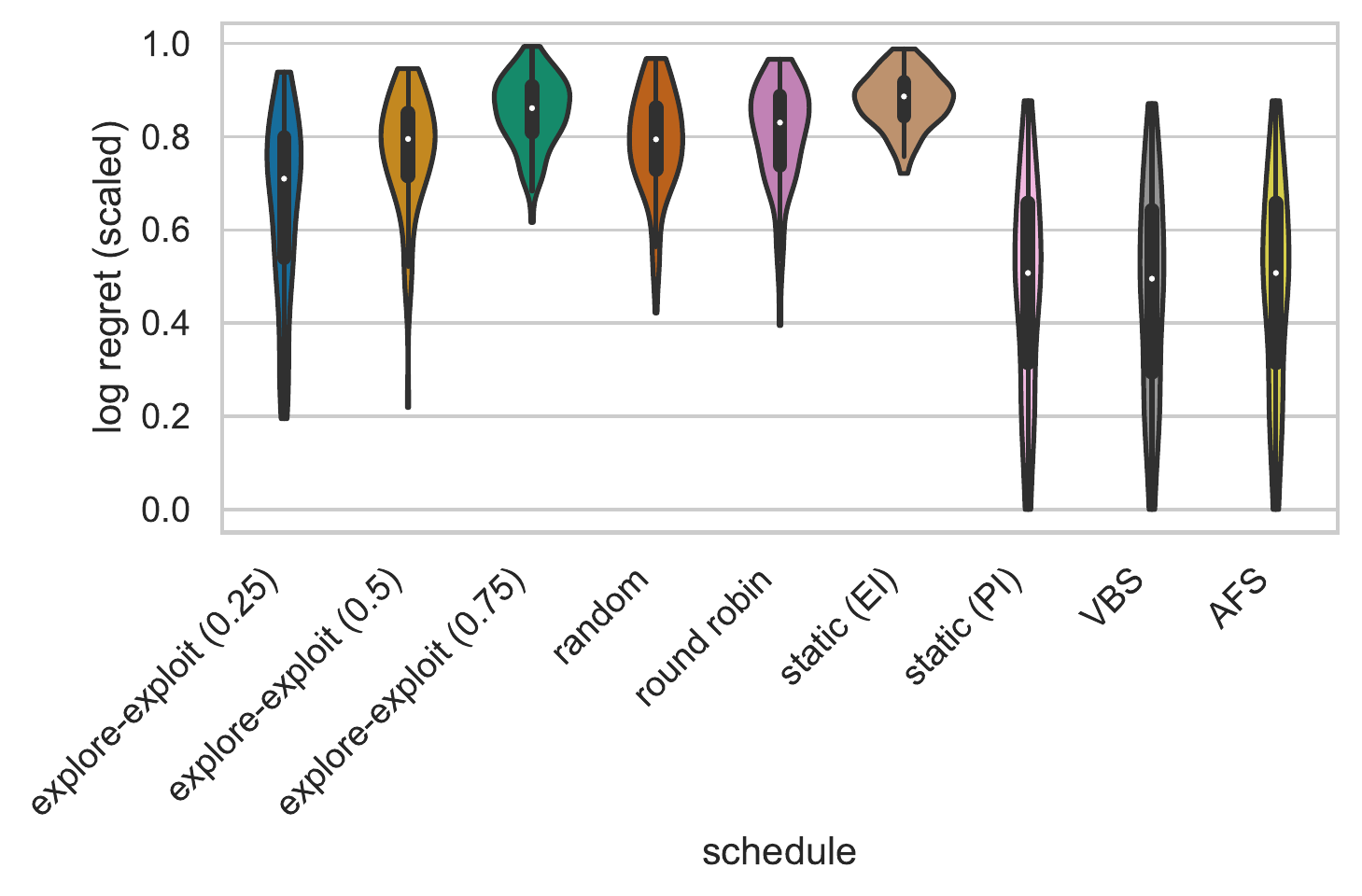}
        \caption{Final Log Regret (Scaled)}
        \label{subfig:boxplot_12}
    \end{subfigure}
    \hfill
    \begin{subfigure}[b]{0.45\textwidth}
        \centering
        \includegraphics[width=\textwidth]{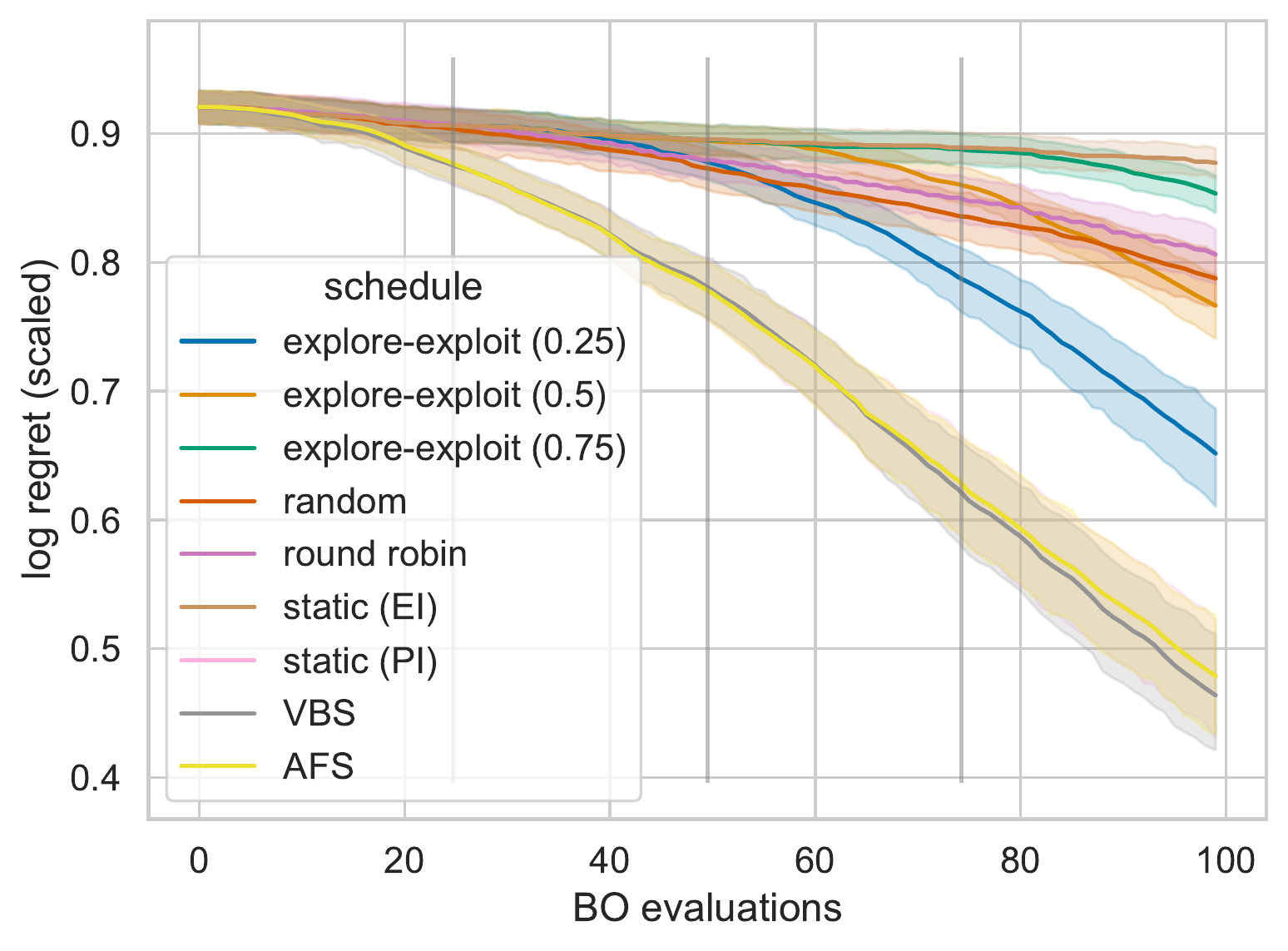}
        \caption{Log-Regret (Scaled) per Step}
        \label{subfig:convergence_12}
    \end{subfigure}\\
    \vspace*{3mm}
    \centering
    \begin{subfigure}[b]{\textwidth}
        \centering
        \includegraphics[width=\textwidth]{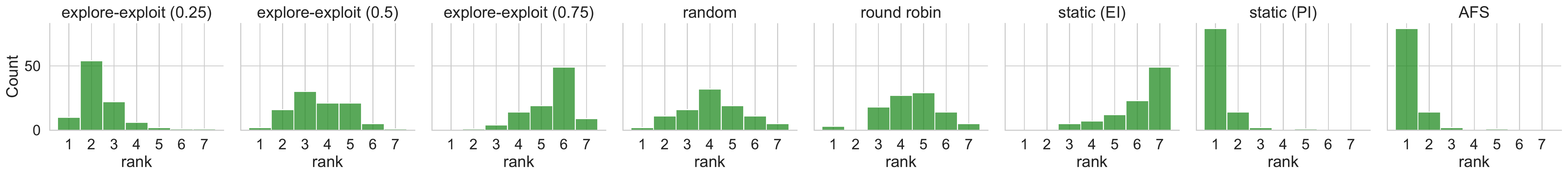}
        \caption{Rank}
        \label{subfig:rank_12}
    \end{subfigure}
    \caption{BBOB Function 12}
    \label{fig:bbob_function_12}
\end{figure}

\begin{figure}[h]
    \centering
    \begin{subfigure}[b]{0.45\textwidth}
        \centering
        \includegraphics[width=\textwidth]{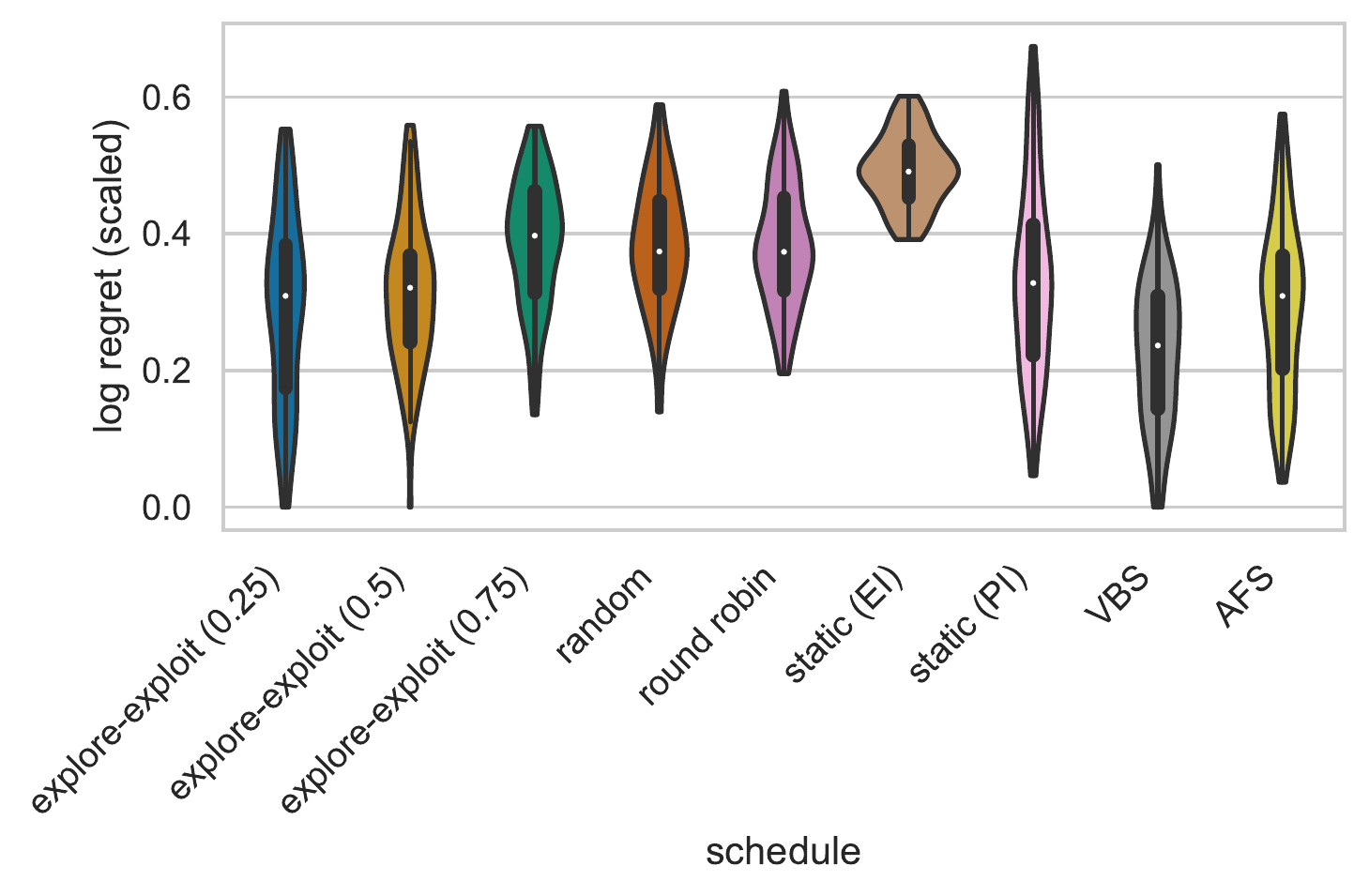}
        \caption{Final Log Regret (Scaled)}
        \label{subfig:boxplot_13}
    \end{subfigure}
    \hfill
    \begin{subfigure}[b]{0.45\textwidth}
        \centering
        \includegraphics[width=\textwidth]{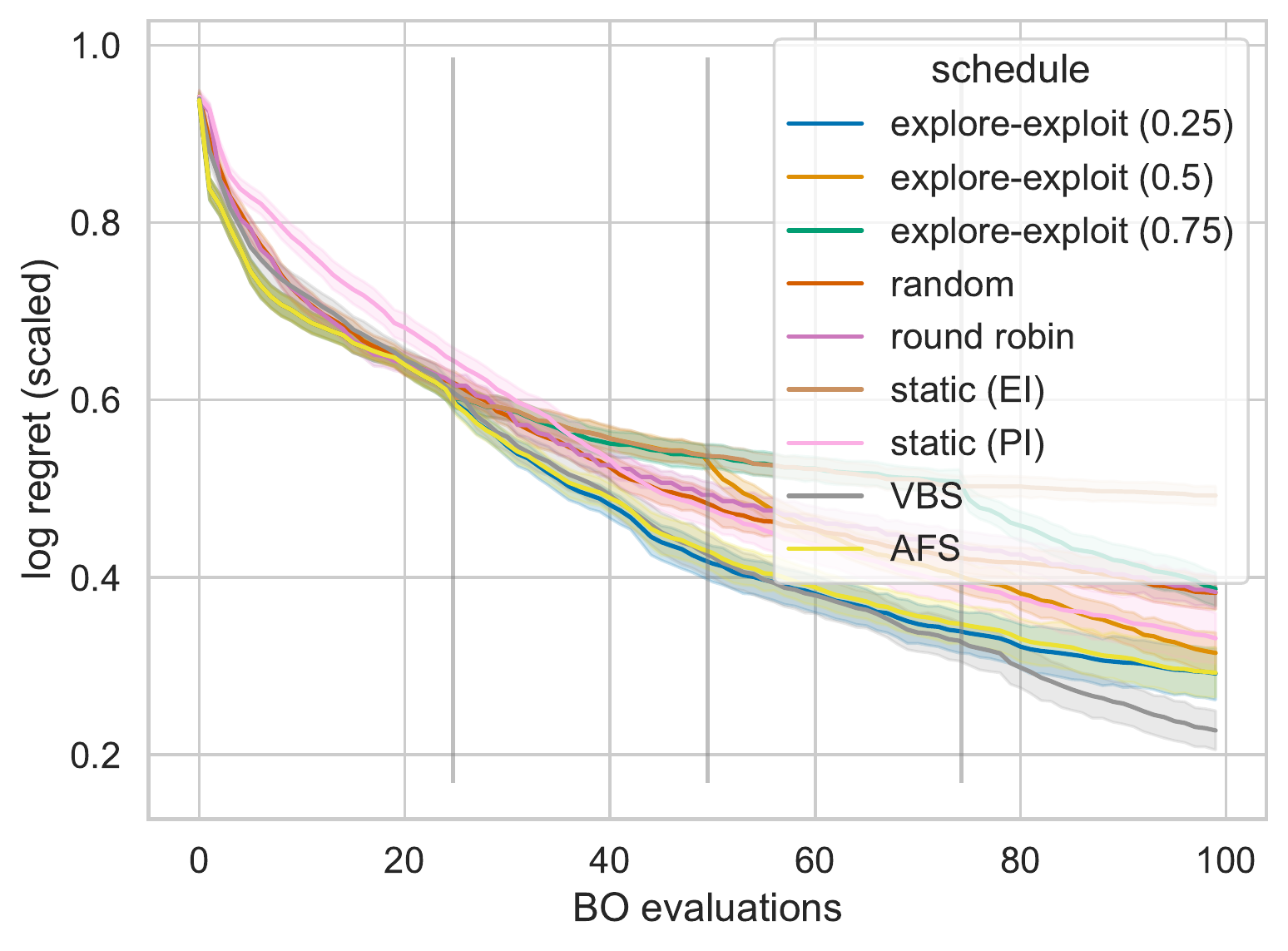}
        \caption{Log-Regret (Scaled) per Step}
        \label{subfig:convergence_13}
    \end{subfigure}\\
    \vspace*{3mm}
    \centering
    \begin{subfigure}[b]{\textwidth}
        \centering
        \includegraphics[width=\textwidth]{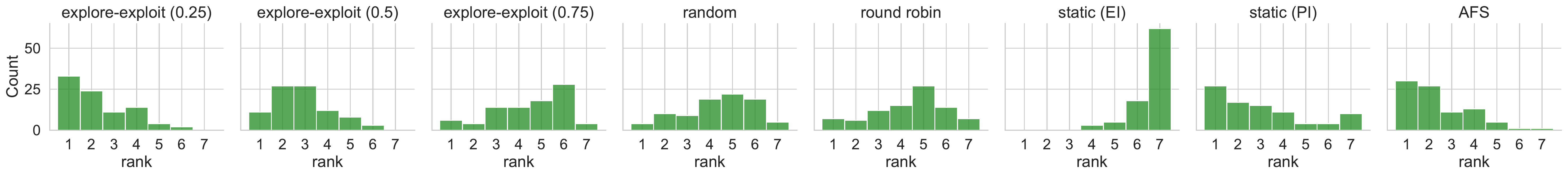}
        \caption{Rank}
        \label{subfig:rank_13}
    \end{subfigure}
    \caption{BBOB Function 13}
    \label{fig:bbob_function_13}
\end{figure}

\begin{figure}[h]
    \centering
    \begin{subfigure}[b]{0.45\textwidth}
        \centering
        \includegraphics[width=\textwidth]{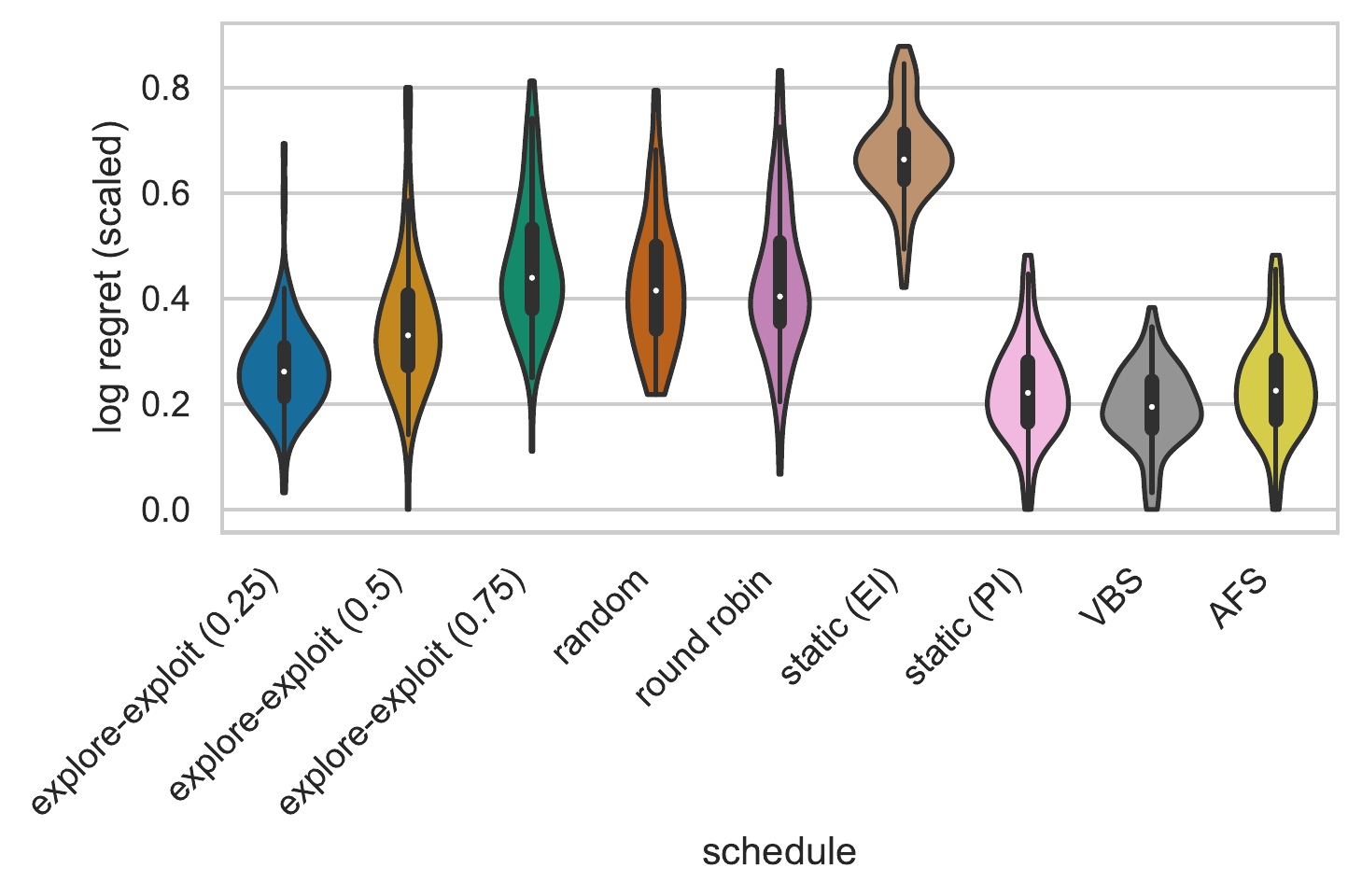}
        \caption{Final Log Regret (Scaled)}
        \label{subfig:boxplot_14}
    \end{subfigure}
    \hfill
    \begin{subfigure}[b]{0.45\textwidth}
        \centering
        \includegraphics[width=\textwidth]{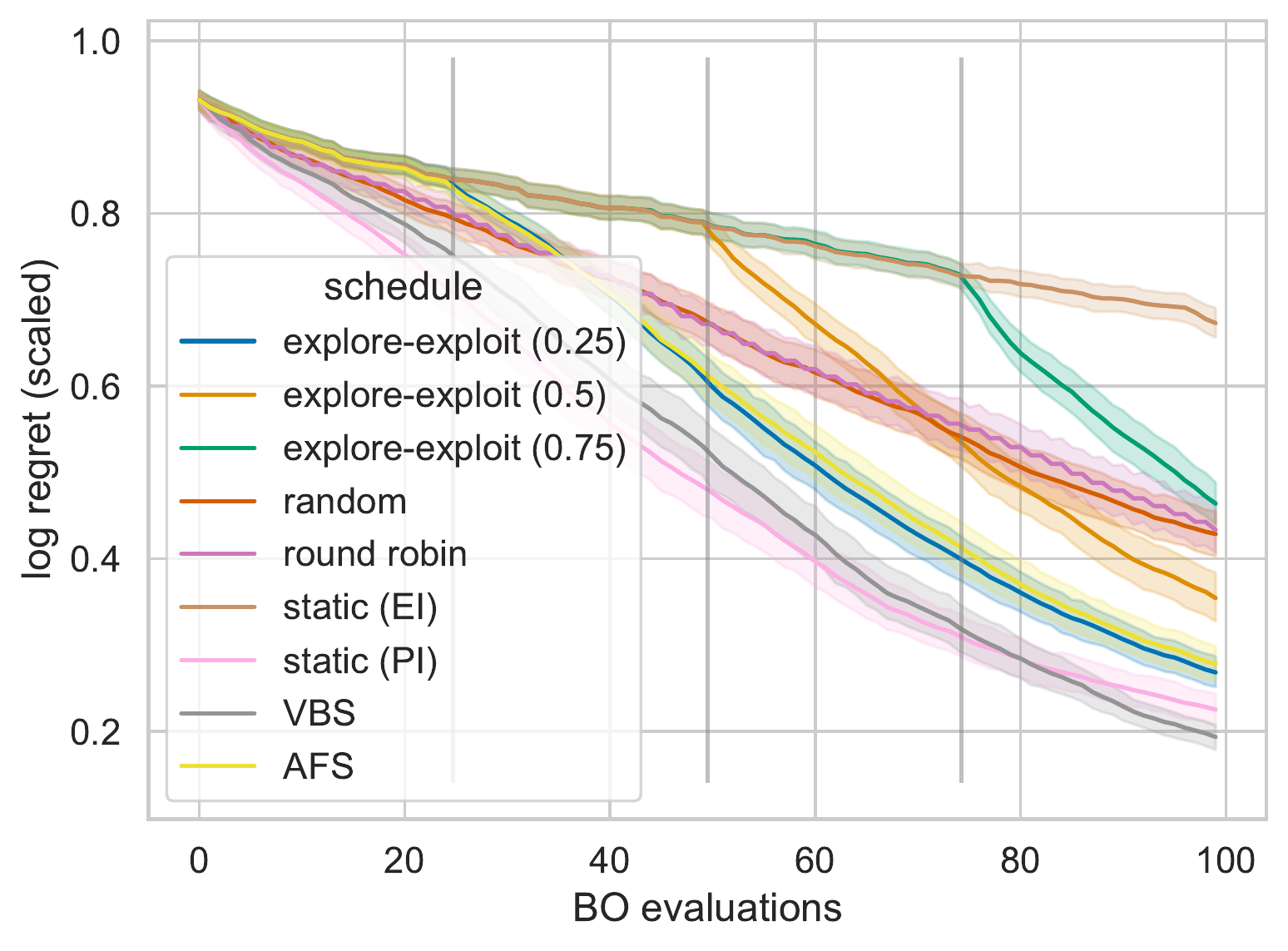}
        \caption{Log-Regret (Scaled) per Step}
        \label{subfig:convergence_14}
    \end{subfigure}\\
    \vspace*{3mm}
    \centering
    \begin{subfigure}[b]{\textwidth}
        \centering
        \includegraphics[width=\textwidth]{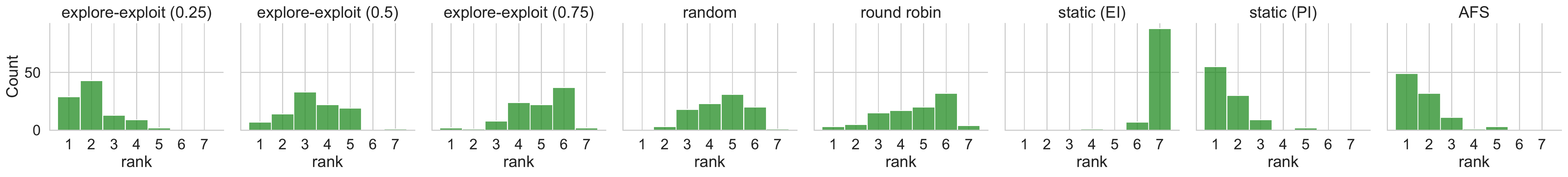}
        \caption{Rank}
        \label{subfig:rank_14}
    \end{subfigure}
    \caption{BBOB Function 14}
    \label{fig:bbob_function_14}
\end{figure}

\begin{figure}[h]
    \centering
    \begin{subfigure}[b]{0.45\textwidth}
        \centering
        \includegraphics[width=\textwidth]{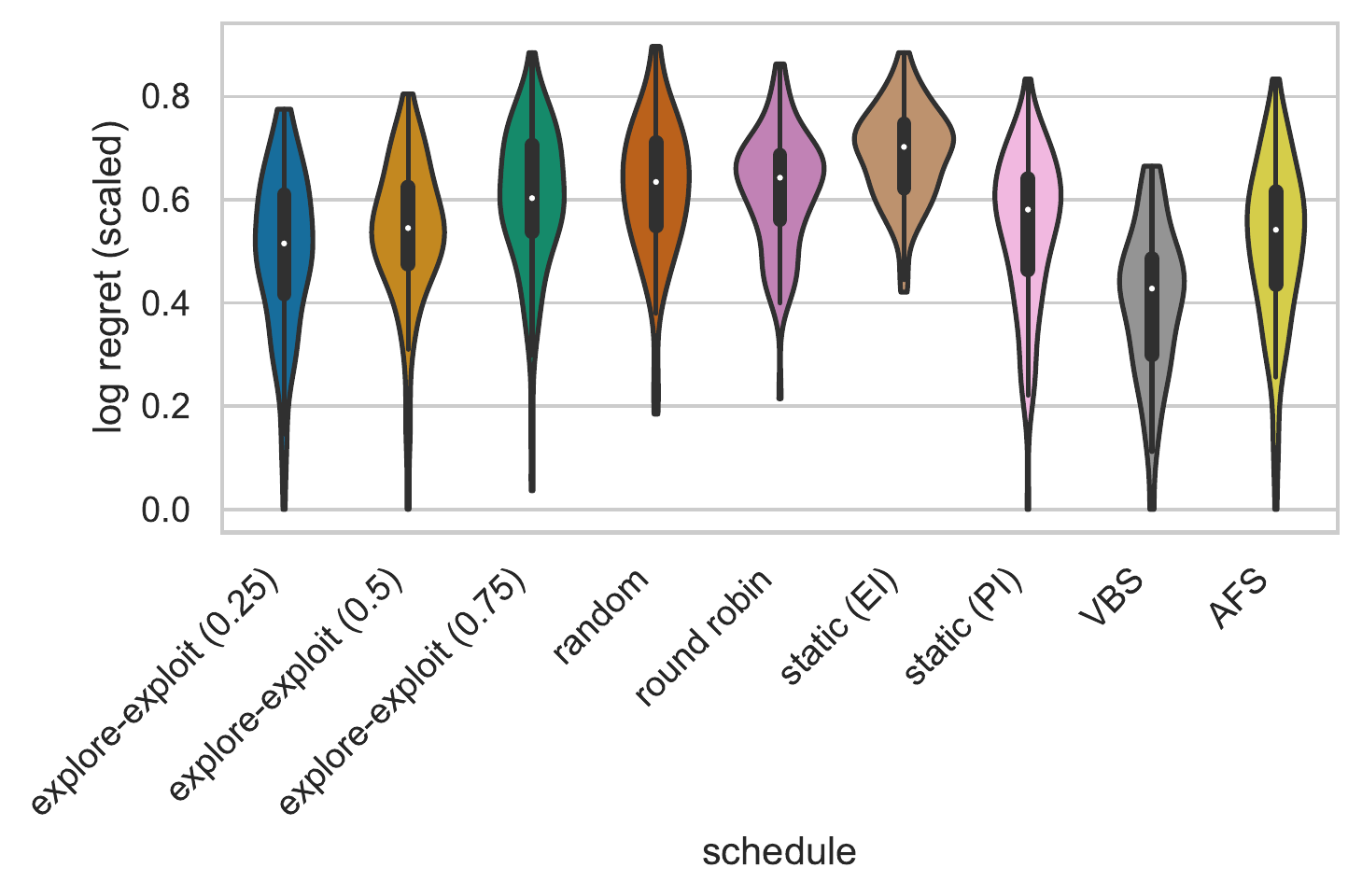}
        \caption{Final Log Regret (Scaled)}
        \label{subfig:boxplot_15}
    \end{subfigure}
    \hfill
    \begin{subfigure}[b]{0.45\textwidth}
        \centering
        \includegraphics[width=\textwidth]{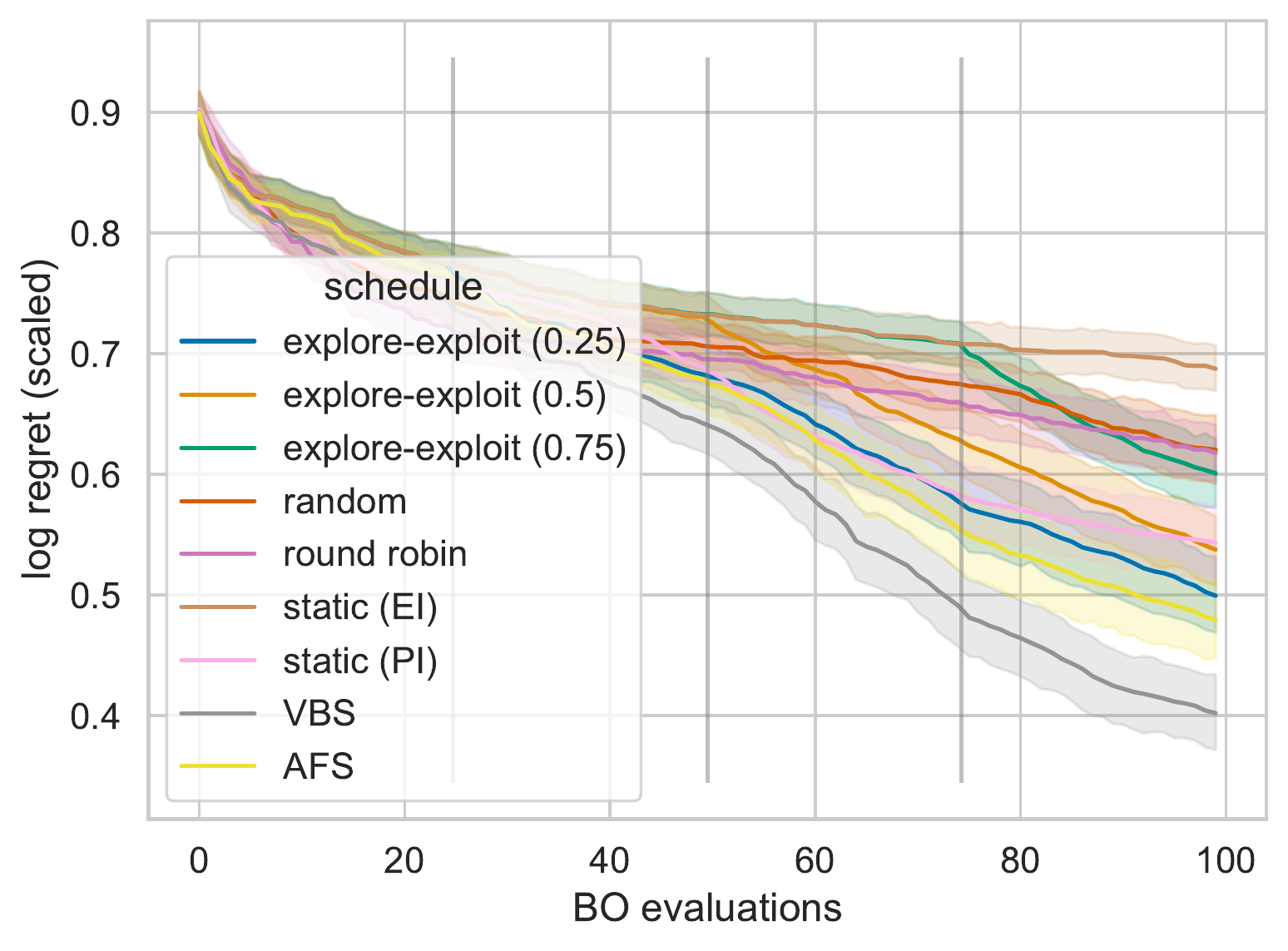}
        \caption{Log-Regret (Scaled) per Step}
        \label{subfig:convergence_15}
    \end{subfigure}\\
    \vspace*{3mm}
    \centering
    \begin{subfigure}[b]{\textwidth}
        \centering
        \includegraphics[width=\textwidth]{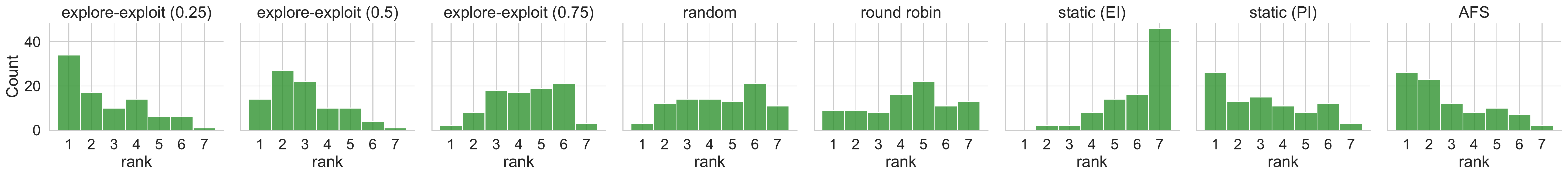}
        \caption{Rank}
        \label{subfig:rank_15}
    \end{subfigure}
    \caption{BBOB Function 15}
    \label{fig:bbob_function_15}
\end{figure}

\begin{figure}[h]
    \centering
    \begin{subfigure}[b]{0.45\textwidth}
        \centering
        \includegraphics[width=\textwidth]{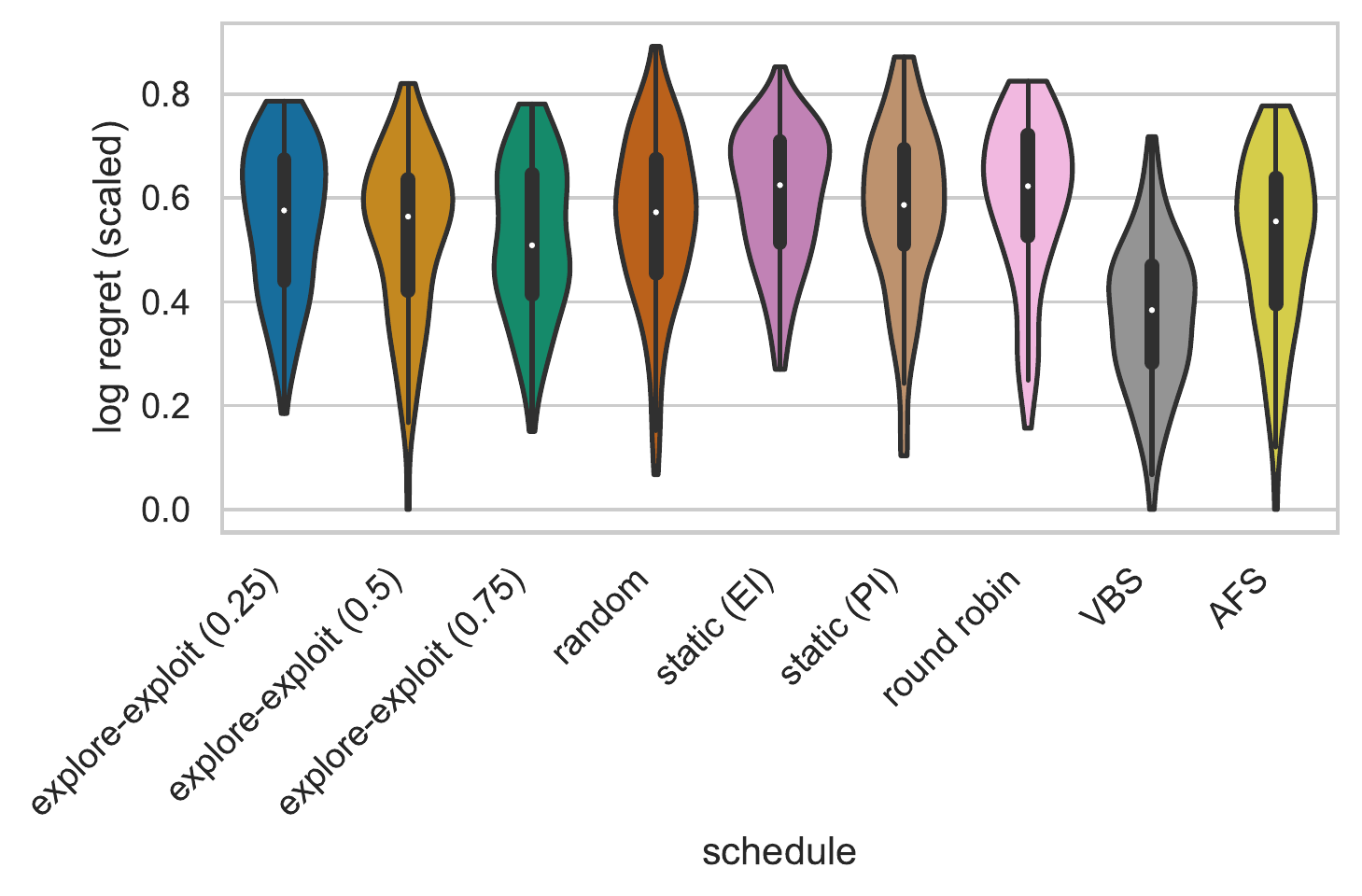}
        \caption{Final Log Regret (Scaled)}
        \label{subfig:boxplot_17}
    \end{subfigure}
    \hfill
    \begin{subfigure}[b]{0.45\textwidth}
        \centering
        \includegraphics[width=\textwidth]{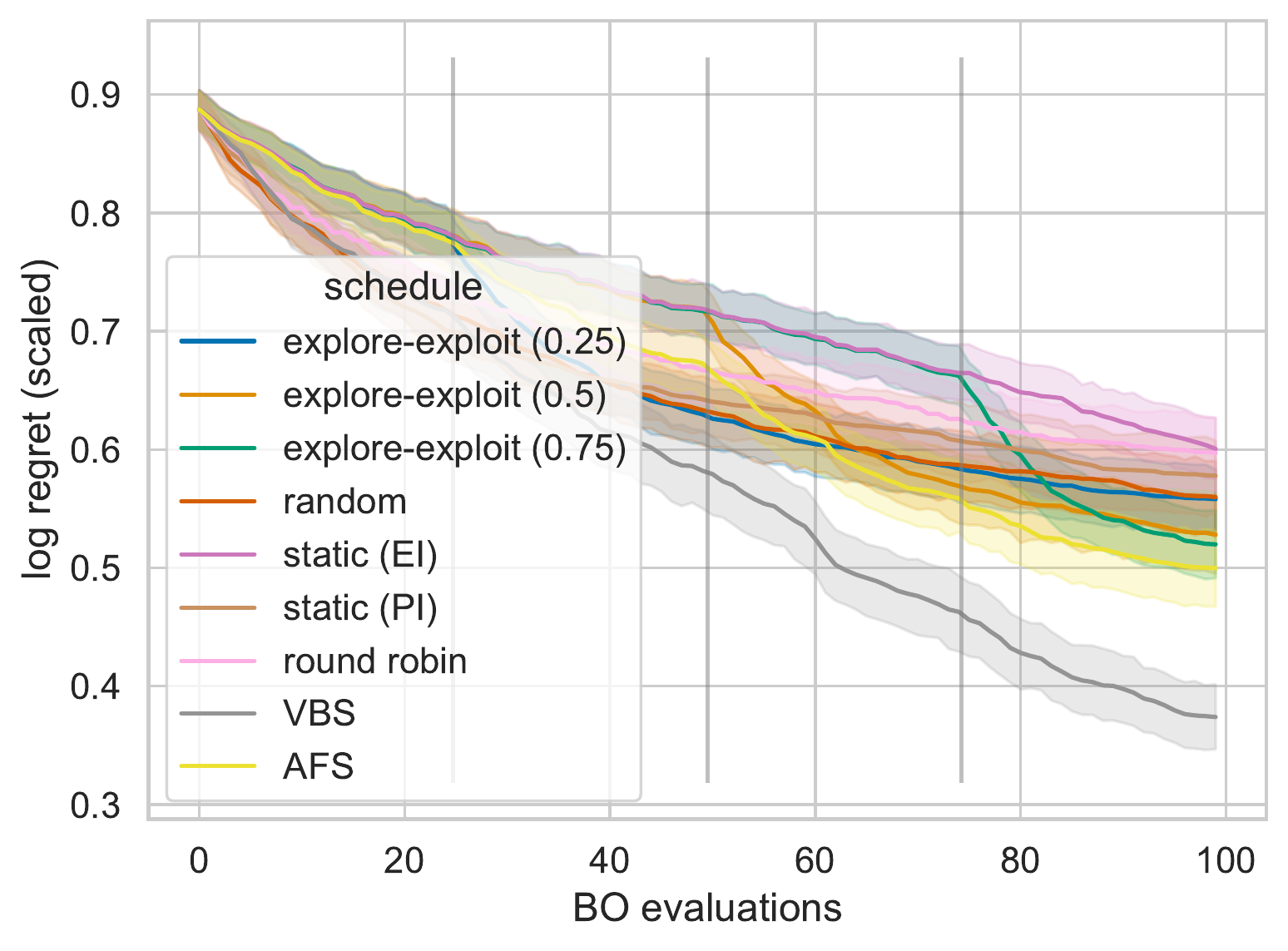}
        \caption{Log-Regret (Scaled) per Step}
        \label{subfig:convergence_17}
    \end{subfigure}\\
    \vspace*{3mm}
    \centering
    \begin{subfigure}[b]{\textwidth}
        \centering
        \includegraphics[width=\textwidth]{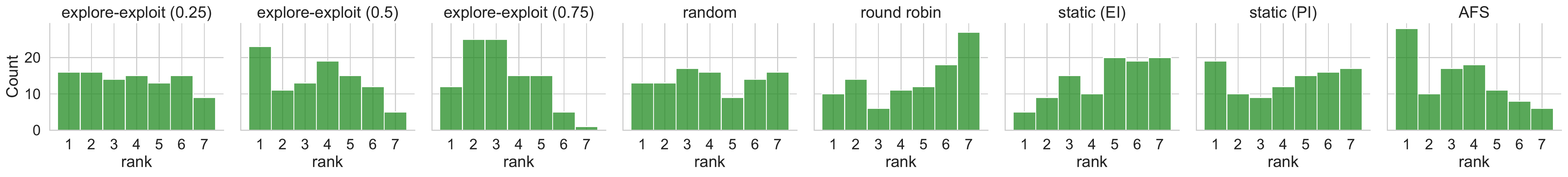}
        \caption{Rank}
        \label{subfig:rank_17}
    \end{subfigure}
    \caption{BBOB Function 17}
    \label{fig:bbob_function_17}
\end{figure}

\begin{figure}[h]
    \centering
    \begin{subfigure}[b]{0.45\textwidth}
        \centering
        \includegraphics[width=\textwidth]{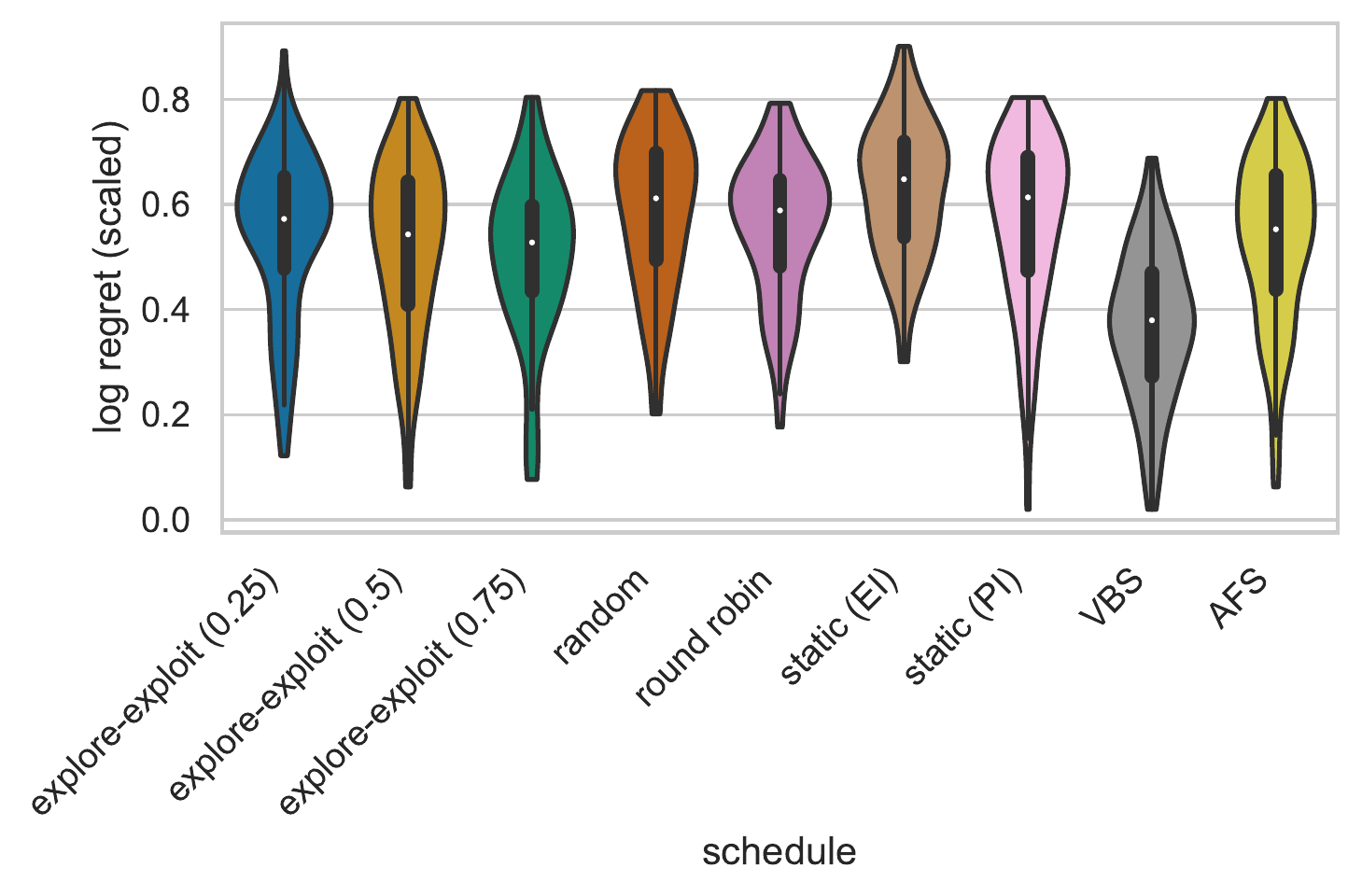}
        \caption{Final Log Regret (Scaled)}
        \label{subfig:boxplot_18}
    \end{subfigure}
    \hfill
    \begin{subfigure}[b]{0.45\textwidth}
        \centering
        \includegraphics[width=\textwidth]{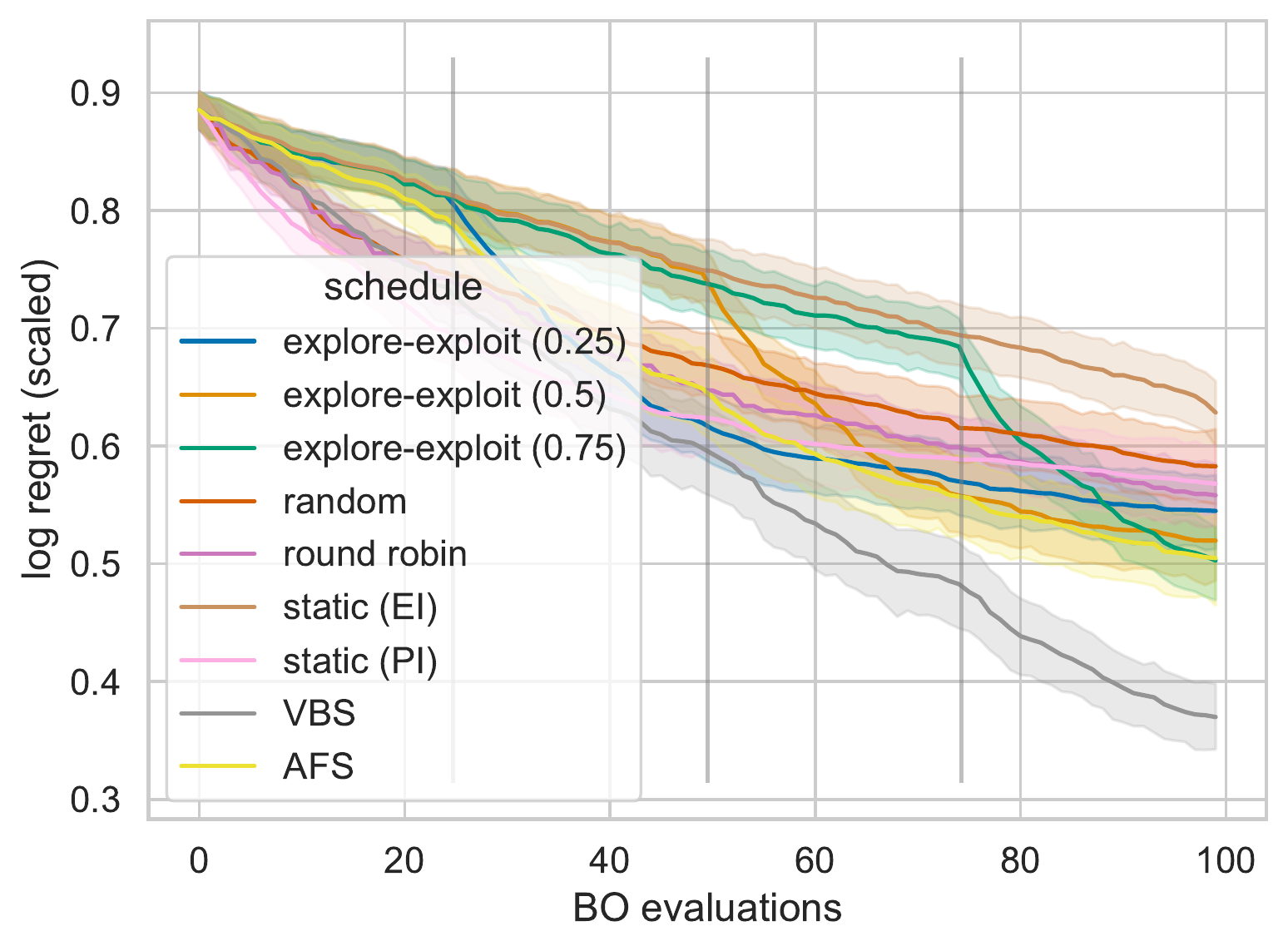}
        \caption{Log-Regret (Scaled) per Step}
        \label{subfig:convergence_18}
    \end{subfigure}\\
    \vspace*{3mm}
    \centering
    \begin{subfigure}[b]{\textwidth}
        \centering
        \includegraphics[width=\textwidth]{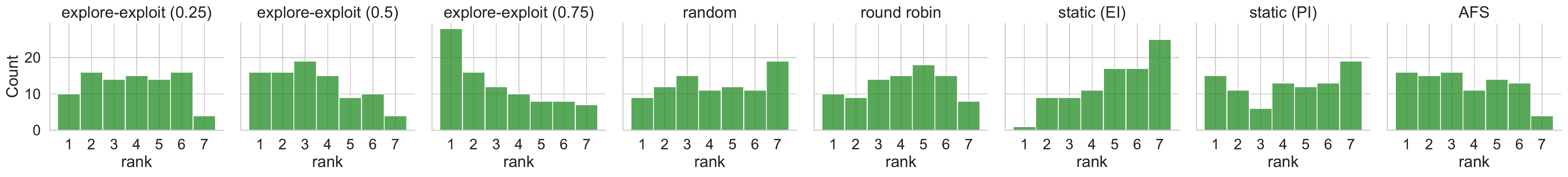}
        \caption{Rank}
        \label{subfig:rank_18}
    \end{subfigure}
    \caption{BBOB Function 18}
    \label{fig:bbob_function_18}
\end{figure}

\begin{figure}[h]
    \centering
    \begin{subfigure}[b]{0.45\textwidth}
        \centering
        \includegraphics[width=\textwidth]{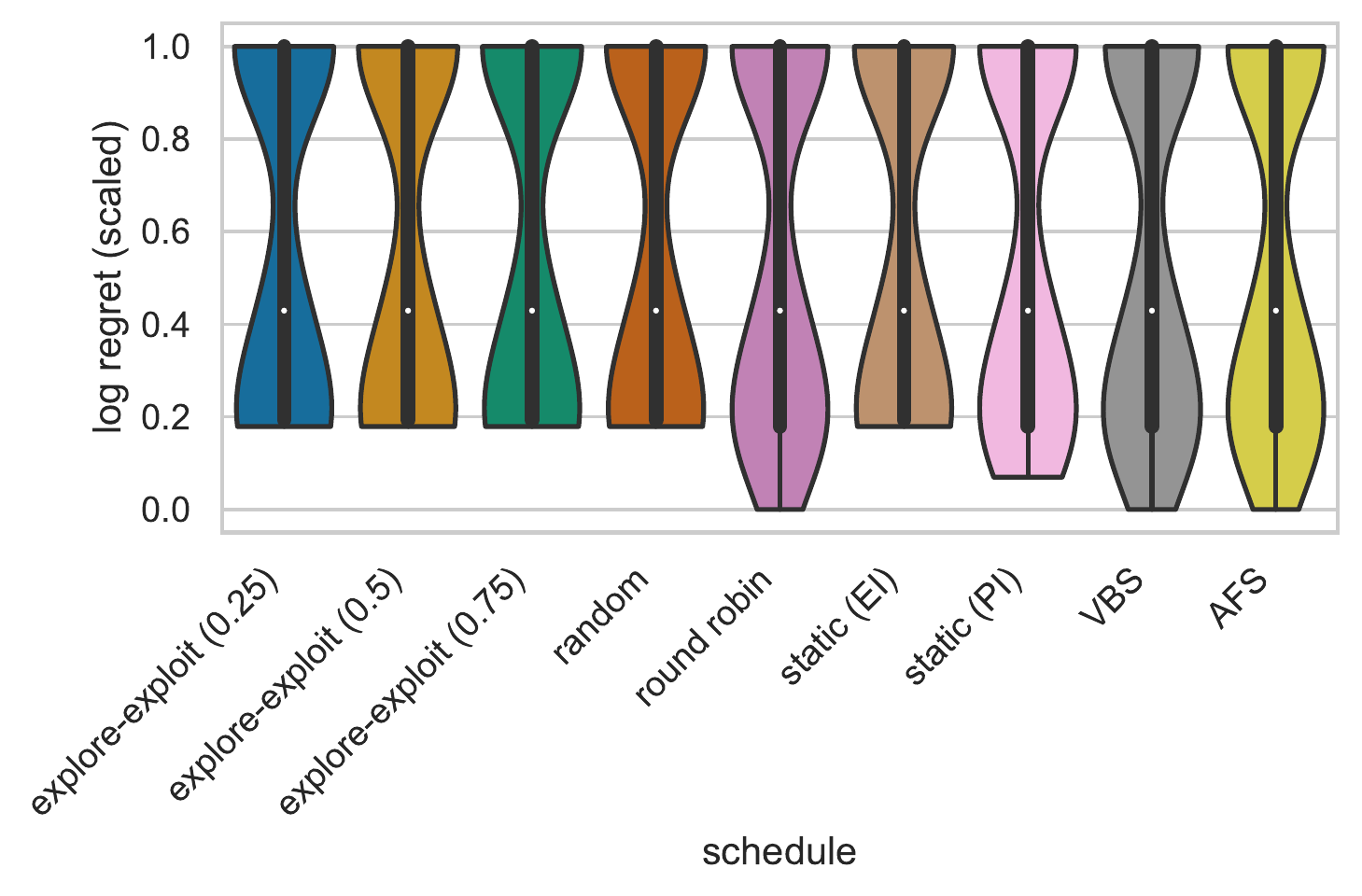}
        \caption{Final Log Regret (Scaled)}
        \label{subfig:boxplot_19}
    \end{subfigure}
    \hfill
    \begin{subfigure}[b]{0.45\textwidth}
        \centering
        \includegraphics[width=\textwidth]{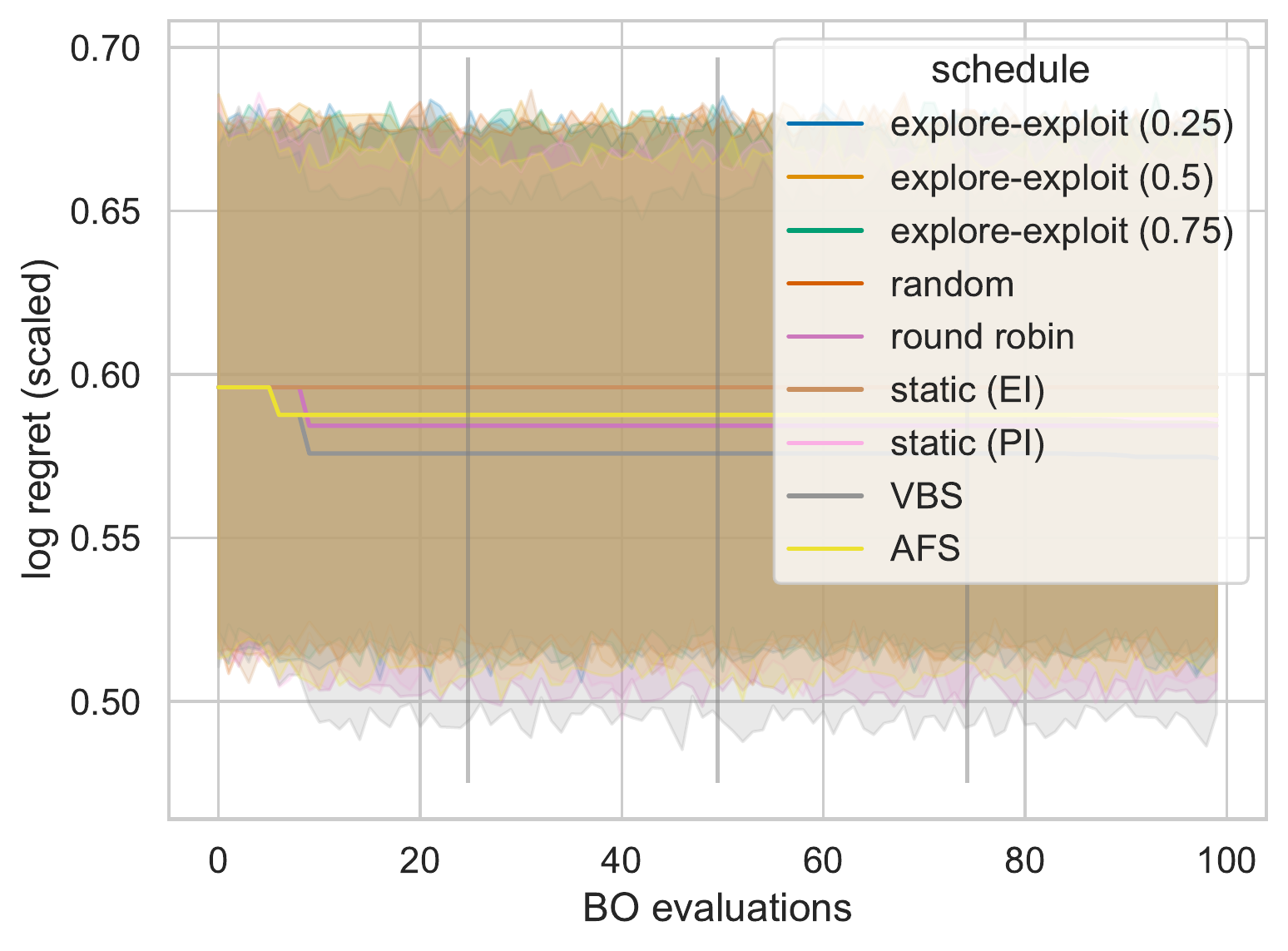}
        \caption{Log-Regret (Scaled) per Step}
        \label{subfig:convergence_19}
    \end{subfigure}\\
    \vspace*{3mm}
    \centering
    \begin{subfigure}[b]{\textwidth}
        \centering
        \includegraphics[width=\textwidth]{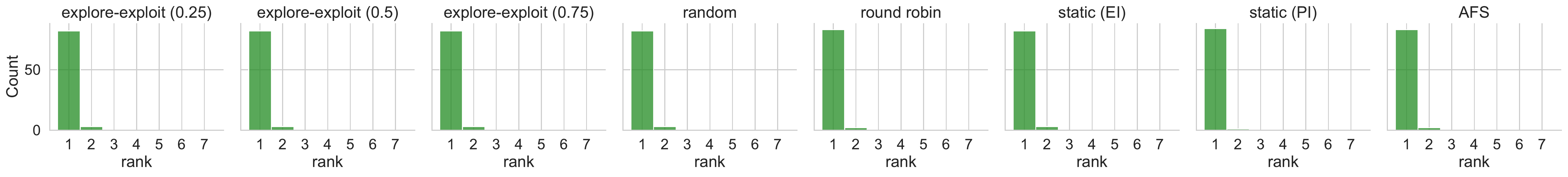}
        \caption{Rank}
        \label{subfig:rank_19}
    \end{subfigure}
    \caption{BBOB Function 19}
    \label{fig:bbob_function_19}
\end{figure}

\begin{figure}[h]
    \centering
    \begin{subfigure}[b]{0.45\textwidth}
        \centering
        \includegraphics[width=\textwidth]{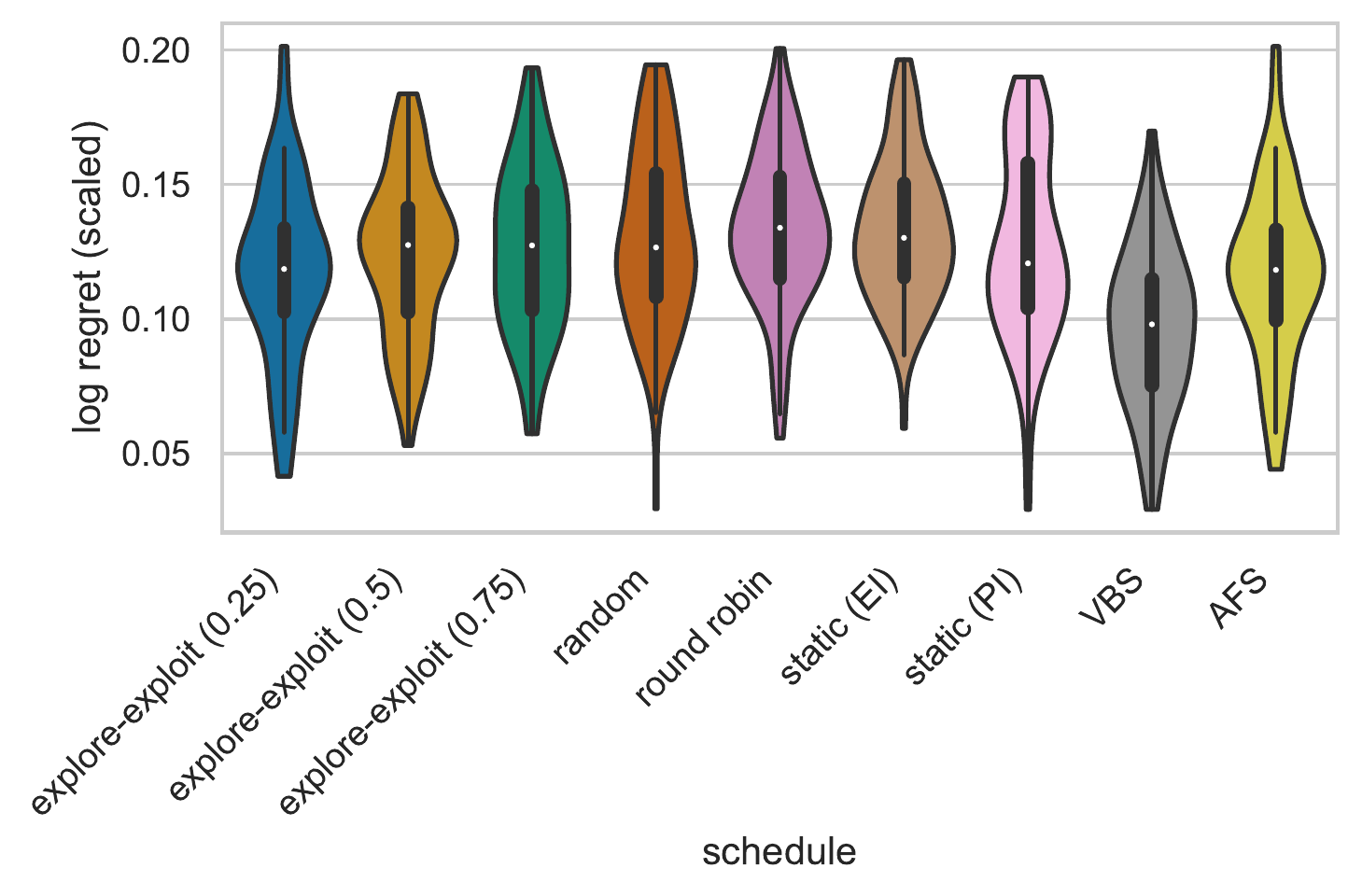}
        \caption{Final Log Regret (Scaled)}
        \label{subfig:boxplot_20}
    \end{subfigure}
    \hfill
    \begin{subfigure}[b]{0.45\textwidth}
        \centering
        \includegraphics[width=\textwidth]{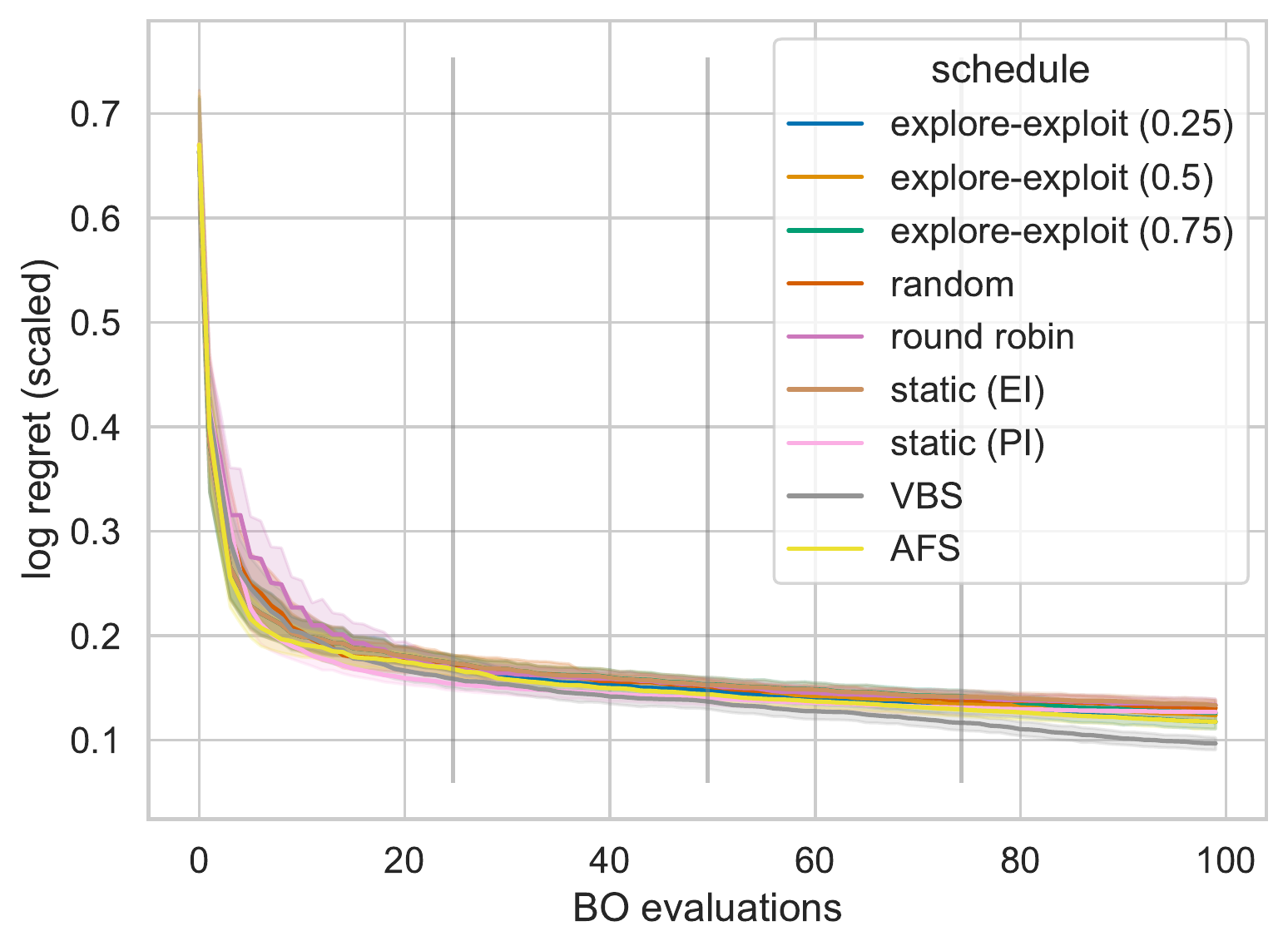}
        \caption{Log-Regret (Scaled) per Step}
        \label{subfig:convergence_20}
    \end{subfigure}\\
    \vspace*{3mm}
    \centering
    \begin{subfigure}[b]{\textwidth}
        \centering
        \includegraphics[width=\textwidth]{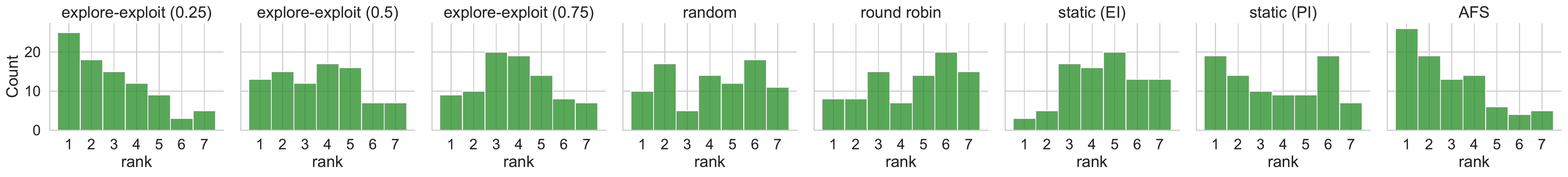}
        \caption{Rank}
        \label{subfig:rank_20}
    \end{subfigure}
    \caption{BBOB Function 20}
    \label{fig:bbob_function_20}
\end{figure}

\begin{figure}[h]
    \centering
    \begin{subfigure}[b]{0.45\textwidth}
        \centering
        \includegraphics[width=\textwidth]{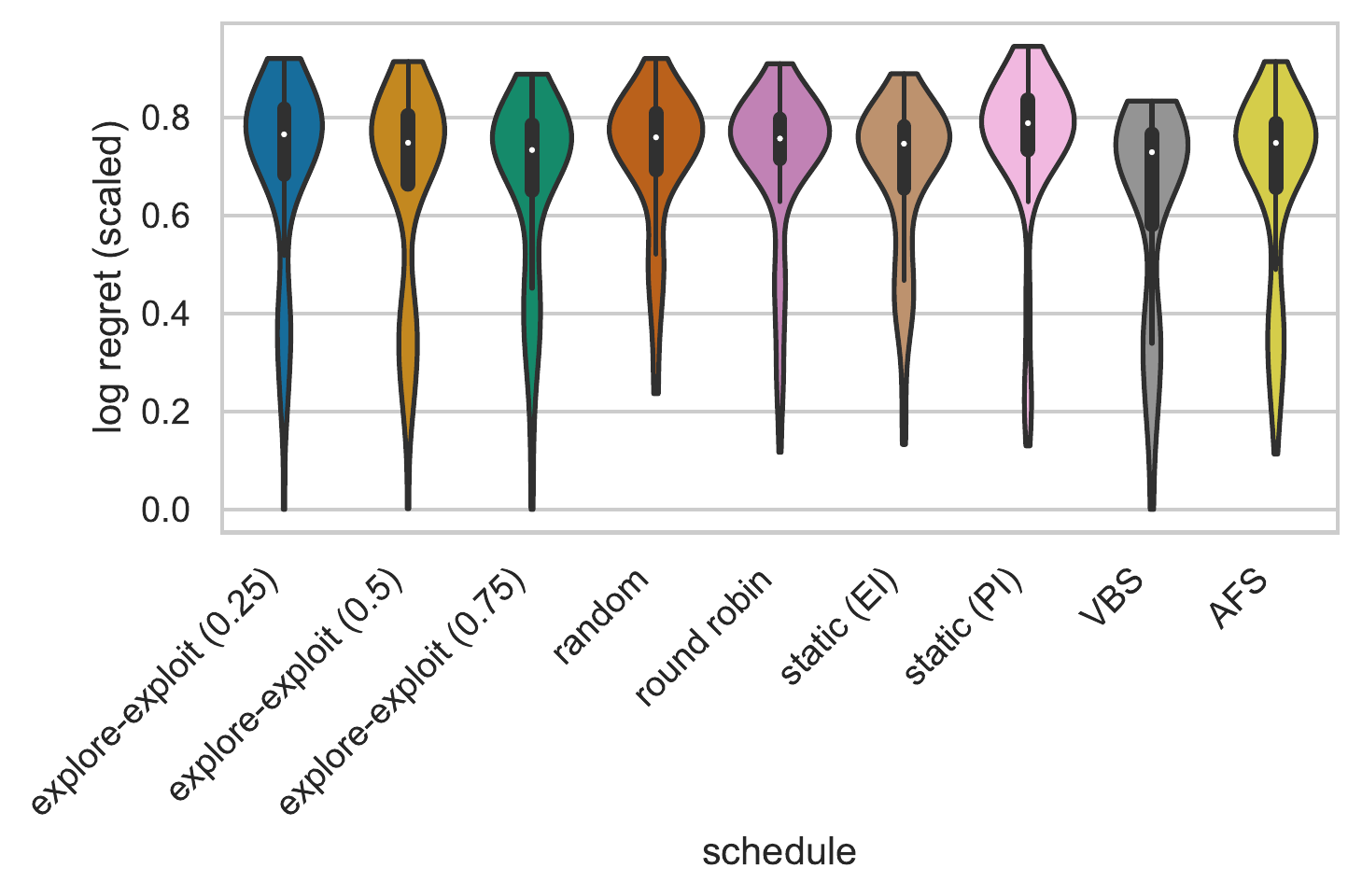}
        \caption{Final Log Regret (Scaled)}
        \label{subfig:boxplot_21}
    \end{subfigure}
    \hfill
    \begin{subfigure}[b]{0.45\textwidth}
        \centering
        \includegraphics[width=\textwidth]{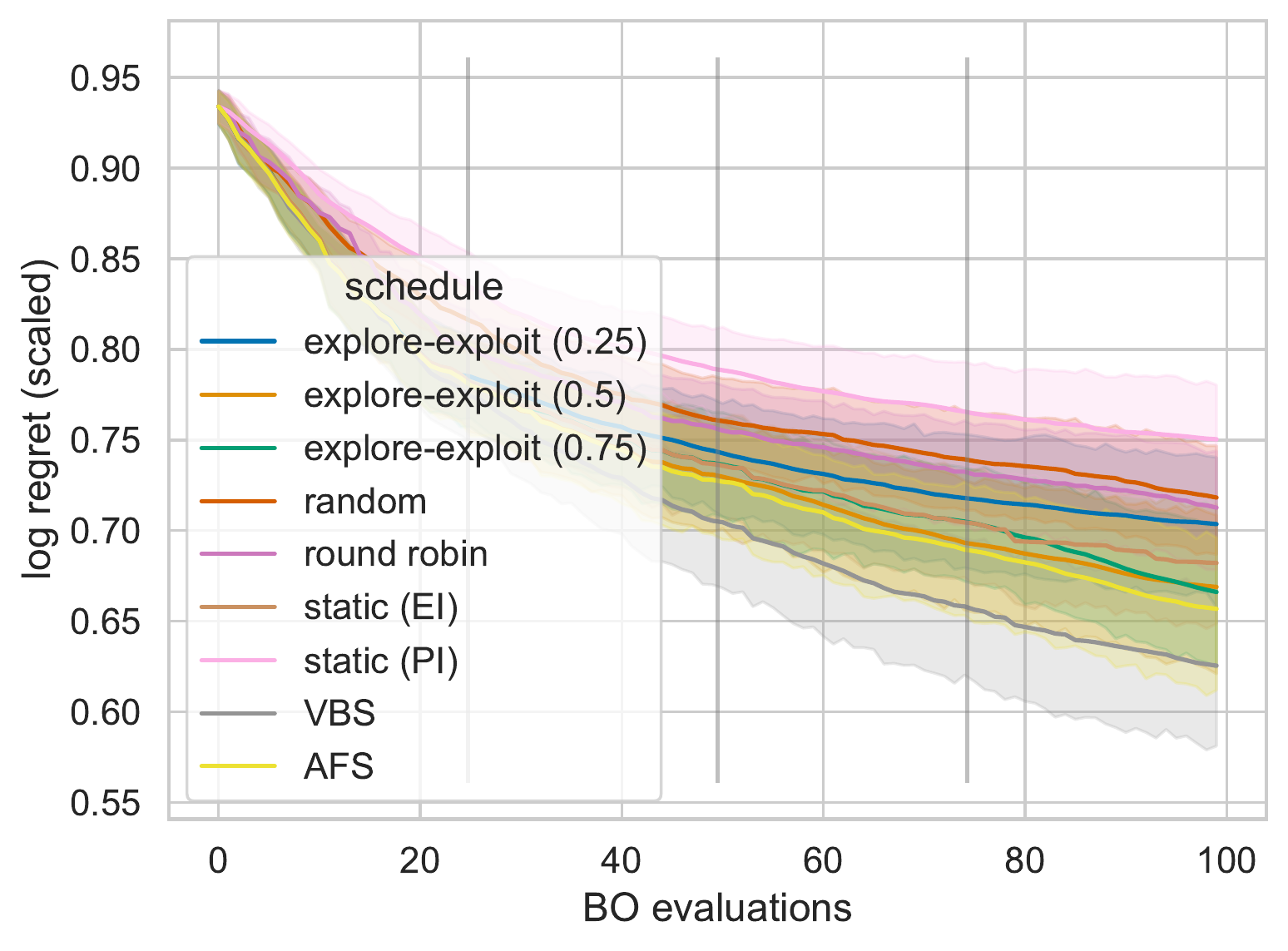}
        \caption{Log-Regret (Scaled) per Step}
        \label{subfig:convergence_21}
    \end{subfigure}\\
    \vspace*{3mm}
    \centering
    \begin{subfigure}[b]{\textwidth}
        \centering
        \includegraphics[width=\textwidth]{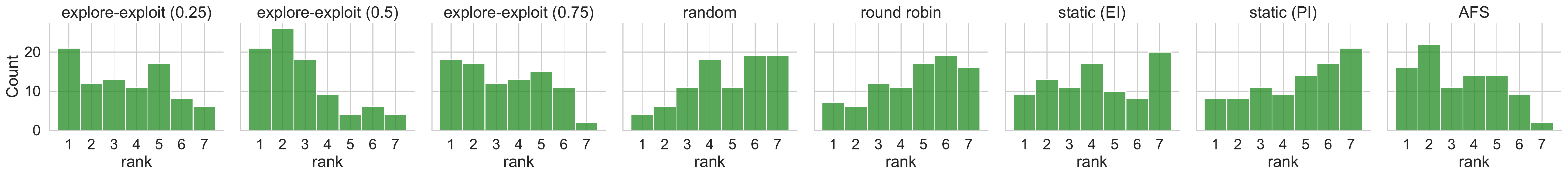}
        \caption{Rank}
        \label{subfig:rank_21}
    \end{subfigure}
    \caption{BBOB Function 21}
    \label{fig:bbob_function_21}
\end{figure}

\begin{figure}[h]
    \centering
    \begin{subfigure}[b]{0.45\textwidth}
        \centering
        \includegraphics[width=\textwidth]{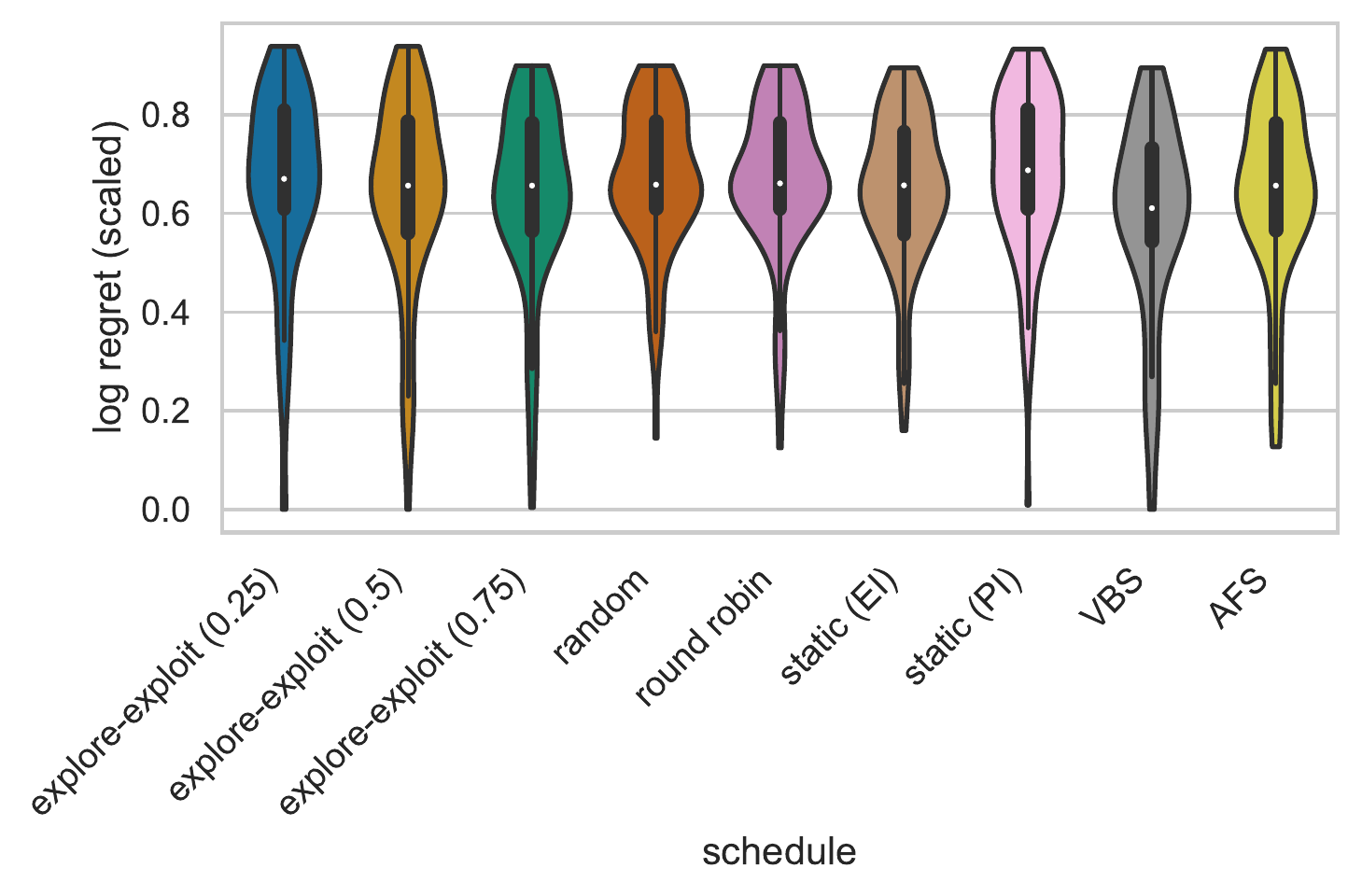}
        \caption{Final Log Regret (Scaled)}
        \label{subfig:boxplot_22}
    \end{subfigure}
    \hfill
    \begin{subfigure}[b]{0.45\textwidth}
        \centering
        \includegraphics[width=\textwidth]{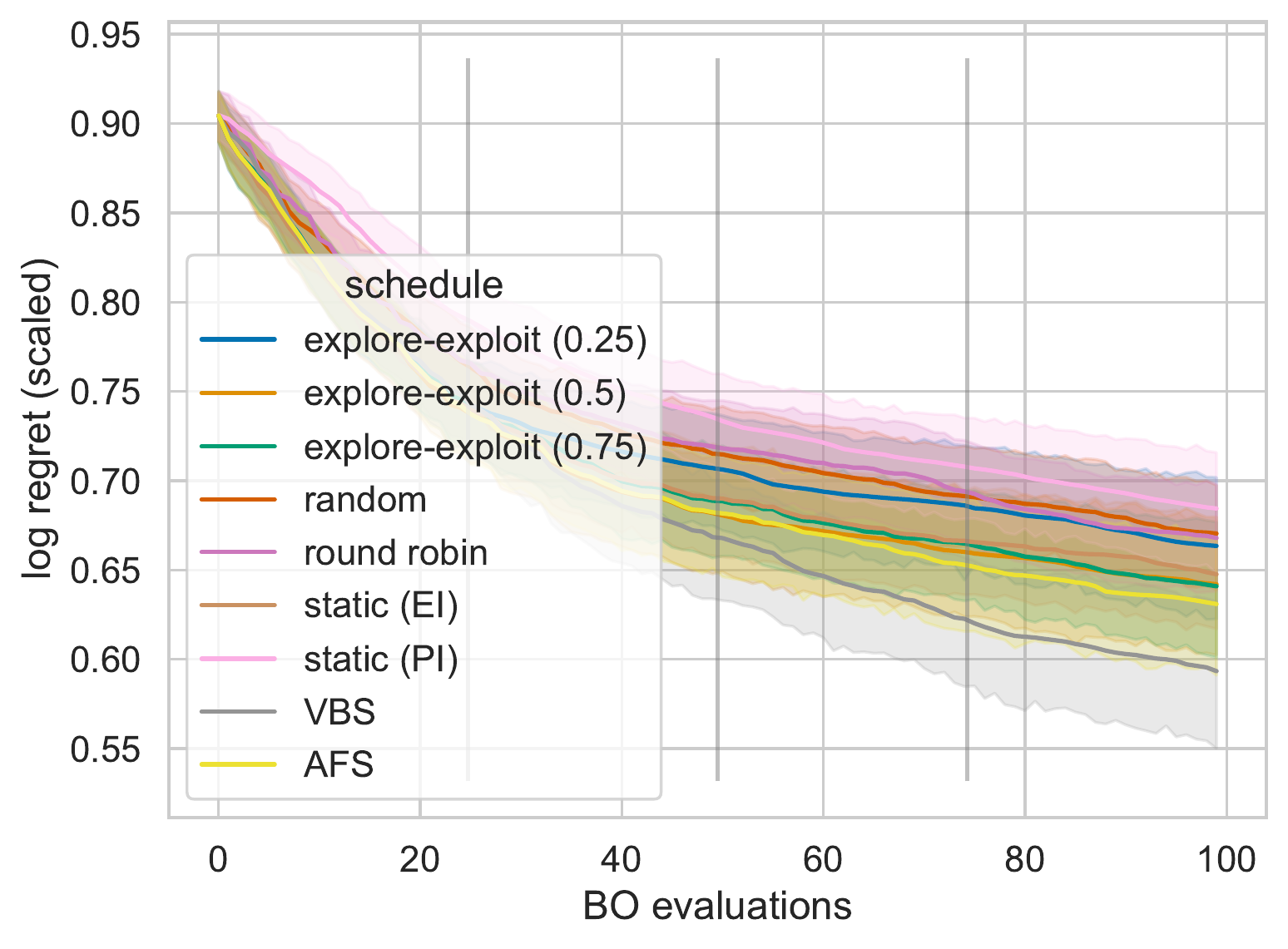}
        \caption{Log-Regret (Scaled) per Step}
        \label{subfig:convergence_22}
    \end{subfigure}\\
    \vspace*{3mm}
    \centering
    \begin{subfigure}[b]{\textwidth}
        \centering
        \includegraphics[width=\textwidth]{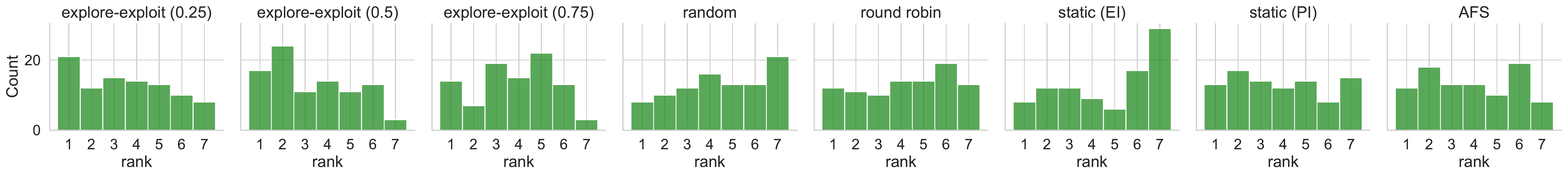}
        \caption{Rank}
        \label{subfig:rank_22}
    \end{subfigure}
    \caption{BBOB Function 22}
    \label{fig:bbob_function_22}
\end{figure}

\begin{figure}[h]
    \centering
    \begin{subfigure}[b]{0.45\textwidth}
        \centering
        \includegraphics[width=\textwidth]{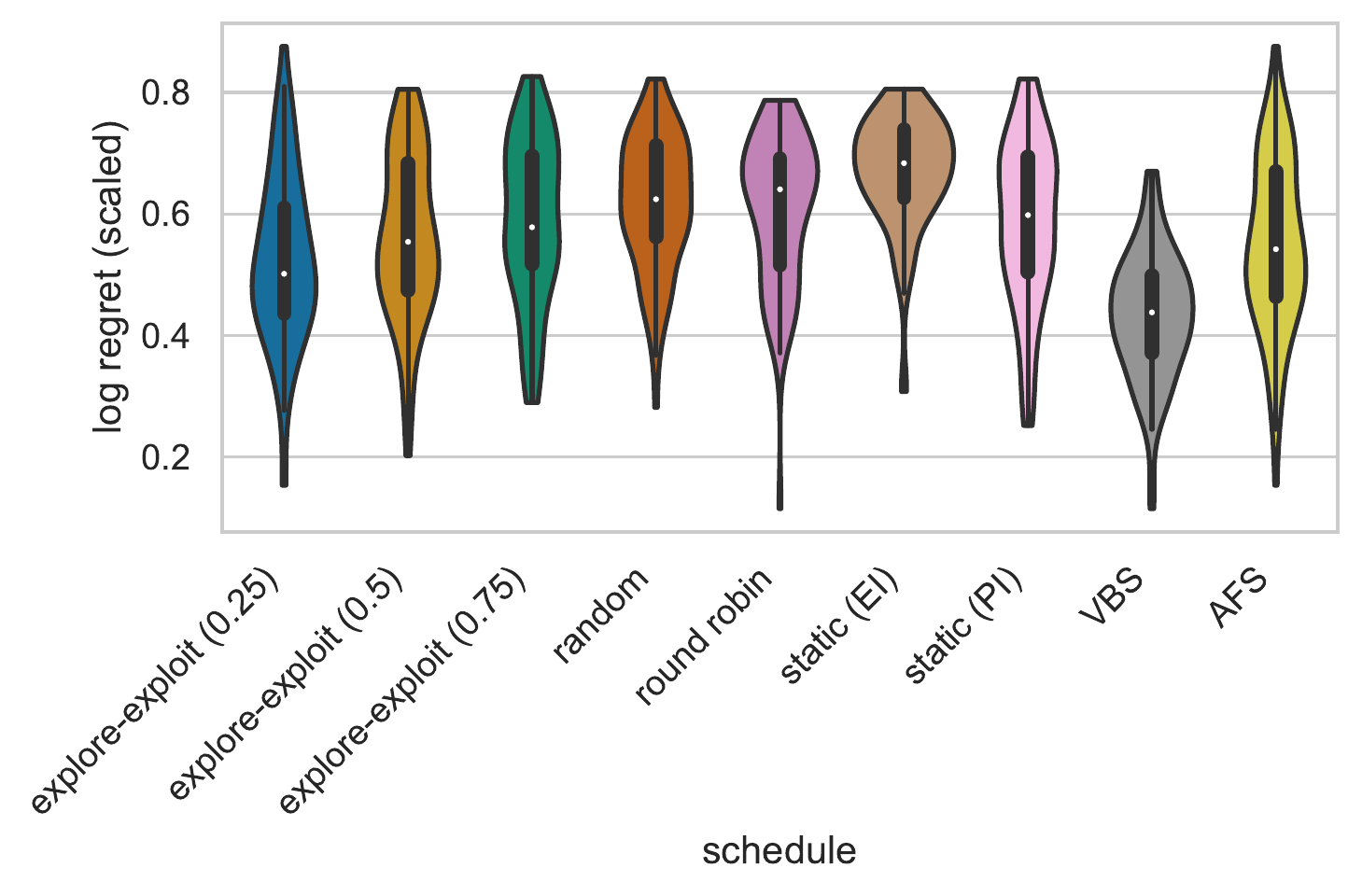}
        \caption{Final Log Regret (Scaled)}
        \label{subfig:boxplot_24}
    \end{subfigure}
    \hfill
    \begin{subfigure}[b]{0.45\textwidth}
        \centering
        \includegraphics[width=\textwidth]{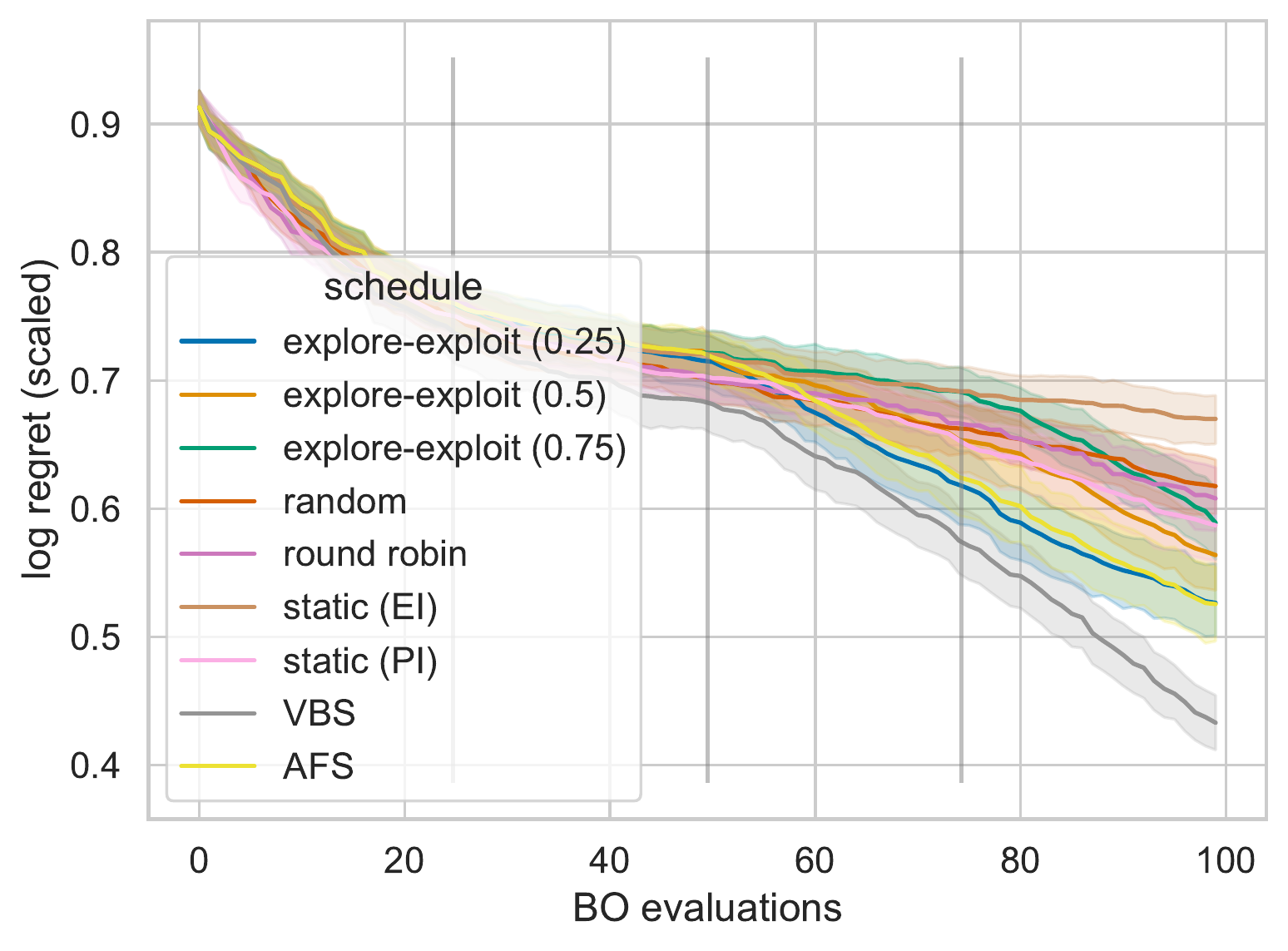}
        \caption{Log-Regret (Scaled) per Step}
        \label{subfig:convergence_24}
    \end{subfigure}\\
    \vspace*{3mm}
    \centering
    \begin{subfigure}[b]{\textwidth}
        \centering
        \includegraphics[width=\textwidth]{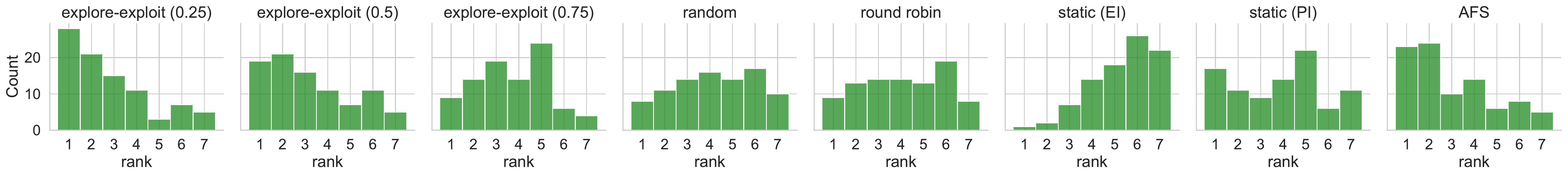}
        \caption{Rank}
        \label{subfig:rank_24}
    \end{subfigure}
    \caption{BBOB Function 24}
    \label{fig:bbob_function_24}
\end{figure}


\end{document}